\documentclass[10pt,twocolumn,letterpaper]{article}

\usepackage[pagenumbers]{wacv}

\usepackage{graphicx}
\usepackage{amsmath}
\usepackage{amssymb}
\usepackage{booktabs}
\usepackage[shortlabels]{enumitem}
\usepackage{multirow} 
\usepackage{comment}
\usepackage[nolist]{acronym}
\usepackage{makecell}
\usepackage{pifont}%
\usepackage{mathtools}
\usepackage{mathrsfs}
\usepackage{tikz}
\newcommand*\circled[1]{\tikz[baseline=(char.base)]{
            \node[shape=circle,draw,inner sep=1.1pt] (char) {#1};}}
\usepackage{wrapfig}
\usepackage{xcolor,colortbl}
\usepackage[export]{adjustbox}
\usepackage[column=0]{cellspace}

\usepackage{multibib}
\newcites{supp}{Supplementary References}

\usepackage{cuted}

\usepackage[pagebackref,breaklinks,colorlinks]{hyperref}

\usepackage[capitalize]{cleveref}
\crefname{section}{Sec.}{Secs.}
\Crefname{section}{Section}{Sections}
\Crefname{table}{Table}{Tables}
\crefname{table}{Tab.}{Tabs.}

\definecolor{col1}{HSB}{0, 255, 192}
\definecolor{col2}{HSB}{51, 255, 192}
\definecolor{col3}{HSB}{102, 255, 192}
\definecolor{col4}{HSB}{135, 255, 192}
\definecolor{col5}{HSB}{204, 255, 192}

\newcommand{\sm}{Suppl. Mat. }

\def\wacvPaperID{148} %
\def\confName{WACV}
\def\confYear{2025}

\begin{document}
    \title{When Cars meet Drones: Hyperbolic Federated Learning for Source-Free Domain Adaptation in Adverse Weather}
    
    \author{Giulia Rizzoli*, Matteo Caligiuri*, Donald Shenaj, Francesco Barbato, Pietro Zanuttigh \\
    University of Padova, Italy  \\
    }
    
    \maketitle
    
    \begin{acronym}
    \acrodef{fl}[FL]{Fed\-er\-at\-ed Learn\-ing}
    \acrodef{freeda}[\textsc{FFreeDA}]{Fed\-er\-at\-ed source Free Do\-main Ad\-ap\-ta\-tion}
    \acrodef{gan}[GAN]{Ge\-ne\-ra\-ti\-ve Ad\-ver\-sa\-rial Net\-work} 
    \acrodefplural{gan}[GANs]{Ge\-ne\-ra\-ti\-ve Ad\-ver\-sa\-rial Net\-works}
    \acrodef{od}[OD]{Ob\-ject De\-tec\-tion}
\end{acronym}

    \begin{abstract}
    In \acf{fl}, multiple clients collaboratively train a global model without sharing private data.
    In semantic segmentation, the \acf{freeda} setting is of particular interest, where clients undergo unsupervised training after supervised pretraining at the server side.
    While few recent works address \ac{fl} for autonomous vehicles, intrinsic real-world challenges such as the presence of adverse weather conditions and the existence of different autonomous agents are still unexplored. To bridge this gap, we address both problems and introduce a new federated semantic segmentation setting where both car and drone clients co-exist and collaborate. 
    Specifically, we propose a novel approach for this setting which exploits a batch-norm weather-aware strategy to dynamically adapt the model to the different weather conditions, while hyperbolic space prototypes are used to align the heterogeneous client representations.
    Finally, we introduce FLYAWARE, the first semantic segmentation dataset with adverse weather data for aerial vehicles.
\end{abstract}

    \begingroup
    \renewcommand\thefootnote{}  %
    \footnotetext{* Equal contribution.}
    \endgroup

    \section{Introduction} \label{sec:intro}
The field of autonomous driving has evolved significantly beyond the traditional concept of cars navigating city streets. In the contemporary landscape, autonomous navigation encompasses various perspectives, including drones and robots, serving a wide range of applications like autonomous delivery and surveillance. The demand for intelligent systems that can seamlessly generalize across these varying viewpoints is quickly increasing, to enhance safety, efficiency, and adaptability.
A key challenge lies in achieving this goal while preserving confidentiality and data security. The importance of robust privacy measures in developing autonomous systems cannot be overstated, as these systems often rely on substantial amounts of data to function effectively. The traditional method of gathering and analyzing data in a centralized manner raises relevant security concerns. %
Federated learning allows tackling this issue by introducing a cutting-edge paradigm that enables the collaborative training of machine learning models without sharing the private data of the participants (\ie, clients).
\begin{figure}[t]
    \centering
    \includegraphics[width=\linewidth]%
    {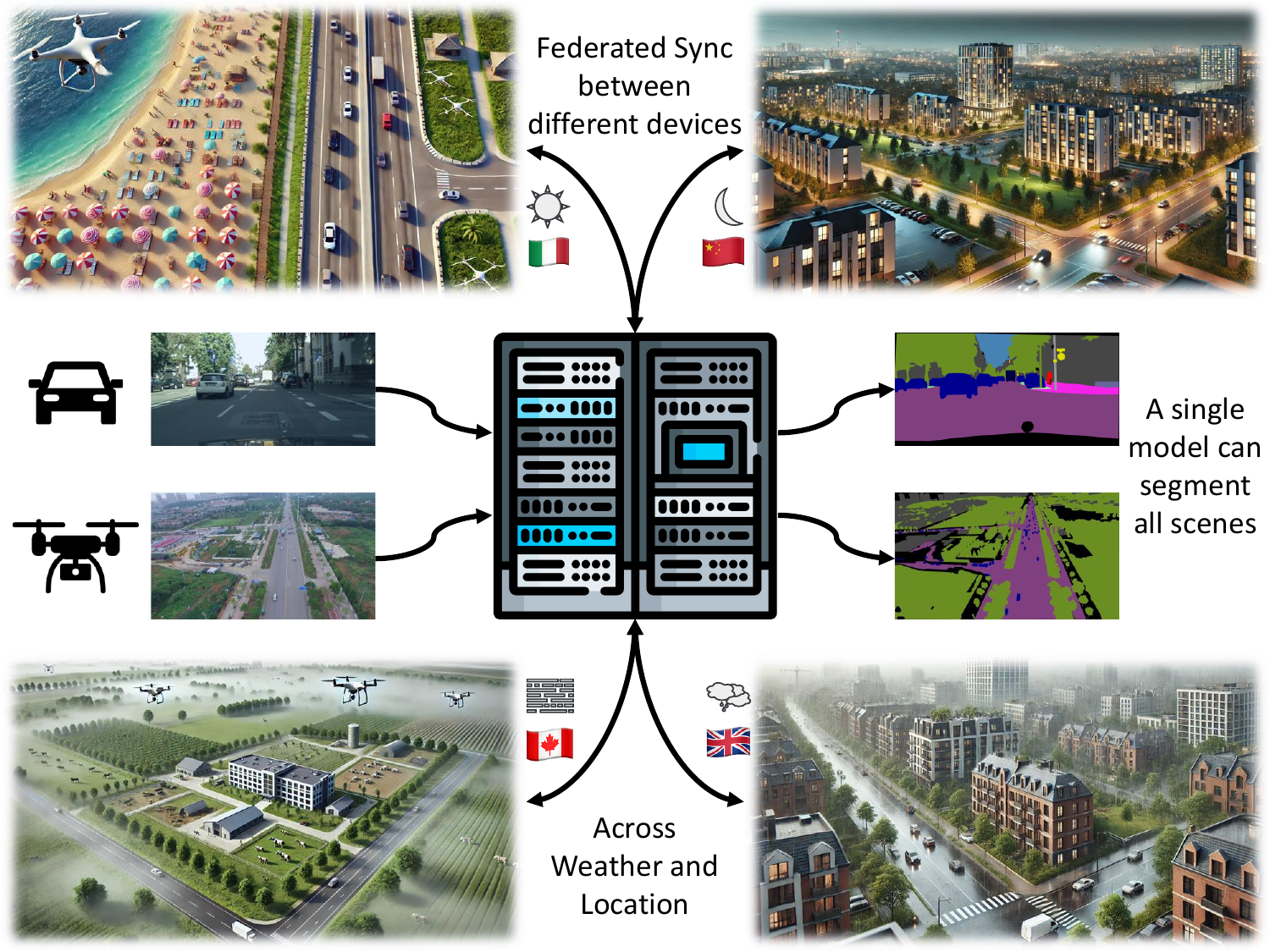}
    \caption{Federated Learning for autonomous driving across diverse conditions: each client trains the segmentation model locally with its own data, which may be biased towards specific environmental conditions. E.g., a taxi operating predominantly at night will have most training samples in nighttime conditions. Despite the local biases, the approach aims to create a single, robust model capable of  generalizing across diverse driving scenarios.
    }
    \label{fig:overview}
\end{figure}
In this paper, we introduce a novel federated learning approach tailored to the specific challenges encountered in driving scenarios, where autonomous agents have different perspectives, specifically the street-view vision for cars and the aerial for UAVs. Multi-viewpoint paradigms have been proven beneficial in various tasks including localization and segmentation \cite{sarlin2024snap,di2020sceneadapt,klinghoffer2023towards}, where jointly training the model with images from different viewpoints improves the objective task. Consequently, we enhance our framework by integrating aerial data during the federated learning process. Importantly, while our approach leverages multi-viewpoint data to learn a more generalizable model, it does not require matching views at inference or training time for individual clients. Each client operates solely with its own perspective, yet benefiting from the collective knowledge gained through federated learning across diverse viewpoints. 

Additionally, we dynamically adapt to ever-changing weather conditions, aligning with the real-world challenges in this dynamic field.
We combine this strategy with the federated paradigm proposed in \cite{shenaj2023learning}, which employs supervised server pretraining to enable unsupervised client-side training. 
In this scenario (shown in \cref{fig:overview}), several key challenges arise: \circled{1}  Clients operating without supervision tend to become overly confident in their predictions, causing the model to diverge; \circled{2} Clients exhibit differing distributions, and varying weather conditions can significantly impact performance (i.e., clients are non-iid, \textit{heterogeneous}); \circled{3} Clients are unbalanced, reflecting real-world scenarios where certain types of agents may be less common (e.g., fewer drones compared to cars).

\noindent To address these challenges, we present several novel contributions:
\begin{itemize}
\itemsep0em 
\item We introduce an unsupervised federated system for learning a global model which can generalize between cities, weather and perspectives.
\item We provide \textbf{FLYAWARE}, the first semantic segmentation dataset with adverse weather imagery for aerial vehicles. 
\item We propose \textbf{HyperFLAW} \footnote{The code and dataset are available at \url{https://github.com/LTTM/HyperFLAW}} 
(\textbf{Hyper}bolic \textbf{F}ederated \textbf{L}earning in \textbf{A}dverse \textbf{W}eather), which addresses domain shifts in diverse weather conditions using weather-aware batch normalization layers.
\item We exploit prototype-based learning in the hyperbolic space to ensure consistent training across the clients in this very challenging setting.
\end{itemize}

    \section{Related Work} \label{sec:related}
\textbf{Domain Adaptation} (DA) primarily focuses on adapting a model from a source to a target domain, typically in scenarios where the target data lacks labels, referred to as Unsupervised Domain Adaptation (UDA) \cite{schwonberg2023survey,toldo2020unsupervised}.
Historically, DA methods aimed to bridge the gap by quantifying domain divergence \cite{tzeng2014deep,saito2018maximum}. Adversarial learning has also been widely used \cite{tsai2018learning,luo2019taking,michieli2020adv}. %
In the contemporary landscape, advanced approaches  \cite{zou2018unsupervised,hoyer2022daformer,mei2020instance} leverage self-learning techniques. %
Several other approaches have been used to align the domains such as image-to-image translation~\cite{yang2020fda,araslanov2021self,choi2019self}, pseudo-label prototypes \cite{zhang2019category,liu2021bapa,zhang2021prototypical}, and confidence thresholding \cite{mei2020instance,zou2018unsupervised}. Within the objective of adapting to adverse weather conditions, Bruggemann et al. \cite{bruggemann2023refign}  propose a solution that requires sunny daytime references.
To tackle nighttime images, Xiao et al. \cite{xia2023cmda} employed a data augmentation method leveraging multiple modalities. %
Finally, Source-Free Domain Adaptation presents a distinct advantage in addressing the performance gap by allowing adaptation without access to labeled source data 
\cite{li2024comprehensive}.

\label{sub:related:federated}
\noindent\textbf{Federated Learning} has captured the interest of the research community due to its practical applicability in various real-world scenarios and its potential to manage sensitive data \cite{fedavg}. 
In literature, it has been applied to different vision tasks \cite{shenaj2023federated}, initially with a large emphasis on classification  \cite{li2020federated,li2021model}, but more recently also focusing on other tasks including semantic segmentation for Autonomous Driving \cite{chellapandi2023federated,fantauzzo2022feddrive,shenaj2023learning}.
Most existing works in FL assume labeled data on remote clients \cite{chellapandi2023federated,fantauzzo2022feddrive}, which is an unrealistic assumption, especially in the driving scenario. 
Notably, Shenaj et al. \cite{shenaj2023learning} introduce the Federated source Free Domain Adaptation (\textsc{FFreeDA}) setting, where only the server accesses a labeled source dataset. We address this challenge by incorporating the management of diverse viewpoints, accommodating various client types, and dealing with the influence of adverse weather conditions.

\label{sub:related:hyperbolic}
\noindent\textbf{Hyperbolic Prototypical Learning. }
Working in Euclidean spaces, implicitly or explicitly assume that data is best represented on regular grids. While this model offers an intuitive and grounded underlying manifold, its properties may not be the most suitable for all types of data~\cite{mettes2023hyperbolic}. Inspired by developments in other fields,  deep learning in hyperbolic space has recently been exploited in computer vision using different isomorphic models, such as the Poincar\'{e} model. 
Several works have proposed prototypes-based hyperbolic embeddings for few-shot learning for classification \cite{khrulkov2020hyperbolic,guo2022clipped,ma2022adaptive}, where hyperbolic space offers advantages over the Euclidean one.
In particular, Khrulkov et al.~\cite{khrulkov2020hyperbolic} demonstrated its competitiveness on simple ConvNets %
while Gao et al.~\cite{gao2021curvature} show the benefits achievable with different curvatures.
Atigh et al.~\cite{atigh2022hyperbolic} introduced approximations for faster computations tailored for use in semantic segmentation. Facing these insights, hyperbolic prototypical learning seems suitable for learning features robust to domain shift and varying weather conditions even when employing different curvatures and lightweight networks,  as those used on clients in  federated  scenarios. Building upon these foundations, our approach exploits hyperbolic learning to align features from diverse viewpoints while reducing domain shift.

    \section{Problem Formulation} 
\label{sec:problem}
Most works in Federated Learning operate under the assumption of labeled data being available on remote clients \cite{chellapandi2023federated,fantauzzo2022feddrive}. Providing semantic segmentation ground truth is extremely time-consuming and costly, thus it is quite unrealistic to expect labeled data on real-world clients, especially during the deployment phase.
Therefore, our objective is to develop a model that can seamlessly adapt from labeled synthetic data to unlabeled real-world images in a distributed paradigm. 
For this reason, we chose to deploy our setting on top of the pre-existing \textsc{FFreeDA} scenario \cite{shenaj2023learning}. Moreover, we discard the assumption that the clients are homogeneous in nature and consider a setting where different autonomous agents work cooperatively in fluctuating atmospheric conditions (\ie, clear sky, night, rain and fog).
In the following, we will first formalize the problem and then move to a detailed description of the methodologies proposed to tackle it in \cref{sec:method}.\\
\noindent\textbf{Source-Free Domain Adaptation (SFDA).}
Consider an image $x \! \in \!\! \mathcal{X} \! \subset \!\! \mathbf{R}^{|\mathcal{I}| x 3}$ and its corresponding label $y \in \mathcal{Y} \subset \mathbf{R}^C$, where $|\mathcal{I}| = H \times W$ are the image dimensions and $C$ is the number of classes.
SFDA entails a pretraining phase on a source dataset %
$\mathcal{D}^S = \{(x^s, y^s)\}^{N^s}$, made of pairs of images and labels, and an adaptation stage on a target dataset %
$\mathcal{D}^T = \{(x^T)\}^{N^T}$, composed of unlabeled images. Specifically, compared to standard DA: 1) the two training phases are distinct; 2) $\mathcal{D}^S$ is unavailable at the adaptation phase.

\noindent\textbf{Federated source-Free Domain Adaptation (\textsc{FFreeDA}).} This setting deals with a distributed setup comprising a central server and a set of clients $\mathcal{K}$ with $\left | \mathcal{K} \right | = K$.
The training datasets are organized as follows: the primary dataset $\mathcal{D}^S$, resides on the server side. Concurrently, each of the $K$ client has a distinct training dataset denoted as $\mathcal{D}^T_k = \{x_{k,i}^T\in \mathcal{X} : \forall i \in |\mathcal{D}^T_k|\}$. Notably, these client datasets are locally stored and managed, ensuring that $\mathcal{D}^T_1 \cap \mathcal{D}^T_2 \cap \ldots \cap \mathcal{D}^T_K = \varnothing$.
In distributed settings, the parameter \( K \) is reasonably large, and the local datasets vary in terms of statistics, \ie size and distribution. However, they typically possess a much smaller size compared to the source dataset (\( N^S \! \gg N^T \)). \textsc{FFreeDA} assumes that the local datasets are drawn from the same meta-distribution, meaning that \( \mathcal{D}_k^T \) comprises images solely from one latent domain.
The objective is to achieve optimal segmentation performance on the target data distribution with the model \( M(\theta): \mathcal{X} \rightarrow \mathbf{R}^{| \mathcal{I}| \times|\mathcal{Y}|} \), parameterized by \( \theta \). This goal is pursued by minimizing an appropriate loss function, expressed as:
\begin{equation}
   \theta^* = \arg\min_{\theta} \sum_{k\in[K]} \frac{|\mathcal{D}_k^T|}{|\mathcal{D}^T|}\mathcal{L}_k(\theta)
   \label{eq:fedavg}
\end{equation}
where \( \mathcal{L}_k \) represents the local loss function, and \( \mathcal{D}^T=\bigcup_{k \in \mathcal{K}} \mathcal{D}^T_k \).\\
\noindent\textbf{Multi-viewpoint \textsc{FFreeDA} in Adverse Weather.} Aiming to assist agents lacking large training datasets or capabilities, such as drones, we consider a setting that enables scene understanding for unsupervised autonomous navigation robust to multiple viewpoints. First of all, we consider $\mathcal{D}^S = \mathcal{D}^S_{car} \cup \mathcal{D}^S_{drone}$. Accounting for adverse weather as a heterogeneity factor, the source data are sampled as $\{(\mathbf{x}^{S}, \mathbf{y}^{S}, w^{S})\}^{N^S}$ which includes $w^S$, denoting the atmospheric condition for each frame, together with the color images $\mathbf{x}^S$ and the ground truth segmentation maps $\mathbf{y}^S$.
Similarly, the client can accommodate either car or drone agents:  $\mathcal{D}^T = \mathcal{D}^T_{car} \cup \mathcal{D}^T_{drone}$, although client datasets are still local and disjoint. 
Please note that within the target set, the drone and car data are considered to have been acquired separately, i.e., they belong to distinct domains. This assumption reflects the practical constraints of real-world scenarios, where obtaining matching viewpoints between aerial %
and terrestrial %
imagery is typically infeasible.

\begin{figure*}[tp]
    \centering
    \includegraphics[width=0.9\textwidth]{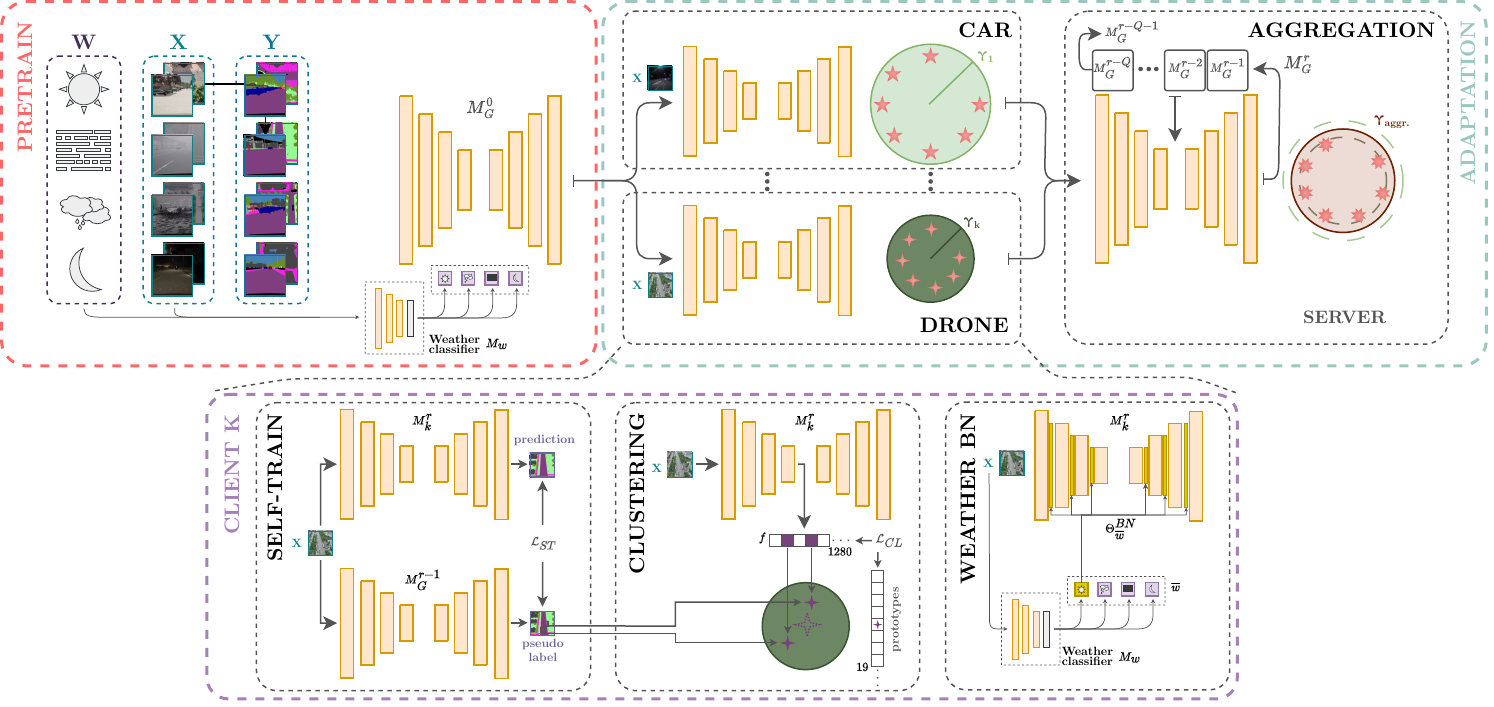}
    \caption{%
    The two-step training process of HyperFLAW: server pretraining on synthetic data and client-side real-world adaptation. We highlight the active modeling of weather conditions using weather-aware batch norm layers and the use of hyperbolic space for feature alignment via prototypical learning, ensuring consistency in features extracted by car and drone agents across various atmospheric conditions.
    At the server, the aggregation with the queue of previous global models reduces the instability introduced by aggregating unlabeled clients.
    }
    \label{fig:architecture}
\end{figure*}

As in the previous setting, each client $k \in \mathcal{K}$ have its own set of unsupervised images both in terms of semantics and weather. %
Notably, the number of samples of the drones' target dataset ($N^T_{drone}$) is typically lower than in the target datasets for cars ($N^T_{car}$). Furthermore, the set of semantic classes could differ between car and aerial vehicles datasets (in our case $C_{drone} \subset C_{car} = C^T = C^S$). 
The identity of each client is fixed, meaning it is designated as either a car or a drone and does not change during training, however, there is no such limitation on the weather conditions, which can change dynamically during the training process. %
The goal is to minimize the objective function of \cref{eq:fedavg}, where the target data distribution holds into its heterogeneity different viewpoints and weathers. 
In \cref{fig:architecture} we present our approach, which will be detailed in \cref{sec:method}.

    \section{Training Procedure} \label{sec:method}
The training is organized in a set of rounds $r=0, \ldots, R$, where $r=0$ corresponds to the pretraining phase. %
At each round $r$, we denote by $M^r_G$ the global model at the server side, while we refer to $M^r_k$ as the model of the $k-th$ client.

\textbf{Pretraining.} We train the initial model $M^0_G$ on a source synthetic dataset $\mathcal{D}^S$, taking advantage of the server's computing capabilities. By sampling batches of cross-agent we maximize the standard cross entropy loss $\mathcal{L}(y^S, M_G^0(x^S))$. Moreover, an additional component $M_W$ consisting of 3 convolutional layers is trained to recognize weather conditions by optimizing $\mathcal{L}(w^S, M_W(x^S))$. This classifier is a lightweight module intended to provide weather information for clients and it is not trained after this phase.
Differently from \cite{shenaj2023learning}, we do not assume to have information on the clients' distribution (i.e., in \cite{shenaj2023learning} the average Fourier domain coefficients from $\mathcal{D}^T$ are used, requiring priors on clients' data).

\textbf{Adaptation.} At each round $r$, we consider that, among all the set of clients $\mathcal{K}$, only a subset denoted as $\mathcal{K}^r$ participate in the training. This assumption is common in FL where not all the clients might be concurrently active and reflects well on our scenario (e.g. a car is parked).  Then, the server transmits the global model from the previous round $M_G^{r-1}$. Note that the weather classifier $M_W$ is only transmitted once as it is fixed. The training for the active clients will be performed separately on $M_k^r, k \in \mathcal{K}^r$. 
Client-side training employs self-training (\cref{sec:client_training}), weather-batch normalization to align feature styles across climate conditions (\cref{sec:weather_BN}), and, finally, prototypical learning for feature space generalization (\cref{sub:method:prototypes}). Each client generates class prototypes $\mathbf{p_{c,k}^r}$, i.e., centroids of class samples in feature space, and computes the manifold curvature $\gamma_k$, both of which are transmitted to the server.
The server aggregates the client models as described in \cref{sec:server}, producing the global model ${M}_{G}^r$ for the next round $r+1$. Global class prototypes $\mathbf{p^{r}_{G,c}}$ are also aggregated from the client prototypes.

\subsection{Client Self-Training (ST)} 
\label{sec:client_training} 

Self-training involves iteratively enhancing a model’s performance by using its own predictions.
In FL, the clients can rely on the server model as the teacher, as ideally, the server aggregates the global knowledge from the targets. %
In our framework, we update the teacher model after the round is over. This implies that the global model received by the clients at the beginning of each round $r$, %
\(M_G^{r-1}\), serves also as the current teacher. %
This approach offers the advantage of eliminating the necessity to transmit two models per round as in \cite{shenaj2023learning} %
where also a model from $R_{u}$ steps before was used, thus requiring only a single model transmission and avoiding the extra hyperparameter $R_{u}$.
The local model of the $k-th$ client is trained using the weak supervision of the pseudo-labels produced by the teacher model $M_G^{r-1}$ and its local model prediction $M^r_k$ by optimizing the self-training loss obtained via cross-entropy as $\mathcal{L}_{st}(M_G^{r-1}(x_k^T), M^r_k(x_k^T))$.
Moreover, to achieve a training that is consistent across the heterogeneous clients 
and weather conditions, the self-training loss is used in combination with the prototype-based one, i.e., 
\begin{equation} \label{eq:local_training}
    \mathcal{L}_k(\theta)= \mathcal{L}_{st}(\theta) + \lambda_{cl}\mathcal{L}_{cl}(\theta)%
\end{equation}
where $\theta$ are the model parameters, $\mathcal{L}_{cl}$ is the clustering loss computed using prototypical representation (see \cref{sub:method:prototypes}), and  $\lambda_{cl}$ balances the two losses.

\subsection{Weather Batch Normalization (BN)}
\label{sec:weather_BN}
\begin{figure}
    \centering
    \includegraphics[width=0.9\columnwidth]{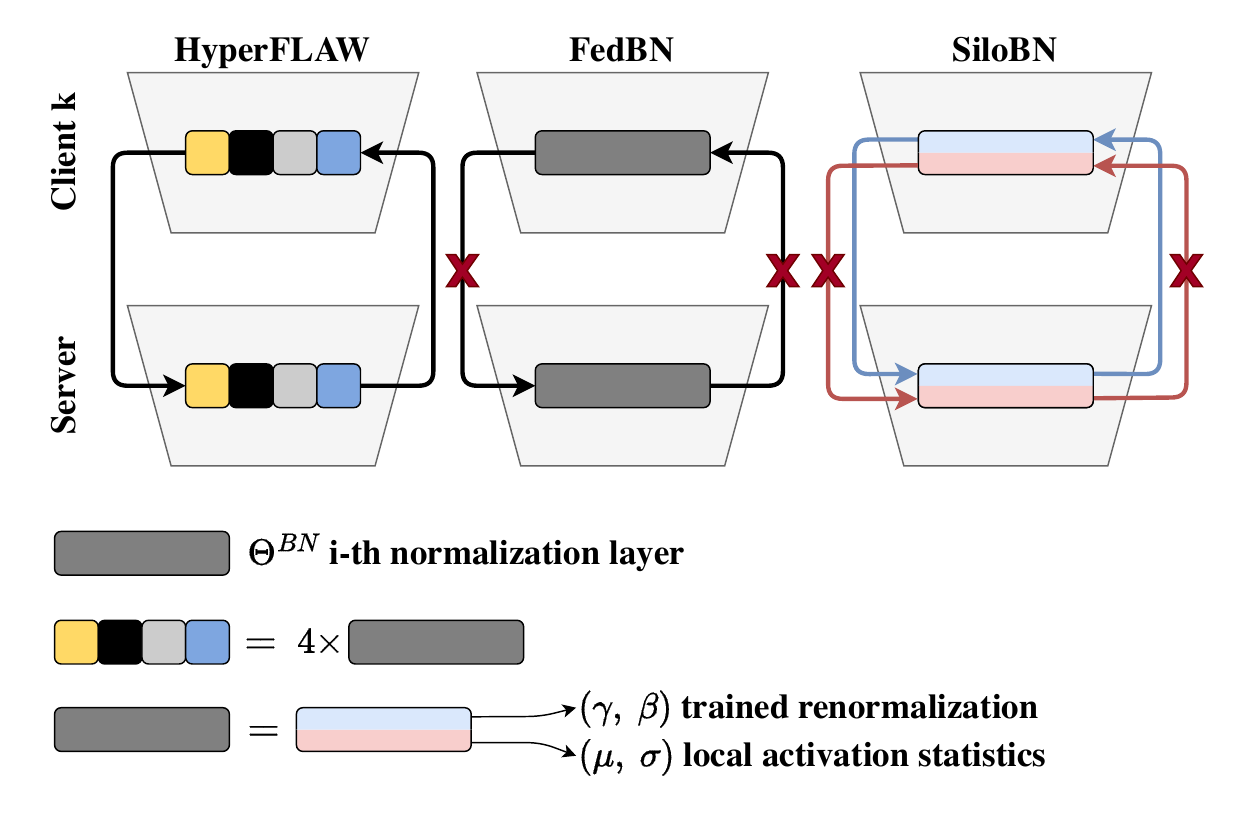}
    \caption{Comparison of Weather-Batch Normalization with BN personalization methods \cite{li2021fedbn,andreux2020siloed}.
    }
    \label{fig:bns}
\end{figure}

Recent methods \cite{li2021fedbn,andreux2020siloed} have explored unsynchronized batch normalization layers to tailor local clusters. Although offering local personalization, these approaches may be suboptimal in driving scenarios, especially when considering real-world applications. Imagine a situation where each client predominantly drives in specific weather conditions, resulting in training samples primarily drawn from a single weather distribution. Ideally, the model should perform equally well across all weather conditions, not just those frequently encountered by individual clients. To address this challenge, we propose a novel strategy: rather than customizing for each client, we adapt the training process for each distinct weather condition. This is accomplished by utilizing a separate set of batch-normalization parameters ${\theta^{BN}_w}$ for each weather condition $w$ (as illustrated in \cref{fig:bns}).

After the pretraining ($r=0$), all the weather-specific BNs inherit the global BN statistics $\theta^{BN}$, while in each subsequent round, the statistics of the BNs are averaged only over those of the same weather condition.
Specifically, on the client side, after obtaining the prediction $\bar{w} = M_W(x^T)$ from the classifier, the client batch-norm parameters are optimized using Eq. \ref{eq:local_training} only on the selected weather-BNs $\theta^{BN}_{\bar{w}}$. 
This allows us to enforce a more general feature extraction from the convolutional layers while exploiting the domain-specific nuances of road scenes under different weather conditions.

\subsection{Prototypical Learning} \label{sub:method:prototypes}
While the various devices explore road scenes, their distinct viewpoints force the local models to conform more closely to the features within each specific perspective. 
In the pretraining phase, drone and car samples are provided jointly, however, this simultaneity does not apply to clients, thereby the network predictions could diverge, making the subsequent aggregation on the server side more challenging.
To ensure the ongoing alignment of features across the two data types, we propose the exploitation of joint class prototypes as a viable solution.
For each client, we compute a set of class prototypes $\mathcal{P}$, where each prototype $\mathbf{p}_{c} \in \mathcal{P}$ is defined as a function of the set $\mathcal{F}_c$ of all feature vectors $\textbf{f}_{c}$ from the encoder output  that are associated to class $c$:%
\begin{equation}
    \mathbf{p_c} = g(\mathcal{F}_c), \; 
        \mathcal{F}_c \!=\ \{\mathbf{f} = E(\mathbf{x})(h,w) : \; \hat{\mathbf{y}}(h,w) = c\} 
        \end{equation}
where $\hat{\mathbf{y}}$ are the teacher pseudo-labels subsampled at the feature space resolution and the function $g$ computes either the mean or the midpoint depending on the considered manifold: Euclidean or Poincar\'e Ball, respectively.

At the beginning of each round $r$, a participating client $k$ initializes its local prototypes $\mathbf{p_{k,c}^{r,0}}$ using the previous round ones received from the server: $\mathbf{p_{k,c}^{r,0}} = \mathbf{p_{G,c}^{r-1}}$.
Then, at each training step $s$ the class prototypes at client side  $ %
\mathbf{p_{k,c}^{r,s}}$ are updated using exponential smoothing with rate $\beta = 0.85$ to stabilize their evolution as follows:
\begin{equation}
    \mathbf{p_{k,c}^{r,s}} = \beta \; \mathbf{p_{k,c}^{r,s-1}} + (1 - \beta) \;  \mathbf{p_{k,c}^{r,s}}
    \label{eq:smooth1}
\end{equation}
where $\mathbf{p_{k,c}^{r,s-1}}$ is the prototype from the previous training step.
After each round, prototypes  $\mathbf{p_{k,c}^r}, \forall c \in \mathcal{C}$ computed by each client are sent back to the server for aggregation. 

\noindent\textbf{Hyperbolic Protoypes.} As mentioned before, to further improve performance, we exploited the capacity of hyperbolic space - specifically of the Poincar\'e ball model - to fit complex feature geometries better and used it to embed the prototypes. 
We refer to \cite{mettes2023hyperbolic,van2023poincar} for further discussion. 
The hyperbolic space represented in Riemannian geometry, with dimension $n$ and constant negative curvature $-\gamma$ is expressed as:
\begin{equation}
    \mathbb{B}_{\gamma}^{n} = \{\mathbf{x} \in \mathbb{R}^n: \|\mathbf{x}\|^2 <  1 / \gamma \}
\end{equation}
On such space, the distance between two points $\mathbf{x}$, $\mathbf{y} \in \mathbb{B}_{\gamma}^{n}$ can be computed as:
\begin{equation}
    d_{\gamma}(\mathbf{x},\mathbf{y}) = %
    2\gamma^{-1/2}\tanh^{-1} \left( \sqrt{\gamma} \|\mathbf{-x} \oplus_{\gamma} \mathbf{y}\|\right)
\end{equation}
where $\mathbf{x} \oplus_{\gamma} \mathbf{y}$ is the M\"{o}bius addition:
\begin{equation}
    \mathbf{x} \oplus_{\gamma} \mathbf{y} = \frac{(1 + 2 \gamma \langle \mathbf{x}, \mathbf{y} \rangle + \gamma \|\mathbf{y}\|^2) \mathbf{x} + (1 - \gamma \|\mathbf{x}\|^2) \mathbf{y}}{1 + 2 \gamma \langle \mathbf{x}, \mathbf{y} \rangle + \gamma^2 \|\mathbf{x}\|^2 \|\mathbf{y}\|^2}
\end{equation}
The translation from the Euclidean to the hyperbolic space is obtained given the exponential mapping in the direction of a tangent vector $\mathbf{v} \in \mathcal{T}\mathbb{B}_{\gamma}^{n}$:
\begin{equation}
    \exp_{\mathbf{x}}^{\gamma}(\mathbf{v}) = \mathbf{x} \oplus_{\gamma} \left( \tanh\left(\frac{\sqrt{\gamma} \|\mathbf{v}\|}{1 - \gamma \|\mathbf{v}\|^2}\right) \frac{\mathbf{v}}{\sqrt{\gamma} \|\mathbf{v}\|} \right)
    \label{eq:expmap}
\end{equation}
The prototypes are determined as the midpoints of the hyperbolic feature vectors $\{\mathbf{f}_{i} := \mathbf{f}^{hyp}_{i} \in \mathbb{B}_{c}^{n}\}$, using an approximation of the Frechet mean:
\begin{equation}
    \mathbf{p} = \frac{1}{2} \otimes_{\gamma} \frac{\sum_{\mathbf{f} \in \mathcal{F}}\lambda_{\mathbf{f}}^{\gamma}\mathbf{f}}{\sum_{\mathbf{f} \in \mathcal{F}}(\lambda_{\mathbf{f}}^{\gamma}-1)}
    \label{eq:protohyp}
\end{equation}
where the M\"{o}bius scalar multiplication is:
\begin{equation}
    r \otimes_{\gamma} \mathbf{x} = \frac{1}{\sqrt{\gamma}} \tanh\left(r \tanh^{-1}(\sqrt{\gamma}\|\mathbf{x}\|)\right) \frac{\mathbf{x}}{\|\mathbf{x}\|}
\end{equation}

Since %
the curvature $\gamma$ is a scalar value, we can convert it to a learnable parameter of the network and optimize it during the training process \cite{spengler2023hypll}. This is done on the client side using self-supervision and prototype loss, then,
at the end of each round, the curvature of the active clients is aggregated using \textit{FedAvg} algorithm, i.e., the average is weighted by the number of seen samples.

\noindent\textbf{Prototypical Loss.} Due to the distribution mismatch between source and target domains and the different viewpoints, feature vectors from these domains become misaligned, leading to inaccuracies in target representations and a subsequent decline in segmentation accuracy. To address this issue, we adopt a clustering objective in the latent space by leveraging prototypical representations that exploits the distance between each extracted feature vector and the prototype available at client side for the corresponding class: %
\begin{equation}
    \mathcal{L}_{cl} = \frac{1}{|\hat{\mathcal{C}}|} \sum_{c \in \hat{\mathcal{C}}} \frac{1}{|\mathcal{F}_c|} \sum_{\mathbf{f_{c} \in \mathcal{F}_c}}%
    d(\mathbf{f}_{c}, \mathbf{p}_{k,c}^r)
    \label{eq:clustering_loss}
\end{equation}
where $\hat{\mathcal{C}} \; \coloneqq \; \{c \in \mathcal{C}: \; |\mathcal{F}_c| > 0\}$ is the set of active classes in the current batch.
\subsection{Server Aggregation (AG)} 
\label{sec:server}
\textbf{Regularization.} As mentioned earlier, the wide range of data variability across client devices, caused by various types of autonomous agents and changing weather conditions, makes aggregation at the server side challenging. Moreover, unrestricted self-training on the client side can lead to instability (see training curve when only $\mathcal{L}_{st}$ is active in  
\cref{fig:ablation})
and to a decrease in performance after several rounds.
To stabilize the results with respect to standard averaging \cite{fedavg}, we used a combination of the models received from the clients with the global ones available at server side from previous steps. {In this way, during the first rounds, there is a smoother adaptation from the pretrained model.
More in detail, the aggregated model on the server side is computed as:

\begin{equation}
    M_G^{r}= \frac{
    \frac{\sum_{k \in \mathcal{K}^R} M^r_k}{|\mathcal{K}^R|} +  \sum_{j=1}^Q M_G^{r-j}} {Q+1}
    \label{eq:ssagg}
\end{equation}
i.e., we get the output of federated averaging and smooth it by averaging with the $Q$ previous rounds, thus making the learning more stable \cite{caldarola2023window, shenaj2023asynchronous}.
Other approaches \cite{shenaj2023learning} tackle regularization by exploiting the global model from several snapshots ago, \ie \( M_G^0 \). The idea behind this is to distill knowledge from past predictions, where the server has higher confidence. However, this introduces additional transmission requirements at each round or necessitates clients to allocate resources for storing \( M_G^0 \). Moreover, clients must perform additional inference and computation for each image. We address these challenges by implementing a strategy that only requires effort from the server side, \ie, by maintaining a queue of previous model snapshots (\( Q \)). We ensure that the burden falls on the server, where memory and computational capabilities are less constrained. We will denote this aggregation strategy as AG in the results to distinguish it from standard FedAvg.

\textbf{Prototype aggregation.} Prototypes are also aggregated on the server side. To this aim, we used a combination of the previously computed prototypes and the ones received by the clients.
To determine the updates, we calculated a weighted average of the prototypes received by the clients and smoothed the result by considering the prototypes from the previous round:

\begin{equation}
    \mathbf{p}_{G,c}^{r+1} = \beta '  \mathbf{p}_{G,c}^{r-1} + (1 - \beta ')  
    \sum_{k \in \mathcal{K}^r} N_{k,c}^r \mathbf{p}_{k,c}^{r}
\end{equation}
where $N_{k,c}^r = | \mathcal{F}_{k,c} |$ at round $r$ and $N_{c}^{r,tot} = \sum_{k \in \mathcal{K}} N_{k,c}$. The  smoothing parameter is the same of (\ref{eq:smooth1}), $\beta' = \beta = 0.85$.\\

    \begin{table}
    \centering
    \resizebox{.8\columnwidth}{!}{
      \begin{tabular}{cccccccc}
           \toprule & &      & \textbf{Clear} & \textbf{Night} & \textbf{Rain}  & \textbf{Fog}   \\
          \midrule
          \multirow{4}{*}{\rotatebox{90}{SYNTH}}
          & \multirow{2}{*}{\textbf{SELMA}}          & Train & 24735 & 24735 & 24735 & 24735 \\
           & &Test  & 3087  & 3087  & 3087  & 3087  \\
           \cline{2-7}
          & \multirow{2}{*}{\textbf{FLYAWARE-S}}   & Train & 24771 & 24731 & 24808 & 24631 \\
           & & Test  & 3147  & 3139  & 3128  & 2935  \\
           \hline
           \multirow{6}{*}{\rotatebox{90}{REAL}}
           &\multirow{2}{*}{\textbf{ACDC$^\dagger$}} & Train & 1060  & 400   & 400   & 400   \\
           & & Test  & 214   & 100   & 100   & 100
          \\
           \cline{2-7}
          & \multirow{2}{*}{\textbf{FLYAWARE-R}}      & Train & 94    & 35    & 35    & 35    \\
           & & Test  & 18    & 18    & 17    & 17    \\
          & \multirow{2}{*}{\textbf{FLYAWARE-R-XL}}      & Train & 565    & 565    & 565    & 565    \\
           & & Test  & 112    & 53    & 52    & 52    \\
          \bottomrule
      \end{tabular}
    }
    \caption{Data distribution. $\dagger$: Cityscapes as clear, see \sm}
    \label{tab:adverse}
\end{table}

\begin{figure}[tp]
    \centering
    \begin{subfigure}{\textwidth}
        \rotatebox{90}{~\tiny{FLYAWARE-S}}
        \begin{subfigure}{.11\textwidth}
            \includegraphics[width=\textwidth]{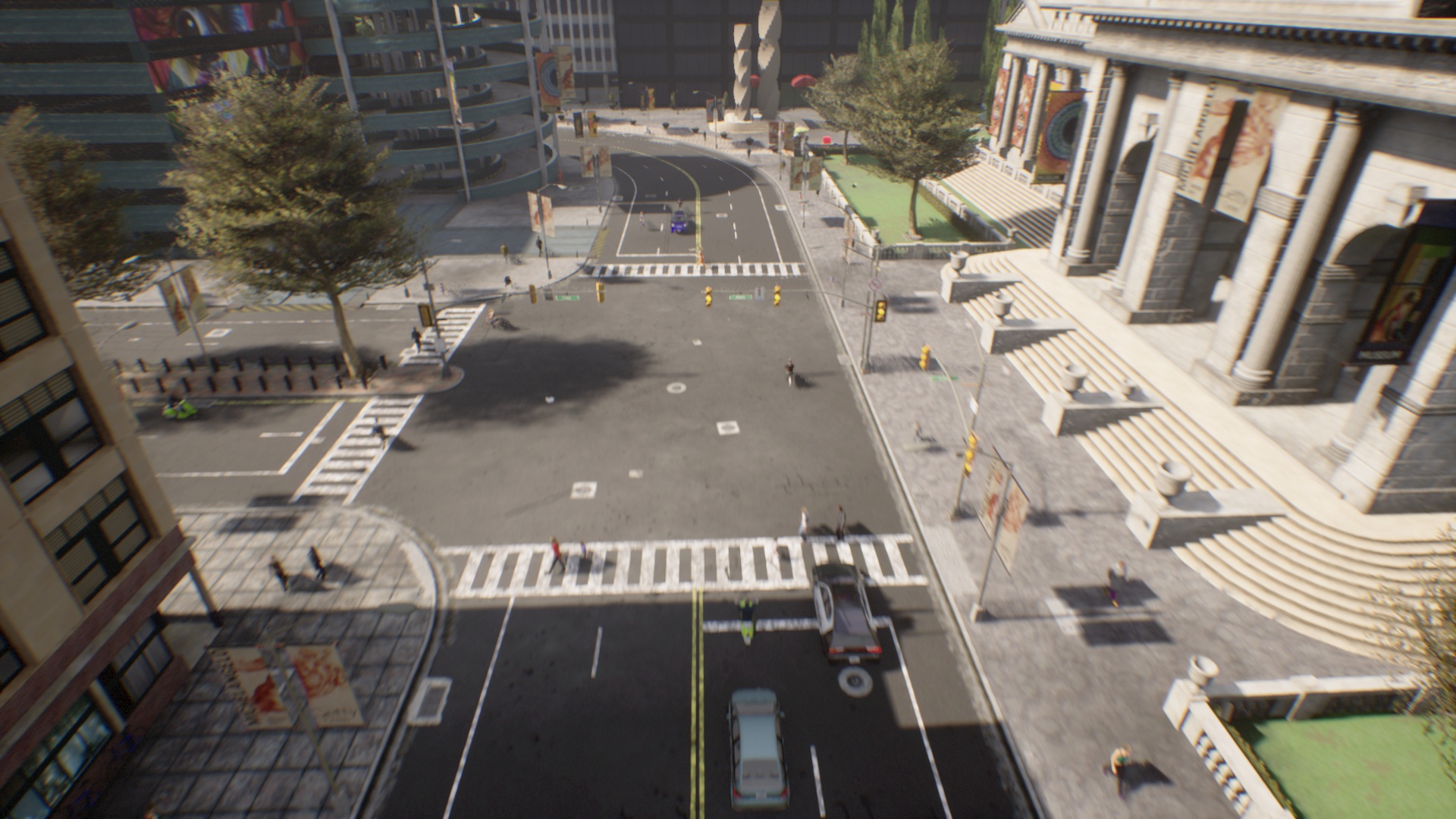}
        \end{subfigure}
        \begin{subfigure}{.11\textwidth}
            \includegraphics[width=\textwidth]{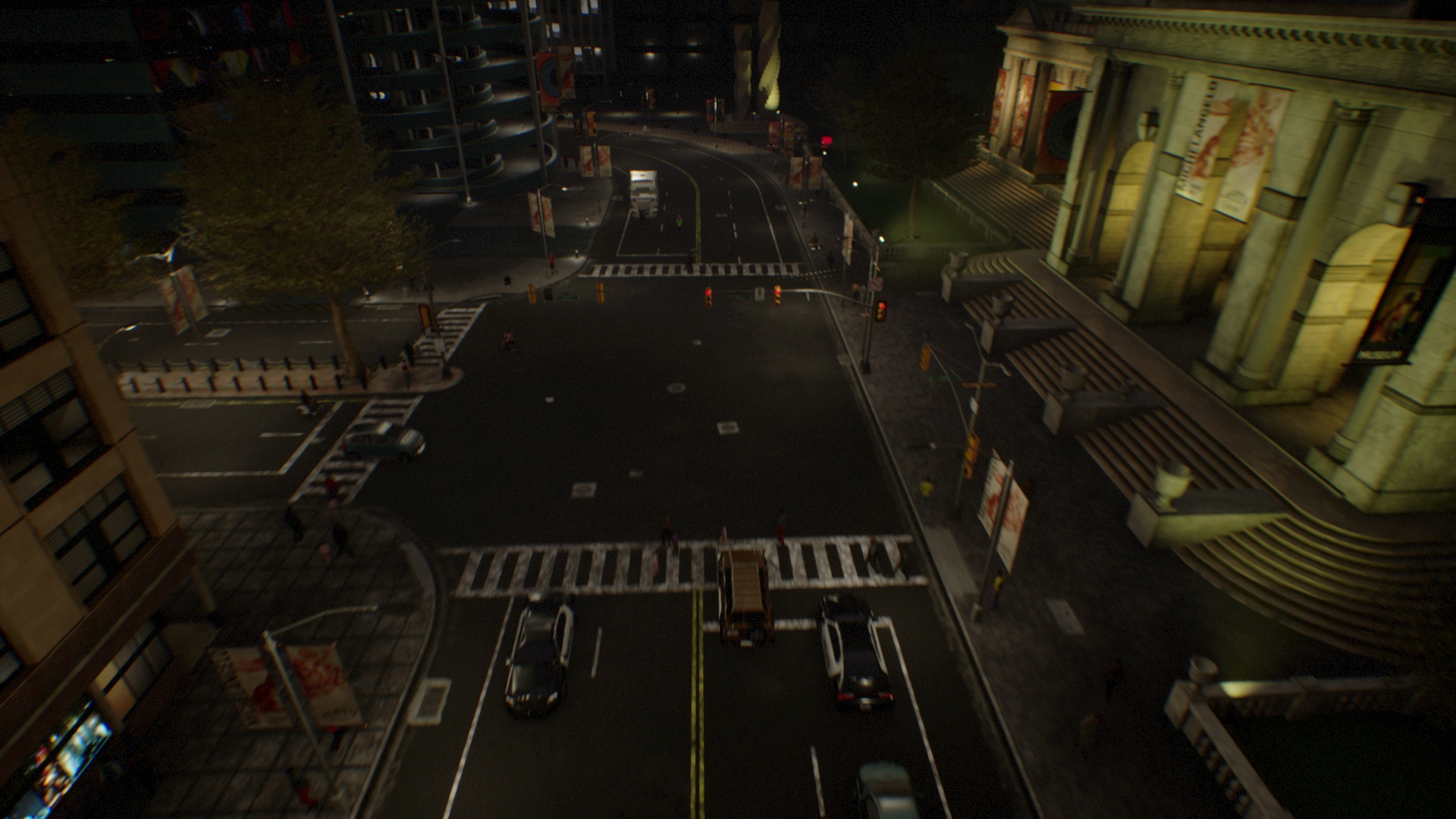}
        \end{subfigure}
        \begin{subfigure}{.11\textwidth}
            \includegraphics[width=\textwidth]{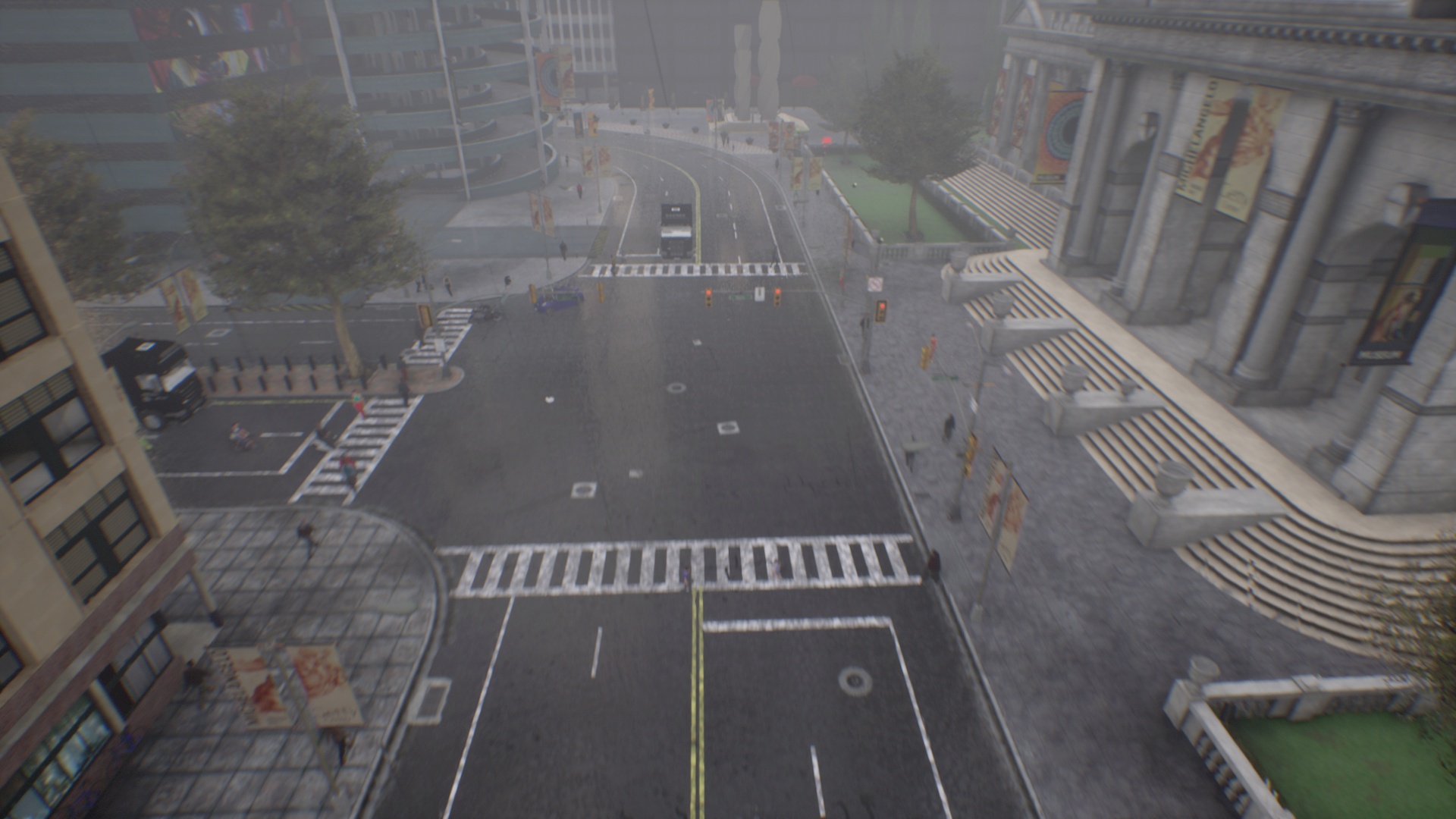}
        \end{subfigure}
        \begin{subfigure}{.11\textwidth}
            \includegraphics[width=\textwidth]{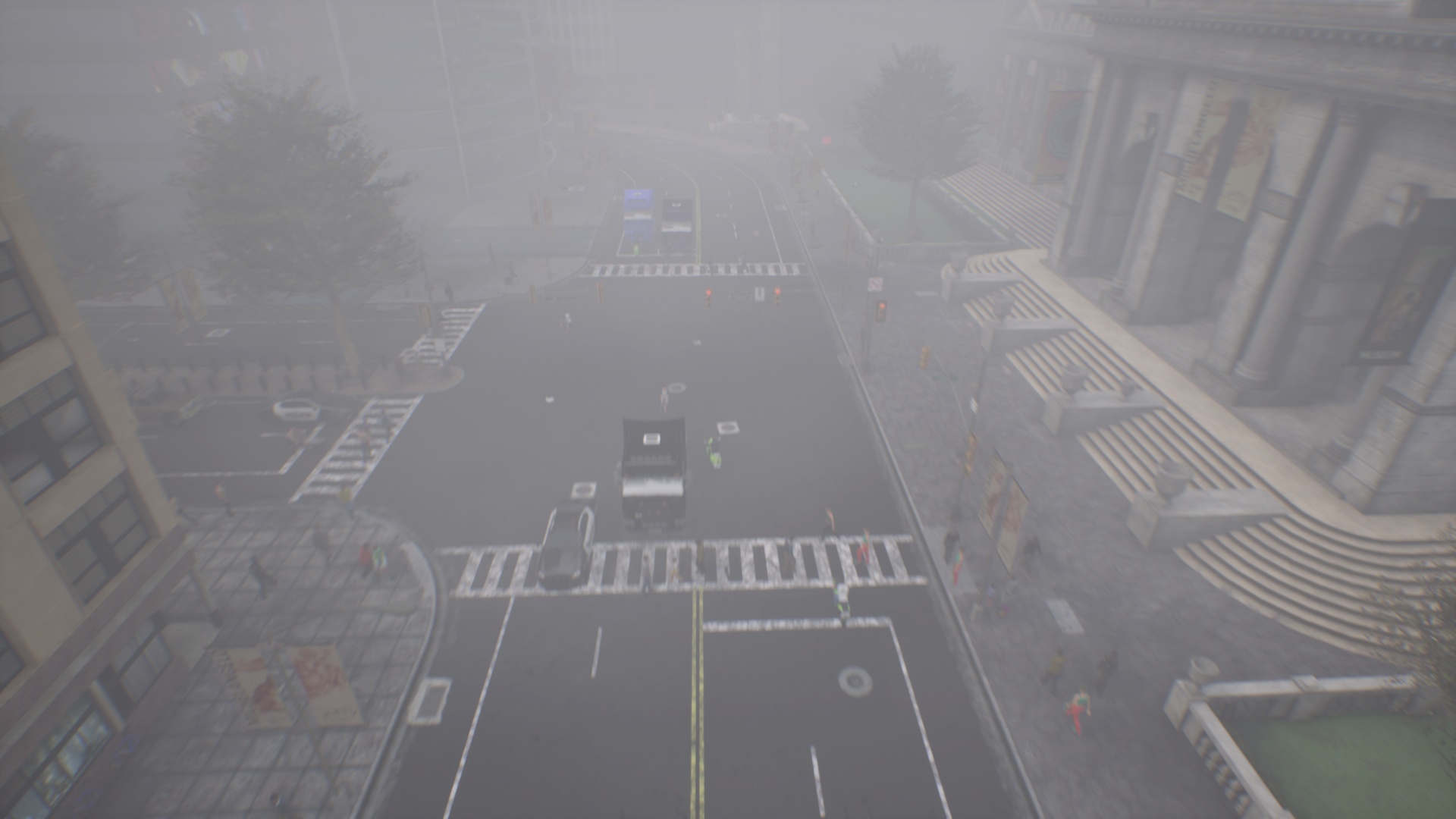}
        \end{subfigure}
    \end{subfigure}
    \begin{subfigure}{\textwidth}
        \rotatebox{90}{~~~~~\tiny{FLYAWARE-R}}
        \begin{subfigure}{.11\textwidth}
            \includegraphics[width=\textwidth]{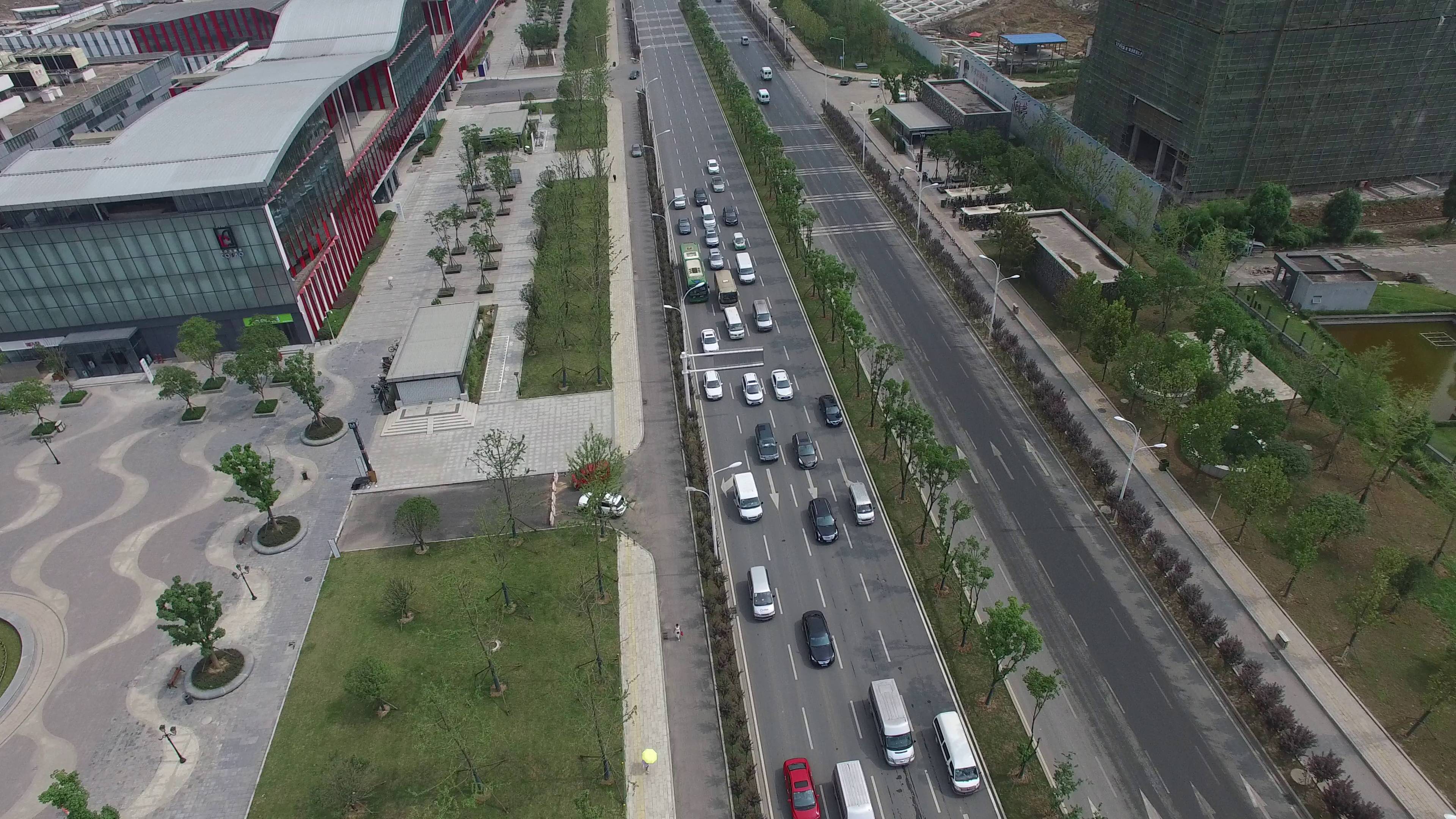}
            \caption*{Day}
        \end{subfigure}
        \begin{subfigure}{.11\textwidth}
            \includegraphics[width=\textwidth]{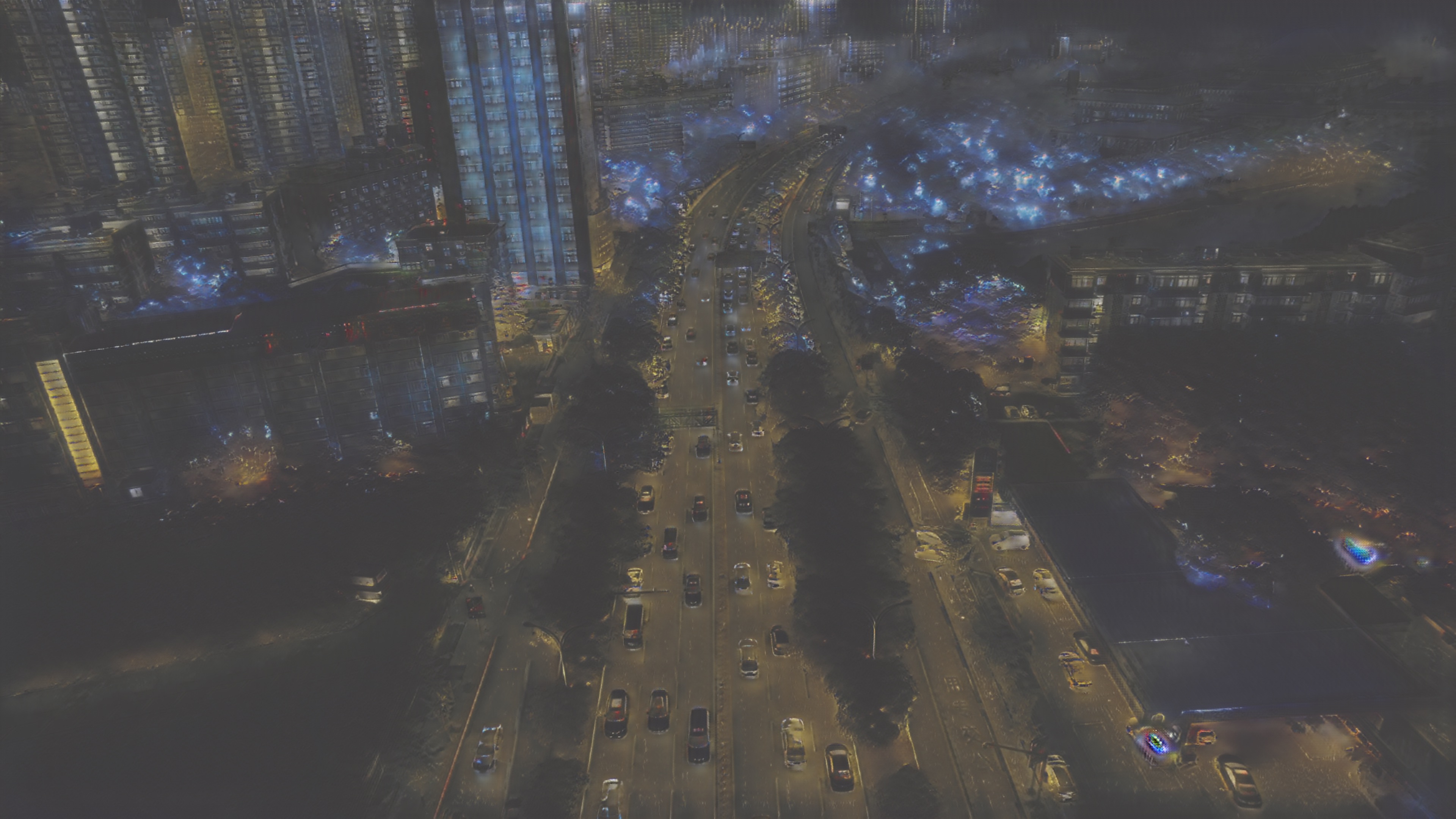}
            \caption*{Night}
        \end{subfigure}
        \begin{subfigure}{.11\textwidth}
            \includegraphics[width=\textwidth]{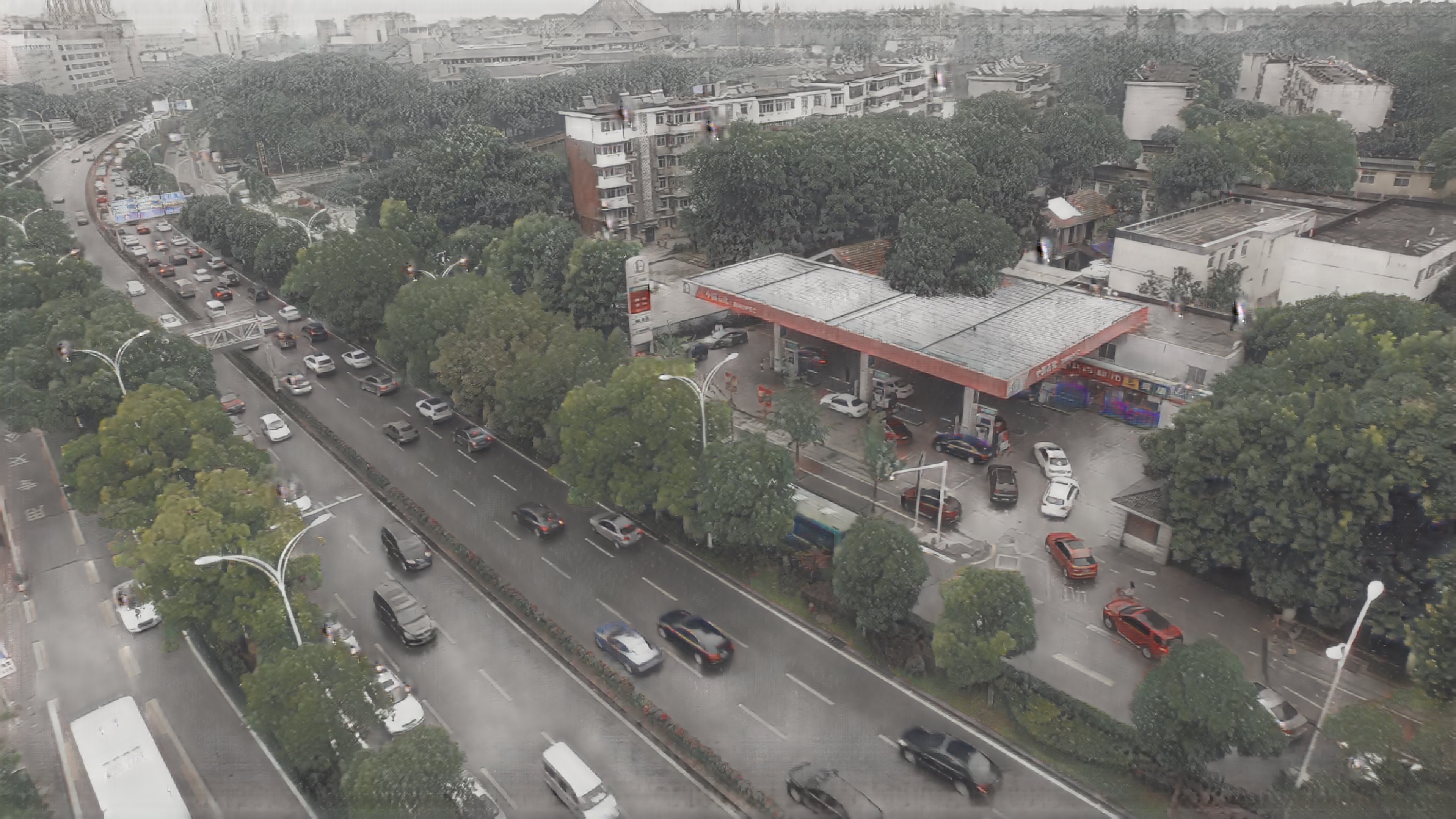}
            \caption*{Rain}
        \end{subfigure}
        \begin{subfigure}{.11\textwidth}
            \includegraphics[width=\textwidth]{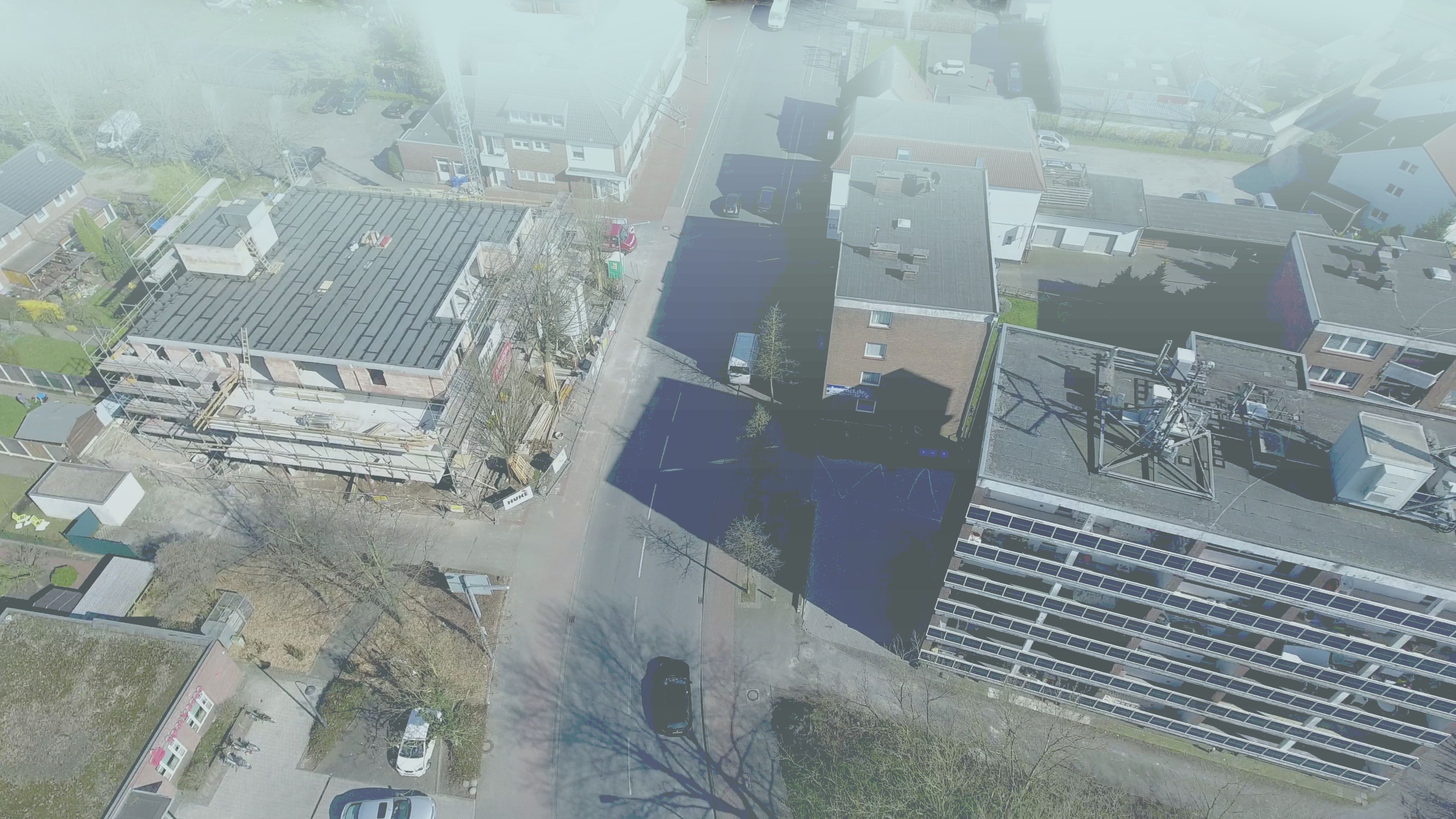}
            \caption*{Fog}
        \end{subfigure}
    \end{subfigure}
    \caption{\textbf{FLYAWARE}: samples under different weathers.}
    \label{fig:adverse}
\end{figure}

\vspace{-0.5cm}
\section{The FLYAWARE Adverse Dataset} \label{sec:dataset}

In standard autonomous driving for semantic segmentation, there is a sufficient number of datasets providing a diversity of adverse conditions \cite{testolina2023selma,sakaridis2021acdc}. However, in the case of aerial viewpoints, a scarcity of such datasets is evident, thus making the development of a dataset of this type a priority to unlock further research on the topic. 
To this aim, we introduce a new benchmark \textbf{FLYAWARE} - \textbf{FLY}ing in \textbf{A}dverse \textbf{W}eather: \textbf{A}rtificial and \textbf{RE}al data.  %
In the following section, we outline the structure of the synthetic and real aerial datasets, $\mathcal{D}_{drone}^S$ and $\mathcal{D}_{drone}^T$, respectively. They contain over 100K images for which we contribute with 108180 generated images and 1287 domain-translated ones. %
\cref{fig:adverse} shows samples from \textbf{FLYAWARE}. Please consult the \sm for details on the dataset construction. %

\begin{table*}[t]
\begin{minipage}[t]{0.66\linewidth}
    \centering
    \resizebox{\textwidth}{!}{%
    \begin{tabular}{llllccc}
        \toprule
        \textbf{Setting} & \textbf{Train} & \textbf{Test} & \textbf{Method} & \textbf{Car} & \textbf{Drone} & \textbf{All} \\
        \midrule
        centralized & SELMA+FA-S & ACDC+FA-R &Source Only  &  19.96 & 33.11  & 22.70 \\ %
        centralized & ACDC+FA-R & ACDC+FA-R & Target Only  &  60.17 &  50.14 & 58.08 \\ %
        federated & ACDC+FA-R & ACDC+FA-R &  Fine-Tuning & 34.35 & 39.10 & 35.34 \\ %
        \midrule
        \textsc{FFreeDA} & ACDC+FA-R & ACDC+FA-R & FedAvg \cite{fedavg}+ST &  25.00 & 30.25 & 26.10 \\ %
        \textsc{FFreeDA} & ACDC+FA-R & ACDC+FA-R & LADD \cite{shenaj2023learning} & 25.95 & 30.79 & 26.96 \\ %
        \textsc{FFreeDA} & ACDC+FA-R & ACDC+FA-R & \textbf{HyperFLAW} &  \textbf{27.45} & \textbf{34.69} & \textbf{28.96} \\  %
        \midrule
        \midrule
        centralized & FA-S+SELMA & ACDC+FA-R-XL &Source Only  & 19.96 & 32.61  & 22.61 \\ %
        \midrule
        \textsc{FFreeDA} & ACDC+VisDrone & ACDC+FA-R-XL & FedAvg \cite{fedavg}+ST & 22.76 & 31.55 & 24.59 \\ %
        \textsc{FFreeDA} & ACDC+VisDrone & ACDC+FA-R-XL & LADD \cite{shenaj2023learning} & 26.32 & 28.93 &  26.86 \\ %
        \textsc{FFreeDA} & ACDC+VisDrone & ACDC+FA-R-XL & \textbf{HyperFLAW} &  \textbf{27.10} & \textbf{35.45} & \textbf{28.84} \\  %
        \bottomrule
    \end{tabular}
    }
    \subcaption{mIoU on the target datasets (ACDC and FLYAWARE-R (FA-R))}
    \label{tab:finalresults}
\end{minipage}%
\begin{minipage}[t]{0.32\linewidth}
    \centering
    \vspace{-6.6em}\resizebox{0.9\columnwidth}{!}{%
    \begin{tabular}{cccc}
    \toprule
    \multicolumn{1}{c}{\multirow{2}{*}{\begin{tabular}[c]{@{}c@{}}\textbf{Pretrain}\\ on\end{tabular}}} & \multicolumn{1}{c}{\multirow{2}{*}{\begin{tabular}[c]{@{}c@{}}Source\\ Only\end{tabular}}} & \multicolumn{2}{c}{\textbf{Adapted}} \\ %
    \multicolumn{1}{c}{} & \multicolumn{1}{c}{} & Separately & Jointly \\ 
    \midrule
    Car+Drone & 22.70 & 28.95 & 28.96 \\
    Car only & 18.23 & 25.17 & 25.24 \\
    Drone only & 15.25 & 14.37 & 13.68 \\ 
    \bottomrule
    \end{tabular}
    }
    \subcaption{Viewpoint Generalization}
    \label{tab:pretraining}
    \vspace{.7em}
    \resizebox{0.9\columnwidth}{!}{%
    \begin{tabular}{ccccccc}
    \toprule
    \textbf{Pretrain} & \textbf{Adapt} & \textbf{Clear} & \textbf{Night} & \textbf{Rain} & \textbf{Fog} & \textbf{All} \\
    \midrule
    Clear & - & 23.43 & 6.43 & 24.93 & 25.50 & 18.28 \\
    All & - & 25.41 & 13.62 & 26.72 & 26.08 & 22.69 \\ \midrule
    Clear & Clear & 26.87 & 1.02 & 8.02 & 9.86 & 14.36 \\
    Clear & All & 25.97 & 7.87 & 25.70 & 23.16 & 18.11 \\
    All & Clear & 30.96 & 10.85 & 22.22 & 26.38 & 25.03 \\
    All & All & \textbf{31.73} & \textbf{14.18} & \textbf{34.19} & \textbf{29.86} & \textbf{28.96} \\
    \bottomrule
    \end{tabular}
    }
    \subcaption{Weather conditions' effects}
    \label{tab:weather}
\end{minipage}
\vspace{-.5em}
\caption{Main results of the work over the multi-view setting. \cref{tab:finalresults} reports the mIoU on multiple target datasets. \cref{tab:pretraining} shows a viewpoint generalization study where during adaptation, clients train either two separate models or jointly train a single model. \cref{tab:weather} presents a study on the generalization to different weather conditions.}
\label{tab:global}
\vspace{-.5em}
\end{table*}

\textbf{Datasets.}
\label{sec:experiments:dataset}
In \cref{tab:adverse}, we present the datasets used in this study. Our research leverages two synthetic driving datasets based on the CARLA simulator \cite{dosovitskiy2017carla}: the car-centric SELMA dataset \cite{testolina2023selma} and our newly introduced drone-centric FLYAWARE-S dataset. Both encompass a range of challenging illumination and weather conditions. Additionally, although lacking aerial counterpart, we include GTAV \cite{richter2016playing}  dataset in the \sm for comparison with previous supervised federated methods.

We then adapt our method to real-world datasets from different domains in a federated fashion. For cars, we utilize the ACDC \cite{sakaridis2021acdc} dataset, supplemented by Cityscapes \cite{cordts2016cityscapes} for clear weather samples (detailed in \sm). For drones, we created FLYAWARE-R, an adverse-weather translated dataset, using two source datasets: UAVid \cite{lyu2020uavid}, which is labeled for segmentation but limited in size, and VisDrone \cite{zhu2021detection}, which is intended for object detection. In FLYAWARE-R, we train and test on UAVid domain-translated samples. For enhanced robustness, we also created an extended set (XL) where we train on VisDrone domain-translated samples and evaluate on UAVid, allowing for more comprehensive testing.
The adverse weather translation process is described in \sm All datasets and splits are referenced in \cref{tab:adverse}. %

\textbf{Federated Scenarios.} \label{sub:dataset:fedsplit}
We examine two main federated learning settings:  
        (i) an \textbf{Unbalanced} setting with 40 clients (32 automotive, 8 drones);
        (ii) a  \textbf{Balanced} one with 64 clients (32 automotive, 32 drones), with drone training and test sets from different domains.
Following \cite{shenaj2023learning}, we sampled a heterogeneous distribution across clients: automotive clients have 69-72 samples each, while drone clients have 24-25 samples.
To simulate real-world conditions, we set clear daytime as the predominant weather, with clients experiencing various combinations of adverse conditions. For weather distribution across clients, see the \sm
Additionally, we explored a limit-case scenario: (iii) 64 balanced clients (as in (ii)), each assigned a single weather condition, to isolate weather variability effects. %

    \begin{table*}[t]
\caption{Ablation studies. \cref{tab:single_weather} compares different BN strategies in three different scenarios: (Scenario (i)) where clients are in ratio 4:1 (car:drones); (Scenario (ii)) with ratio 1:1;  (Scenario (iii)) where clients (in 1:1 ratio) have data from a single weather condition. \cref{tab:ablation} reports a more complete ablation study on all components of our strategy.}
\centering
\begin{minipage}{.3\textwidth}
    \centering
    \resizebox{\linewidth}{!}{%
    \begin{tabular}{cccccc}
    \toprule
     Method & Aggr. & W-BNs & \textbf{(i)} & \textbf{(ii)} &\textbf{(iii)} \\
     \midrule
     SiloBN \cite{andreux2020siloed} &  &  & 25.09 & 24.50 & 24.93 \\
     FedBN \cite{li2021fedbn} &  &  & 25.20  & 24.70 & 25.44 \\
     Ours & \checkmark &  & 28.51 & 27.07 & 25.49 \\  
     Ours & \checkmark & \checkmark & \textbf{28.96} & \textbf{28.84} & \textbf{28.04} \\
     \bottomrule
    \end{tabular}
    }
    \subcaption{Batch normalization strategies.}
    \label{tab:single_weather}
    \centering
    \resizebox{\linewidth}{!}{%
      \begin{tabular}{ccccccccc}
          \toprule
          $\mathcal{L}_{st}$ & \multicolumn{1}{c}{\textbf{AG}} & & $\mathcal{L}_{cl}$ &\multicolumn{1}{c}{\textbf{BN}}  & \multicolumn{1}{c}{\textbf{Car}} & \multicolumn{1}{c}{\textbf{Drone}} & \multicolumn{1}{c}{\textbf{All}} \\
          \midrule
          &  &  &  & & 19.96 & 33.11 & 22.70 \\ %
          \checkmark &  &  &  & & 25.00 & 30.25 & 26.10 \\ %
           \checkmark&  \checkmark &  &  & & 24.75 &  33.49 &  26.57 \\ %
          \checkmark &  \checkmark &  &  & \checkmark  & 24.91 &  34.55 &  26.92 \\ %
           \checkmark &  \checkmark &  & \checkmark  &  & 27.17 & 33.60 &  28.51 \\ %
           \checkmark &  \checkmark & & \checkmark &  \checkmark & \textbf{27.45}&  \textbf{34.69} & \textbf{28.96} \\ %
           \bottomrule
      \end{tabular}
      }
    \subcaption{Ablation on components.}
    \label{tab:ablation}
\end{minipage}\hspace{1em}
\begin{minipage}{.32\textwidth}
    \centering
    \includegraphics[width=\textwidth,clip,trim={0 0 0 10pt}]{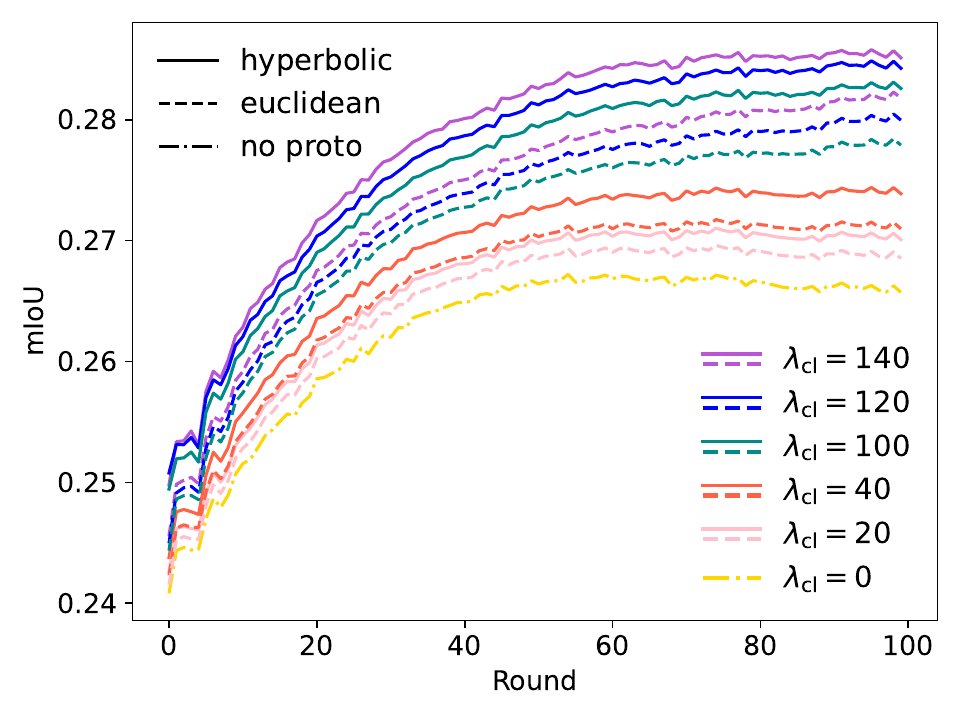}
    \captionof{figure}{Euclidean (dashed lines) vs hyperbolic (solid lines) prototypes.} %
    \label{fig:ablation_protovshyper}
\end{minipage}\hspace{1em}
\begin{minipage}{.32\textwidth}
    \centering
      \includegraphics[width=\linewidth,clip,trim={0 0 0 10pt}]{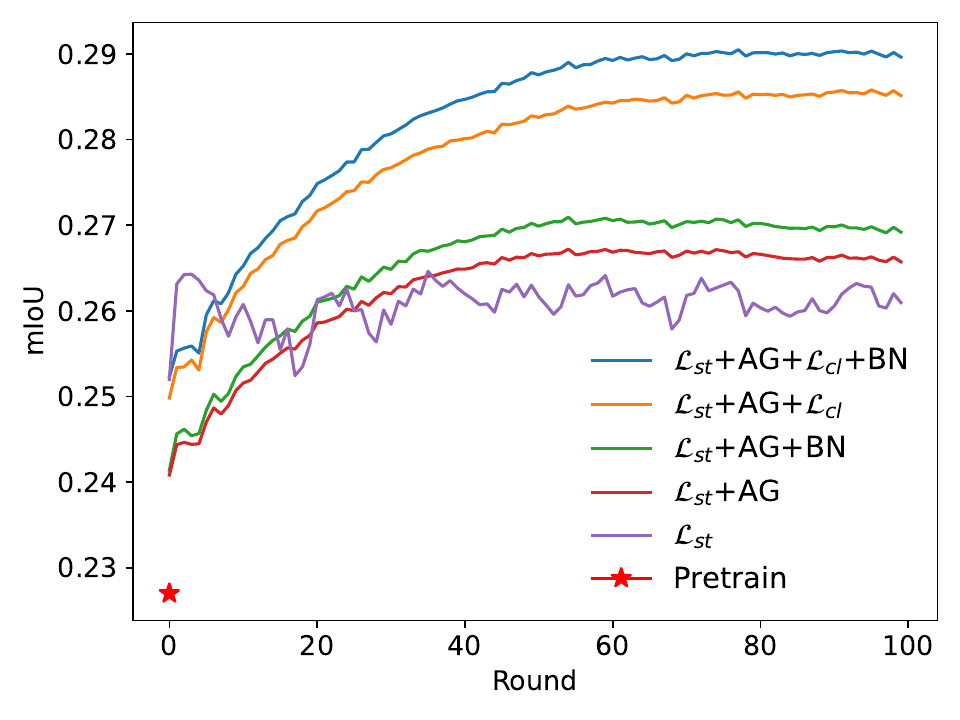}
      \captionof{figure}{Training curves of the ablated method.}
      \label{fig:ablation}
\end{minipage}
\end{table*}

\section{Experimental Results}
\label{sec:results}
The results are detailed in \cref{tab:finalresults}, which reports the average mIoU for cars, drones and a combined score. The latter is computed using the average for the classes present in both datasets and the car mIoU otherwise. See the \sm for implementation and network architecture details.
For comparison, we initially examined three supervised baselines: 1) the source-only approach, which relies exclusively on synthetic labeled data from the source domains for training; 2) the fully centralized approach, where all available data is used to train the network specifically on the target domain; 3) the upper limit scenario for the federated model, where it undergoes fine-tuning with full supervision on the clients.
In the unsupervised setting, (i.e.,  in the context of \textsc{FFreeDA}), we explore two strategies: the standard FedAvg \cite{fedavg} combined with our self-training approach, and the main competitor, LADD \cite{shenaj2023learning}. Notably, unlike our strategy, LADD conducts dataset stylization from the target domain, assuming prior knowledge of the client data distribution during pretraining. We pretrain LADD on our  synthetic adverse weather dataset to ensure comparability, maintaining the target's average style as outlined in \cite{shenaj2023learning}.
The significant domain shift between source and target domains can be observed from the mIoU of 22.70\% achieved by source-only training compared to 58.08\% of the target training. Although fine-tuning the source-only pretraining increases performance by almost 13\%, assuming labeled data on the client side is unrealistic in many scenarios.
Our approach can successfully align the intrinsic distribution shift that different viewpoints introduce, in addition to outperforming other methods. The \sm demonstrates this result holds even when no aerial data are provided in the GTA $\rightarrow$  Cityscapes scenario.

\noindent\textbf{Multi-viewpoint generalization capabilities.}
\cref{tab:pretraining} shows that multi-viewpoint pretraining enhances overall performance compared to pretraining on either car or drone data alone. Specifically, jointly pretraining on both cars and drones improves the overall mIoU by 4.47\% compared to car-only and 7.45\% to drone-only. Importantly, we demonstrate that clients of the other agent type are not necessary for training. Adapting to cars and drones together leads to roughly the same performance as adapting two separate models  - one for cars and one for drones -, indicating good generalization across the two agents even when only trained with data from a single domain.%

\noindent\textbf{Impact of Weather Conditions and Normalization.} 
In \cref{tab:weather}, we present experiments training on the clear-only weather subset to verify generalization abilities across weather conditions. When performing supervised pretraining and unsupervised adaptation (i.e., our full method) on clear-only data and then testing on adverse conditions, the model loses over 14 mIoU points, motivating further exploration of this setting.
To understand the impact of batch-normalization - our main component to address climate heterogeneity - we also tested a limit-case scenario where each client has data of only one weather condition (\cref{tab:single_weather}, Scenario (iii)). In this context, the effect of our weather-specific BNs is enhanced, %
demonstrating that sharing weather-personalized features benefits in learning a global model. Note that not aggregating BN layers, as in SiloBN \cite{andreux2020siloed} and FedBN \cite{li2021fedbn}, worsens the performance over the baseline (i.e., aggregating over non-weather specific BNs), proving that these layers carry weather-specific style content.  Further per-weather results are discussed in the \sm

\noindent\textbf{Impact of prototypes in the Hyperbolic space.}
The key strength of the hyperbolic space lies in its ability to effectively represent distances between feature vectors, a crucial factor in tasks like clustering where having a distance function capturing meaningful relationships among data points is essential.
In Fig. \ref{fig:ablation_protovshyper}, we show the training curves for both Euclidean and Hyperbolic prototypes employing different values of the weighting parameter for the prototype loss $\lambda_{cl}$. Hyperbolic space prototypes consistently allow for increased performance across different $\lambda_{cl}$ values.\\
\noindent\textbf{Ablation studies.} %
\cref{tab:ablation} shows an extensive ablation of the various modules of the proposed approach together and \cref{fig:ablation} the corresponding training curves. 
The aggregation strategy of Eq. (\ref{eq:ssagg}) stabilizes the self-training and avoids performance decrease on unbalanced clients (\ie drones). 
The batch normalization effect is visible in cars and drones ($+3.4\%$ overall). Observe that the testing set of drones is more balanced in terms of weather categories (\textonequarter \, each), making it more noticeable. 
The hyperbolic prototypes allow for improving the generalization of the feature spaces compared to the baseline, %
achieving $28.51\%$. Finally, adding each component leads to the final score of $28.96\%$. Overall, the contributions of all the components jointly produce a gain of over $6.26\%$ of mIoU on the source only. Further ablation on the hyperparameters is in the \sm

    \section{Conclusion} \label{sec:conclusion}
In this paper we introduced a new federated learning scenario, addressing the challenges of heterogeneous autonomous agents in adverse weather conditions.
The contributions include a realistic federated setting where clients are unsupervised and a versatile model seamlessly integrating aerial views. We also introduced the FLYAWARE dataset for semantic segmentation in adverse weather.
The proposed method, HyperFLAW, employs weather-aware batch normalization layers to mitigate domain shifts across diverse weather conditions. Additionally, it leverages prototype-based learning in the hyperbolic space, contributing a novel method for consistent training in this challenging federated setting. %
By extending the scope to different types of vehicles and addressing real-world challenges such as adverse weather and privacy preservation, our contributions lay a foundation for future advancements in autonomous driving.
Further research will address the exploitation of multi-modal data and explore more advanced adaptation strategies for adverse weather conditions, as well as bridging the significant domain gap between car and aerial images.

\textbf{Acknowledgment.~} This work was partially supported by the European Union under the Italian National Recovery and Resilience Plan (NRRP) of NextGenerationEU, partnership on ``Telecommunications of the Future'' (PE00000001- program ``RESTART'').

    {\small
    \bibliographystyle{ieee_fullname}
    \bibliography{main}
    }

    \clearpage

    \renewcommand{\thefigure}{S\arabic{figure}}
    \renewcommand{\theHfigure}{S\arabic{figure}}
    
    \renewcommand{\thesection}{S\arabic{section}}
    \renewcommand{\theHsection}{S\arabic{section}}
    
    \renewcommand{\theequation}{S\arabic{equation}}
    \renewcommand{\theHequation}{S\arabic{equation}}
    
    \renewcommand{\thetable}{S\arabic{table}}
    \renewcommand{\theHtable}{S\arabic{table}}
    
    \setcounter{equation}{0}
    \setcounter{figure}{0}
    \setcounter{table}{0}
    \setcounter{section}{0}
    \setcounter{page}{1}
    
    \clearpage

    \begin{strip}
        \null
        \vskip .375in
        
        \begin{center}
            {\large \bf 
             When Cars meet Drones: Hyperbolic Federated Learning for Source-Free Domain Adaptation in Adverse Weather \\ \textit{Supplementary Material}
            \par}
            \vspace*{12pt}
        \end{center}
    \end{strip}

    \definecolor{road}{rgb}{.502,.251,.502}
\definecolor{sidewalk}{rgb}{.957,.137,.910}
\definecolor{building}{rgb}{.275,.275,.275}
\definecolor{wall}{rgb}{.4,.4,.612}
\definecolor{fence}{rgb}{.745,.6,.6}
\definecolor{pole}{rgb}{.6,.6,.6}
\definecolor{tlight}{rgb}{.980,.667,.118}
\definecolor{tsign}{rgb}{.863,.863,0}
\definecolor{vegetation}{rgb}{.420,.557,.137}
\definecolor{terrain}{rgb}{.596,.984,.596}
\definecolor{sky}{rgb}{0,.510,.706}
\definecolor{person}{rgb}{.863,.078,.235}
\definecolor{rider}{rgb}{1,0,0}
\definecolor{car}{rgb}{0,0,.557}
\definecolor{truck}{rgb}{0,0,.275}
\definecolor{bus}{rgb}{0,.235,.392}
\definecolor{train}{rgb}{0,.314,.392}
\definecolor{motorbike}{rgb}{0,0,.902}
\definecolor{bicycle}{rgb}{.467,.043,.125}
\definecolor{unlabelled}{rgb}{0,0,0}

This document contains the supplementary material for the paper \textit{When Cars meet Drones: Hyperbolic Federated Learning for Source-Free Domain Adaptation in Adverse Weather}. 
We begin by presenting the implementation specifics of our network.  
Next, we introduce additional details on the employed datasets. We then proceed with ablations supporting the usage of the weather batch-normalization and hyperbolic prototypes. %
Finally, we show further experimental data including per-class accuracy scores and qualitative results. 

\section{Implementation Details}  \label{sub:experiments:implementation}
\noindent\textbf{Server pretraining.}
We chose DeepLabV3 architecture with MobileNetV2 as in \citesupp{shenaj2023learning}.
We trained our model on the source synthetic dataset using a decreasing power-law learning rate $\eta$, starting at $\eta = 5 \times 10^{-3}$ with a power of 0.9. The optimization used SGD with momentum 0.9 and no weight decay, lasting for 5 epochs with batch size 16.
For the weather classifier, we implemented a 3-layer ConvNet (see \cref{fig:classifier}), trained for 8 epochs using SGD optimization with a learning rate of $1 \times 10^{-4}$ and a batch size of 88.\\
\noindent\textbf{Clients Adaptation.}
For the target dataset, experiments were run with a fixed learning rate of $\eta = 1 \times 10^{-4}$ with SGD optimizer. The training involved 5 clients per round for a total of $R = 100$ rounds, with $\lambda_{cl}=140$. The pseudo-label teacher model was updated at the end of each round.
For the optimizer of the manifolds, we use RiemmanianAdam as in \citesupp{van2023poincar} with $\gamma = 0.1$ as initialization, learning rate equal to $1 \times 10^{-3}$ and weight decay of $4 \times 10^{-4}$.\\
\noindent\textbf{Training of Competitors\quad} %
To establish performance upper bounds, we conducted two experiments:
\begin{enumerate}
    \itemsep0em
    \item Fully Centralized Training: We performed supervised training on the target dataset for 250 epochs with a learning rate of $5 \times 10^{-3}$.
    \item Federated Fine-Tuning: We implemented federated fine-tuning with full supervision on the clients for 250 rounds, representing an upper limit scenario for the federated model.
\end{enumerate}

\section{Data Selection and Distribution}
In standard autonomous driving for semantic segmentation, a sufficient number of datasets provide a diversity of adverse weather conditions. For our study on car agents, we relied on existing datasets, adapting them to enable distributed learning in adverse weather scenarios. Here, we will first describe the technical choices adopted for this purpose.
Conversely, there is a scarcity of datasets featuring adverse conditions for aerial viewpoints. To address this gap, we introduced the \textbf{FLYAWARE} aerial dataset, as it represents a novel contribution to unlock further research on this topic.
Lastly, we provide additional details on weather distribution among clients.

\begin{figure*}[t]
    \centering
    \begin{subfigure}{.30\textwidth}
        \includegraphics[width=\textwidth]{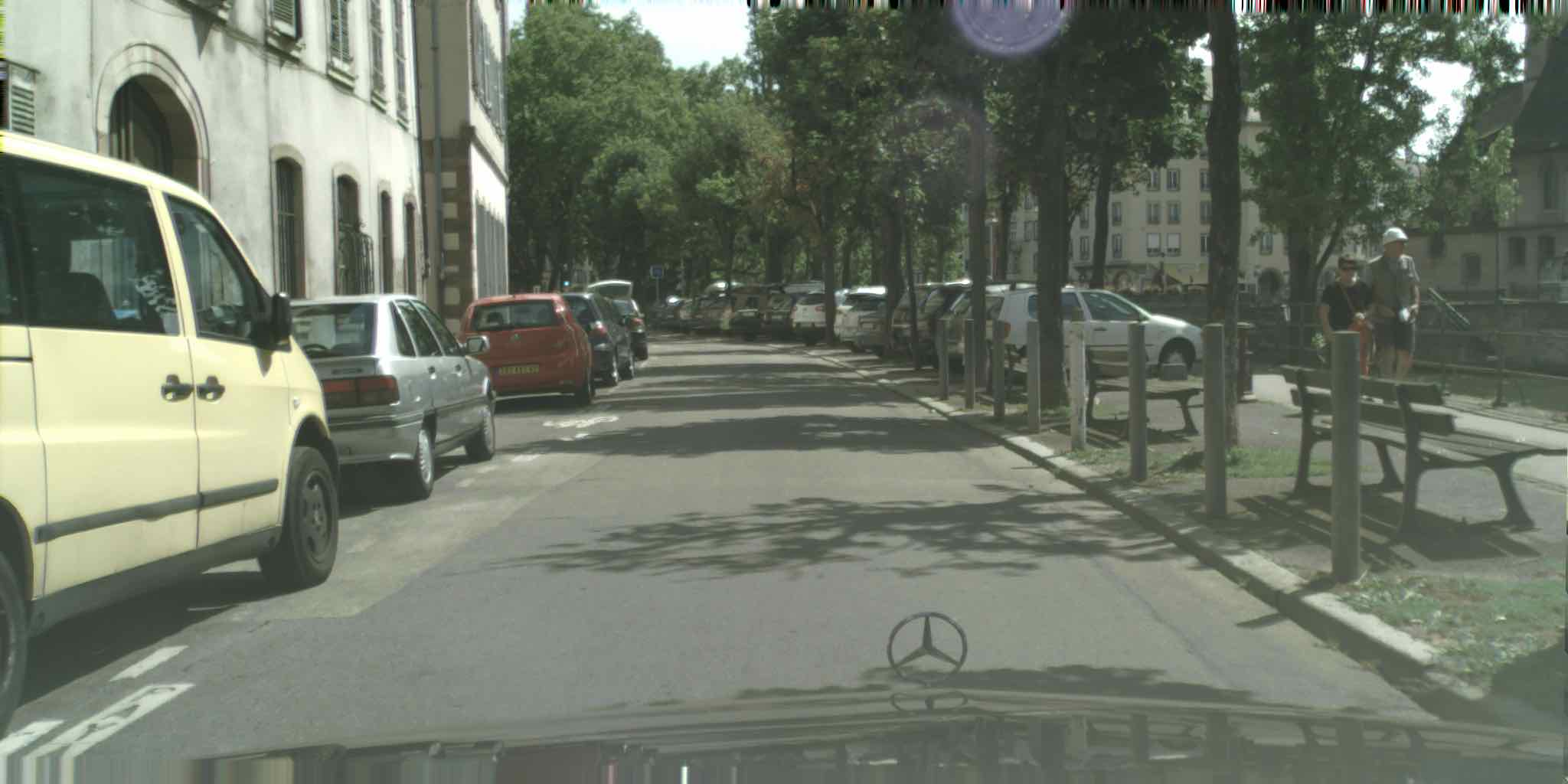}
    \end{subfigure}
    \begin{subfigure}{.30\textwidth}
        \includegraphics[width=\textwidth]{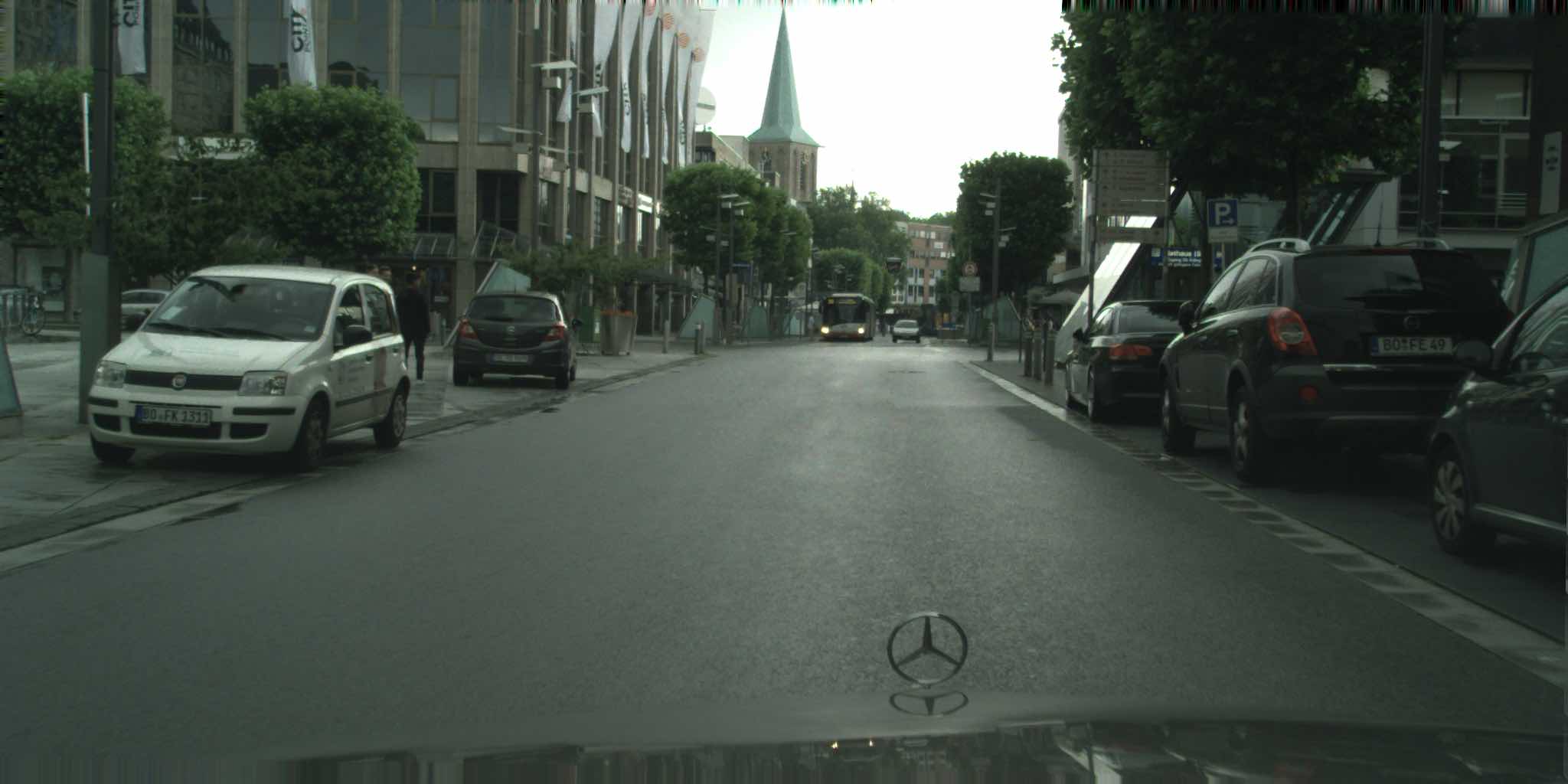}
    \end{subfigure}
    \begin{subfigure}{.30\textwidth}
        \includegraphics[width=\textwidth]{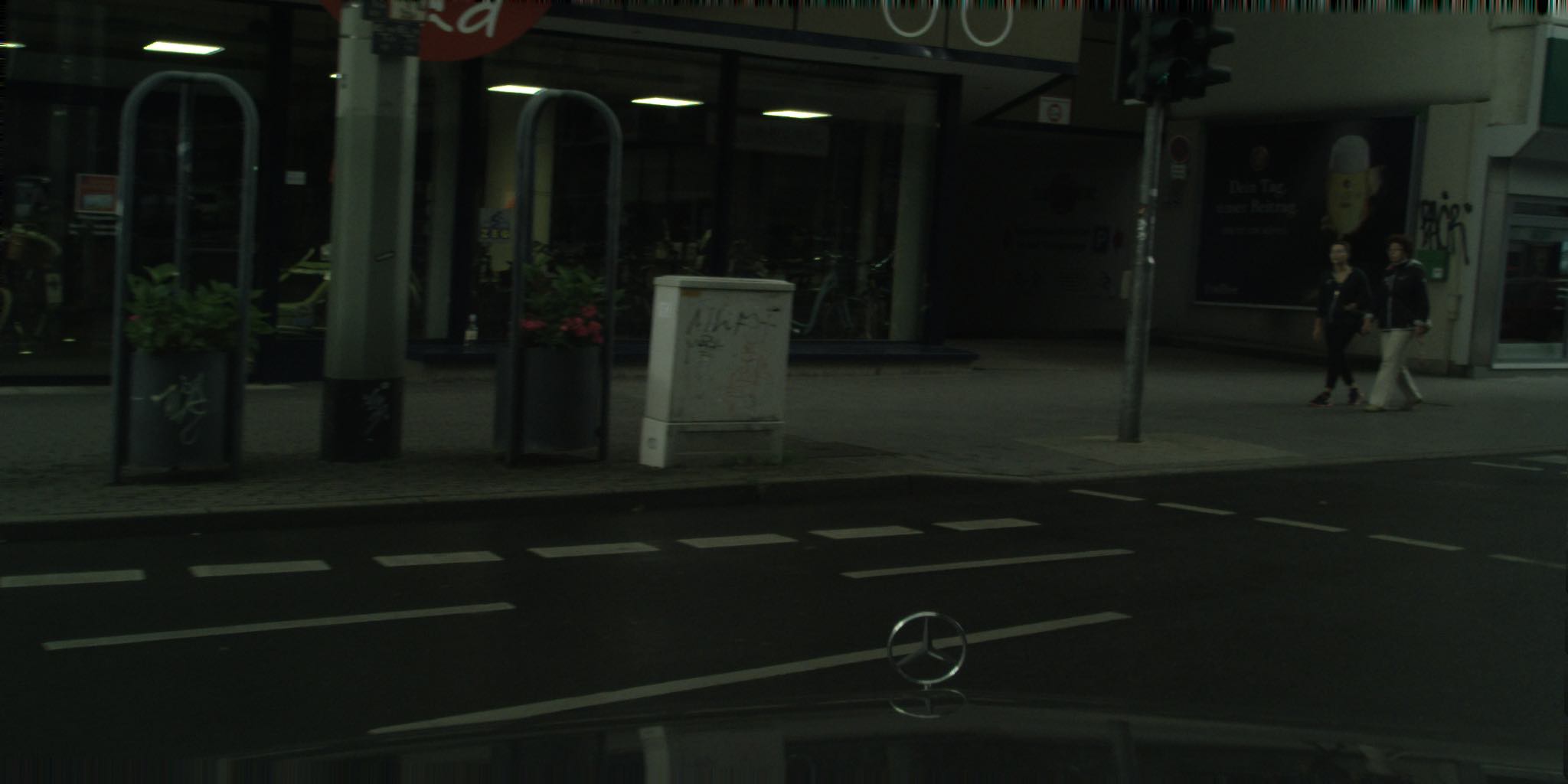}
    \end{subfigure}
    \caption{Cityscapes samples and corresponding \textit{sunlit} score. Decreasing from the left, high/mid/low-score ($\sim300/200/100$).}
    \label{fig:city_sunny}
\end{figure*}
\begin{figure*}[t]
\centering
    \begin{subfigure}{.32\textwidth}
        \centering
        \includegraphics[width=\linewidth,trim={0 0.5cm 0 0.4cm },clip]{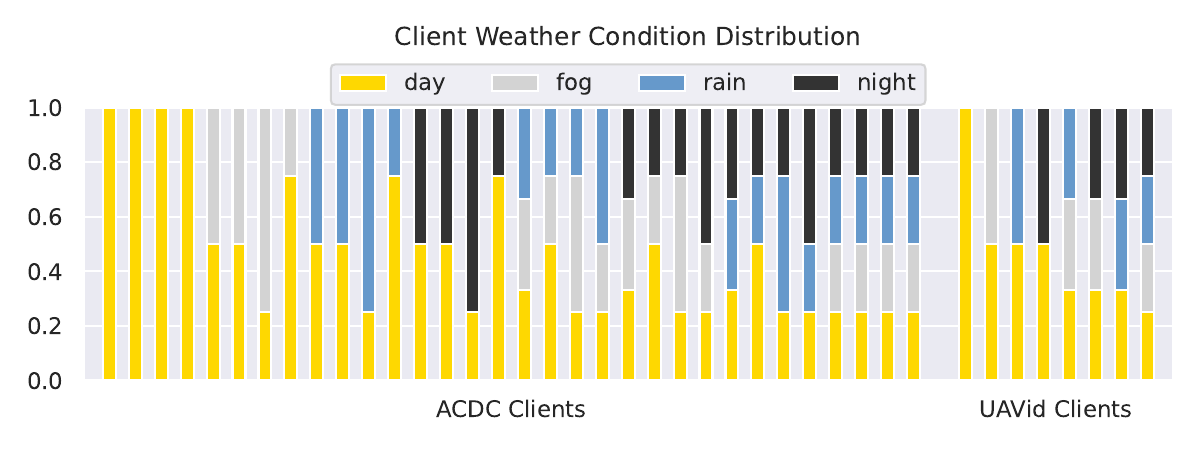}
        \caption{Scenario (i)}
        \label{fig:client_per_weather}
    \end{subfigure}
    \begin{subfigure}{.32\textwidth}
        \centering
        \includegraphics[width=\linewidth,trim={0 0.5cm 0 0.4cm },clip]{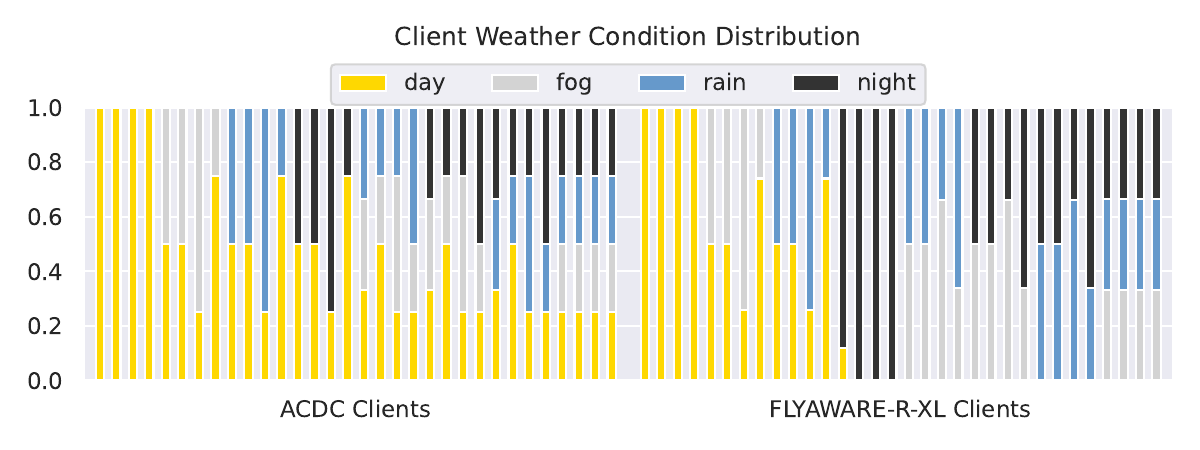}
        \caption{Scenario (ii)}
        \label{fig:client_per_weather_xl}
    \end{subfigure}
    \begin{subfigure}{.32\textwidth}
        \centering
        \includegraphics[width=\linewidth,trim={0 0.5cm 0 0.4cm },clip]{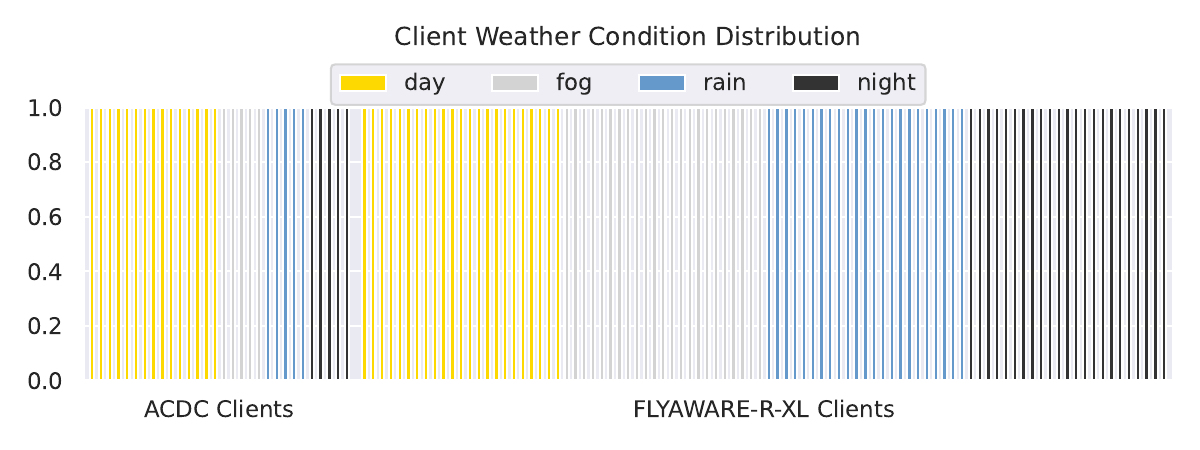}
        \caption{Scenario (iii)}
        \label{fig:client_per_weather_single}
    \end{subfigure}
\caption{Distribution of weather conditions across clients}
\end{figure*}

\subsection{Driving Datasets}
\label{sub:dataset:car}
For the synthetic source dataset for cars, we used the \textbf{SELMA} dataset \citesupp{testolina2023selma}. 
It offers a comprehensive set of 27 weather and daytime conditions, resulting in a vast dataset of over 20M samples. To better align with the weather scenarios considered in real data \citesupp{sakaridis2021acdc}, we opted not to use the standard SELMA split, but downloaded over 24k samples in the \textit{ClearNoon, ClearNight, HardRainNoon, MidFoggyNoon} splits.  
In total, for this work, we employed almost 100k SELMA samples from the Desk Cam point of view that match the one used in the real-world dataset.
As the real counterpart, we used the \textbf{ACDC} dataset, from which we selected the 3 domains — night, rain, and fog — that match our pretraining. As ACDC lacks images in clear weather, we supplemented it with daytime images coming from the Cityscapes dataset \citesupp{cordts2016cityscapes}, which is the most similar to ACDC. Note that, the  ACDC dataset has been built to create an adverse condition version of Cityscapes, and shares the same class-set. 
Cityscapes provides more samples than those in the thematic splits by ACDC. Therefore, we subsampled its training set to match the sizes. Since the goal was to select clear weather conditions, we devised an automatic way of assigning a sunlit level (Eq. \ref{eq:city}) to each image and selected the needed images by sorting them according to this metric:
\begin{equation}
    \label{eq:city}
    \text{sunlit}(\textbf{X}) = \sum\limits_{c \in \{r,g,b\}}\frac{1}{H W}\sum\limits_{i=0}^{W}\sum\limits_{i=0}^{W}\textbf{X}[i,j,c]
\end{equation}
As shown in \cref{tab:adverse} %
in the main paper, we selected a set of images to match the total count of all other conditions combined, aiming for a ratio of three clear sky images to one adverse condition. This decision was driven by the desire to retain ``clear sky'' as the most probable scenario in real-world contexts.
In \cref{fig:city_sunny} we report some examples of samples with their rating (a bright one, a dark one and a mid-range one).

\subsection{Aerial Dataset}
A demo video of the dataset is available at \url{https://github.com/LTTM/HyperFLAW}.

\noindent\textbf{FLYAWARE-S: Adverse Synthetic Dataset.} \label{sub:dataset:advsyndrone}
Recently released, Syndrone \citesupp{rizzoli2023syndrone} is a synthetic dataset based on the CARLA simulator \citesupp{dosovitskiy2017carla}.
While the dataset is richly annotated with 28 semantic classes, it currently lacks imagery in different weather conditions.
We extended their work starting from the codebase provided in \citesupp{rizzoli2023syndrone} by generating images in 3 adverse weather conditions (\ie, rain, fog, and night), while maintaining the capability of the system to produce images from multiple viewpoints and heights. The dataset includes drone views at heights ranging from 20 to 80 meters and with angles varying from 30 to 90 degrees, all with a resolution of 1920x1080. Moreover, we also generated  3D data (depth maps and LiDAR) for future usage in multimodal architectures.\\

\begin{table*}[t]
    \resizebox{\textwidth}{!}{%
    \begin{tabular}{ccccc|ccccccccccccccccccc|c}
    \toprule
    \rotatebox{90}{\textbf{ST}} & \rotatebox{90}{\textbf{AG}} &  & \rotatebox{90}{\textbf{CL}} & \rotatebox{90}{\textbf{BN}} & \rotatebox{90}{road} & \rotatebox{90}{sidewalk} & \rotatebox{90}{building} & \rotatebox{90}{wall} & \rotatebox{90}{fence} & \rotatebox{90}{pole} & \rotatebox{90}{traffic light} & \rotatebox{90}{traffic sign} & \rotatebox{90}{vegetation} & \rotatebox{90}{terrain} & \rotatebox{90}{sky} & \rotatebox{90}{person} & \rotatebox{90}{rider} & \rotatebox{90}{car} & \rotatebox{90}{truck} & \rotatebox{90}{bus} & \rotatebox{90}{train} & \rotatebox{90}{motorcycle} & \rotatebox{90}{bicycle} & \rotatebox{90}{All} \\ \midrule
     & & & & & 35.4 & 5.7 & 55.8 & 4.6 & 4.2 & 14.2 & 13.3 & 10.9 & 61.9 & 6.3 & 43.7 & 26.0 & 17.2 & 11.1 & 3.3 & 3.8 & 1.4 & 8.9 & 10.7 & 22.7 \\
    \checkmark & &  &  &  &   43.7 & 22.0 & 57.2 & 6.6 & 4.7 & 15.8 & 9.8 & 13.9 & 50.2 & 5.7 & 68.4 & 24.3 & 11.4 & 34.6 & 8.9 & 4.7 & 3.5 & 16.0 & 14.9 & 26.1 \\ %
    \checkmark & \checkmark &  &  &  & 43.9 & 23.2 & 59.2 & 7.1 & 5.9 & 16.2 & 11.1 & 14.8 & 55.3 & 6.1 & 58.5 & 26.0 & 11.9 & 34.2 & 9.2 & 4.3 & 3.0 & 15.8 & 13.5 & 26.6\\ %
    \checkmark & \checkmark &  &  & \checkmark & 43.8 & 23.6 & 59.7 & 6.9 & 5.9 & 15.9 & 10.7 & 14.8 & 56.5 & 5.1 & 60.6 & 25.4 & 11.9 & 37.7 & 9.0 & 4.5 & 2.8 & 15.1 & 13.0 &  26.9   \\  %
    \checkmark & \checkmark &  & \checkmark & & 46.8 & 22.4 & 59.9 & 7.4 & 6.7 & 17.6 & 17.0 & 22.4 & 56.6 & 7.1 & 62.6 & 25.4 & 10.5 & 35.5 & 9.5 & 6.9 & 3.7 & 16.6 & 25.7 & 28.5 \\ %
    \checkmark & \checkmark & & \checkmark & \checkmark & 47.2 & 22.7 & 60.6 & 7.0 & 6.7 & 17.3 & 16.8 & 22.7 & 57.7 & 5.9 & 64.9 & 25.4 & 10.6 & 36.9 & 9.3 & 6.8 & 3.5 & 16.8 & 25.5 & 29.0 \\ 
    \bottomrule
    \end{tabular}}
    \caption{Per class IoU for the proposed approach and its ablated versions. %
    }
    \label{tab:per_class_iou}
\end{table*}

\noindent\textbf{FLYAWARE-R: Real Dataset Translation.} \label{sub:dataset:advuavid} Since no adverse weather dataset for aerial vehicles is available we opted for using image translation over standard drone datasets to build the adverse weather imagery.
Addressing the task of converting daytime images into adverse conditions requires the availability of images %
of clear weather samples like sunny or cloudy conditions, and %
of adverse samples occurring in specific conditions of interest, such as rainy weather.
Inspired by \citesupp{gasperini2023robust} which performs adverse domain translations for autonomous driving in the context of depth estimation, we opted for the use of \acp{gan}. %
For the rain and night adverse conditions, we employed the ForkGAN model \citesupp{zheng2020forkgan} %
to translate clear-day training samples from the UAVid dataset \citesupp{lyu2020uavid}. While for the fog samples, we 
have used a combination of Omnidata \citesupp{eftekhar2021omnidata} and FoHIS \citesupp{zhang2017towards} methods. In the following, we will present an in-depth explanation of how we have performed the aforementioned conversion.

\noindent\textbf{Day2Night.} To convert daylight into nighttime, we trained ForkGAN over $14K$ images, half of which represent daytime and the other half nighttime. All the used images were sampled from Visdrone \citesupp{zhu2021detection}, and UAVDT \citesupp{du2018unmanned}, two datasets designed for Object Detection. %
We have decided to sample the training data from different datasets to increase the overall number of nighttime images. This was necessary to ensure that the \ac{gan} architecture reaches good enough reconstruction performance. The training phase lasted $40$ epochs. After the training, for all the nighttime samples in the UAVid dataset, we %
convert clear-day images into their nighttime counterparts using the pretrained model.

\noindent\textbf{Day2Rain.} Training the \ac{gan} models directly on drone data proved advantageous as it helped to mitigate domain shifts. In the rain case however there are no available drone datasets  - not even for other tasks as in the case of night images -,
therefore we had to train  ForkGAN using diverse datasets that included adverse weather scenarios, like BDD100K \citesupp{yu2020bdd100k}, ACDC \citesupp{sakaridis2021acdc} and RainCityscapes \citesupp{hu2019depth} (which are datasets for automotive applications). 
For  daytime to rainy conversion, we followed the same strategy applied for the conversion to nighttime training the ForkGAN model over $9K$ images for $40$ epochs and  we applied %
the pretrained model to convert  images  into their rainy counterparts.

\noindent\textbf{Day2Fog.} To perform daytime to foggy conversion we %
first estimated a depth map for each image in the UAVid dataset %
using Omnidata \citesupp{eftekhar2021omnidata}, then, we exploited the estimated depth %
as input to the FoHIS method \citesupp{zhang2017towards} to apply fog.

\subsection{Weather Heterogeneity}
\textbf{Scenario i}: \cref{fig:client_per_weather} illustrates the weather distribution among clients in the configuration \textbf{ACDC+} \textbf{FLYAWARE-R}. We deliberately supplied clear day samples to each client to emulate a common scenario where clear daylight conditions prevail, with fewer instances of challenging weather conditions. This introduces an inherent challenge as adverse weather data samples are less prevalent. Additionally, the imbalance among clients further complicates the setting.

\begin{figure}[t]
    \centering
    \includegraphics[width=0.4\textwidth]{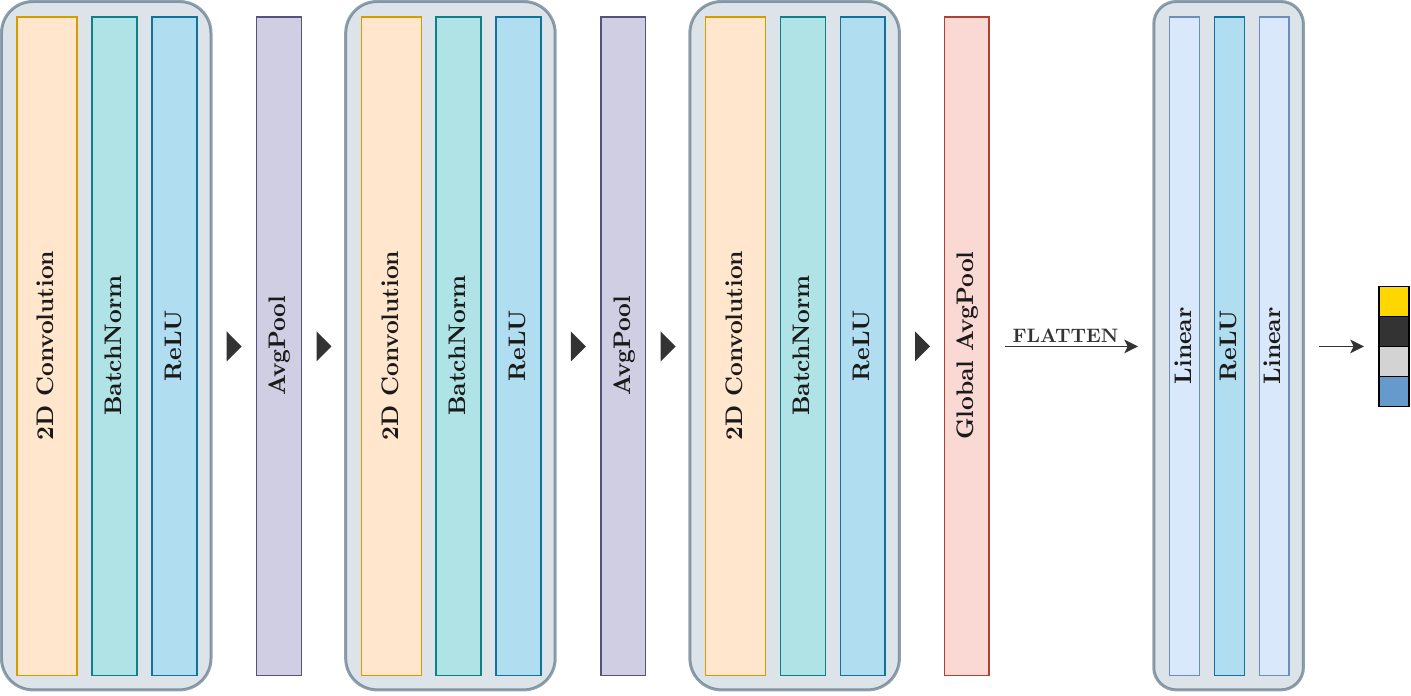}
    \caption{Weather classifier architecture model.}
    \label{fig:classifier}
\end{figure}

\textbf{Scenario ii}: \cref{fig:client_per_weather_xl} shows the weather distribution for clients in the \textbf{ACDC+FLYAWARE-R-XL} setup. While this configuration exhibits better data balance across clients, it introduces new complexities: some drone clients exclusively operate in adverse conditions, and the training and testing domains originate from different datasets (VisDrone \citesupp{zhu2021detection} and UAVid \citesupp{lyu2020uavid}, respectively). Notably, VisDrone images offer real night data, whereas we utilized domain-translated images for fog and rain using the same strategies.

\textbf{Scenario iii}: \cref{fig:client_per_weather_single} shows the weather distribution for clients in the \textbf{ACDC+FLYAWARE-R-XL} setup in the extreme case where each client observes a single weather condition. Although the number of clients is balanced in terms of different viewpoints, operating in a single weather condition implies less training variability.

\section{Impact of Weather-Batch Normalization}
As shown in \cref{tab:single_weather} %
of the main paper, the inclusion of ad-hoc batch norms (BNs) enhances performance in comparison to utilizing a non-personalized network.
First of all, in \cref{fig:classifier}, we show in detail the architecture of the weather classifier.
Although the classifier model is simple, it achieves an accuracy of $98.96\%$ on the source synthetic datasets used to train it and of $72.37\%$ on the real world target ones.

\begin{table}[t]
    \centering
    \begin{tabular}{lcccc}
        \toprule
         \textbf{Modules} & \textbf{Clear} & \textbf{Night} & \textbf{Rain} & \textbf{Fog} \\
         \midrule
        Pretrain & 25.41 & 13.62 & 26.72 & 26.08 \\ %
        $\mathcal{L}_{st}$ + AG w/o BN & 29.27 &  13.47 & 29.47 & 27.86 \\ %
        $\mathcal{L}_{st}$ + AG w BN & 26.81 & 13.50& 30.19 & 28.86  \\ %
        ALL w/o BN & 31.58 & 14.04 & 30.53 & 28.93 \\ %
        ALL w BN & \textbf{31.73} & \textbf{14.18} & \textbf{34.19}  & \textbf{29.86} \\ %
        \bottomrule
    \end{tabular}
    \caption{Effect of the different modules on the weather.} 
    \label{tab:adverseresults}
\end{table}

We further examine in \cref{tab:adverseresults} the advantage of adapting the system to accommodate different weather conditions as a remarkable feature for the decision-making of autonomous driving agents.
The basic self-training strategy, coupled with the proposed server-side aggregation scheme, has shown stability and an overall improvement over pretraining. However, due to the prevalence of clear-day images in most clients' samples, the network tends to learn more about this weather condition.
Introducing the personalized weather BNs helps to mitigate this, improving performances in adverse weather conditions, with an mIoU increase of $3\%$   on Rain and $1.78\%$ on Fog.
Similar results are evident when comparing the full method with and without the weather BNs, showcasing mIoU increases of $3.66\%$  on Rain and $0.93\%$ on Fog.
As a side note, the most difficult scenario remains the night where the improvement over the pretraining is just $0.56\%$ of mIoU.
Overall, we remark on the significance of incorporating personalized weather BNs for mitigating bias towards clear day images and enhancing performance across various weather conditions.

As a final note, considering that modern vehicles often have access to real-time weather information through several sensors (e.g., automatic windshield wipers for rain detection, clock or optical light detection for night and fog), and the weather information might be directly obtained.

\section{Additional Parameters Ablation}

Fig. \ref{fig:ablation_lambda} shows the mIoU for different values of the $\lambda_{cl}$ parameter. Best performances are achieved in the range 140-180, with quite stable maxima.
We report the effect of the queue aggregation parameter $Q$, which stabilizes at value 5 (\cref{fig:ablation_queue}).
Finally, the network does not exhibit sensitivity to the smoothing parameter of the prototypes $\beta$ ($\beta = 0.85) $ ( \cref{fig:ablation_smoothing}).

\begin{figure}[t]
\vspace{-7mm}
    \centering
    \includegraphics[width=0.45\textwidth]{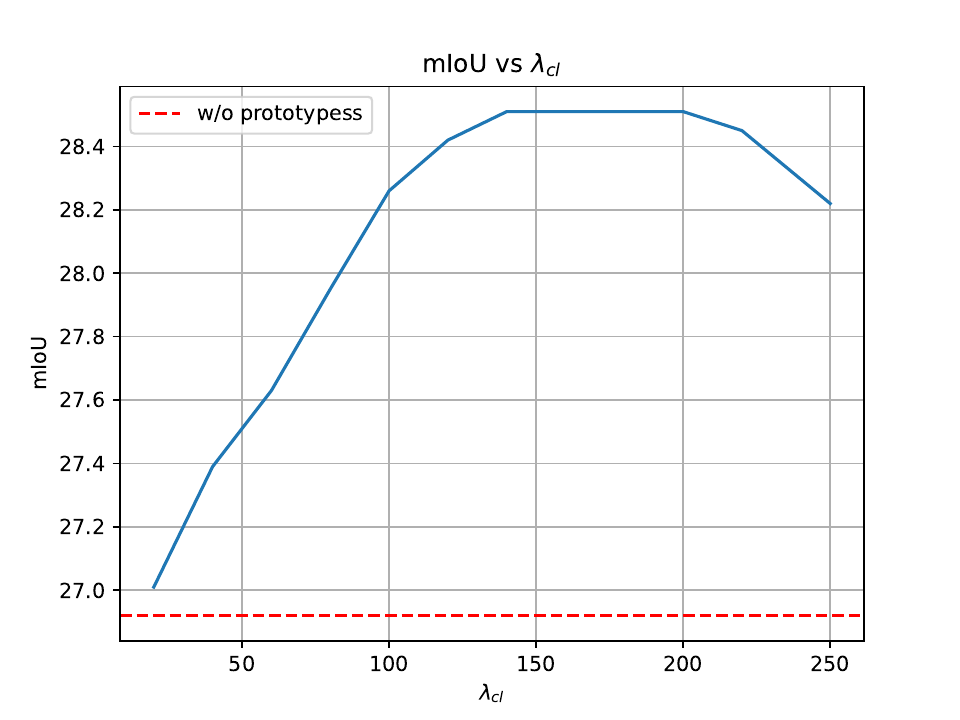}
    \caption{Tuning on $\lambda_{cl}$ weight.}
    \label{fig:ablation_lambda}
\end{figure}
\begin{table}[t]
    \centering
      \resizebox{0.5\textwidth}{!}{%
        \begin{tabular}{cccccc}
            \toprule
            \multirow{2}{*}{Method} & \multirow{2}{*}{Backbone} & \multirow{2}{*}{\# Params} & \multirow{2}{*}{\begin{tabular}[c]{@{}c@{}}Supervised\\ on\end{tabular}} & \multirow{2}{*}{\begin{tabular}[c]{@{}c@{}}Unsup.\\ on\end{tabular}} & \multirow{2}{*}{\begin{tabular}[c]{@{}c@{}}City\\ mIoU\end{tabular}} \\
             &  &  &  \\ \midrule
            FedDrive \citesupp{fantauzzo2022feddrive}  & BiseNetV2 & 8.2M & City & -- & 43.85 \\
            Fed. Oracle† & MobileNetV2 & 3.4M & City & -- & 58.16 \\
            Fed. Fine-Tune† & MobileNetV2 & 3.4M & GTAV $\rightarrow$ City & -- & 59.35 \\ \midrule
            Source Only  & MobileNetV2 & 3.4M & GTAV & -- & 20.23 \\%/ 28.16 \\ %
            LADD \citesupp{shenaj2023learning} † & MobileNetV2 & 3.4M & GTAV & City &  36.49 \\
            Ours & MobileNetV2 & 3.4M & GTAV & City & 38.21 \\ %
            \bottomrule
        \end{tabular}
        }
    \caption{GTA $\rightarrow$ Cityscapes results.}
    \label{tab:gta2cityscapes}
\end{table}

\section{Per-class Results}
Tab. \ref{tab:per_class_iou} contains a more detailed version of the results in \cref{tab:ablation} %
of the main paper that also shows the class-by-class accuracy. As common on the employed datasets, the mIoU is higher in the common classes (e.g., road or building) and lower in the rare and more challenging ones. However, it can be seen that by adding the various components of the model, results tend to increase consistently in most classes, even if a few challenging ones remain hard to detect. Some classes like traffic sign, car or bicycle show impressive improvements.
On the other side, there are a few cases in which our approach does not improve %
the results on some under-represented classes such as terrain and rider.

\section{Qualitative Results}

\begin{figure}[t]
    \vspace{-7mm}
    \centering
    \includegraphics[width=0.45\textwidth]{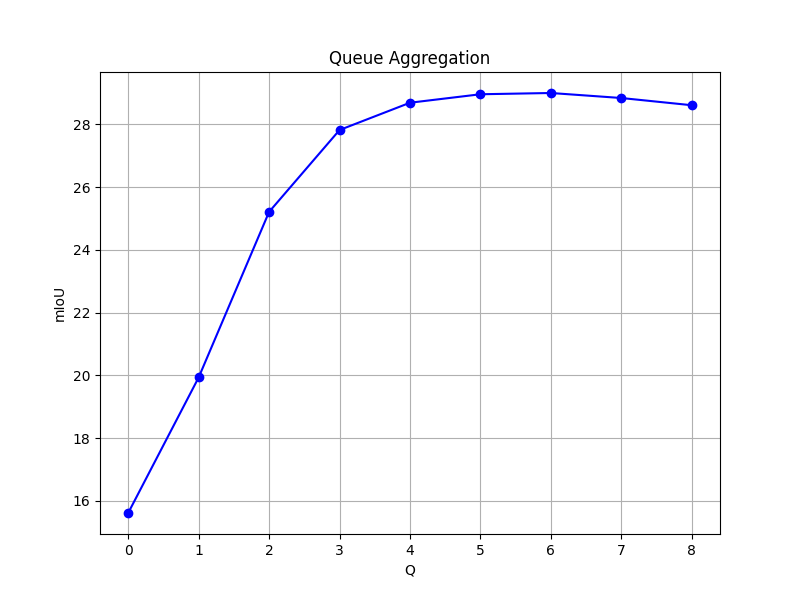}
    \caption{Tuning on the queue aggregation parameter $Q$.}
    \label{fig:ablation_queue}
    \vspace{1em}
    \centering
    \includegraphics[width=0.45\textwidth]{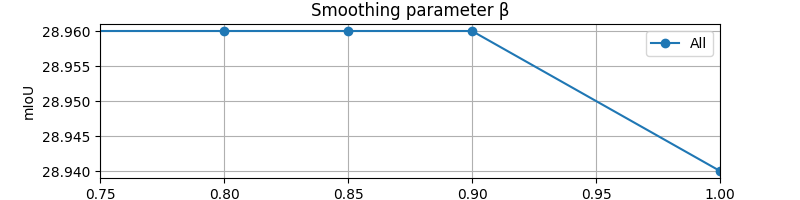}
    \caption{Tuning on the prototype smoothing parameter $\beta$.}
    \label{fig:ablation_smoothing}
\end{figure}

Fig.~\ref{fig:grid} shows some qualitative results for car and drone samples in both clear and adverse weather.
FedAvg \citesupp{fedavg} and LADD \citesupp{shenaj2023learning} appear to be overconfident on the road and car classes, respectively.
While our model adeptly captures the structure of the street better than other models in both adverse and non-adverse conditions, the adverse conditions images remain a challenging aspect.  %
For drones, the discrepancies between the predicted classes and the ground truth arise also due to the fact that the network is trained on a superset of classes beyond those present in the drone dataset. Notice that sometimes the network assigns ``fine'' classes that share a semantic affiliation with the respective ``coarse'' classes, e.g., predict terrain instead of vegetation, but on the drone dataset labeling terrain is not present and terrain samples have ground truth set to vegetation.
The same thing happens for the sidewalk and road or rider and person.
For this reason, we remapped the 19 classes of the full set into the 5 drone ones using the mapping proposed in \citesupp{rizzoli2023syndrone}. The figure shows both the original prediction maps from the network and the ones obtained after remapping the 19 classes into the 5 of the drone datasets.
The comparison with the ground truth shows that some predictions not matching are just due to the different types of labeling in drone and car datasets. %

\begin{figure*}[t]
  \centering
  \resizebox{\textwidth}{!}{%
  \begin{tabular}{0c 0c 0c 0c 0c 0c}
    \multicolumn{1}{c}{} & \multicolumn{1}{c}{\textbf{RGB}} & \multicolumn{1}{c}{\textbf{GT}} &  \multicolumn{1}{c}{\textbf{FedAvg}\citesupp{fedavg}\textbf{+ST}} & \multicolumn{1}{c}{\textbf{LADD}\citesupp{shenaj2023learning}} & \multicolumn{1}{c}{\textbf{HyperFLAW}}\\ 
    \makecell{\textbf{CAR} \\ \textbf{CLEAR}}
    & 
    \includegraphics[width=0.15\linewidth, valign=c]{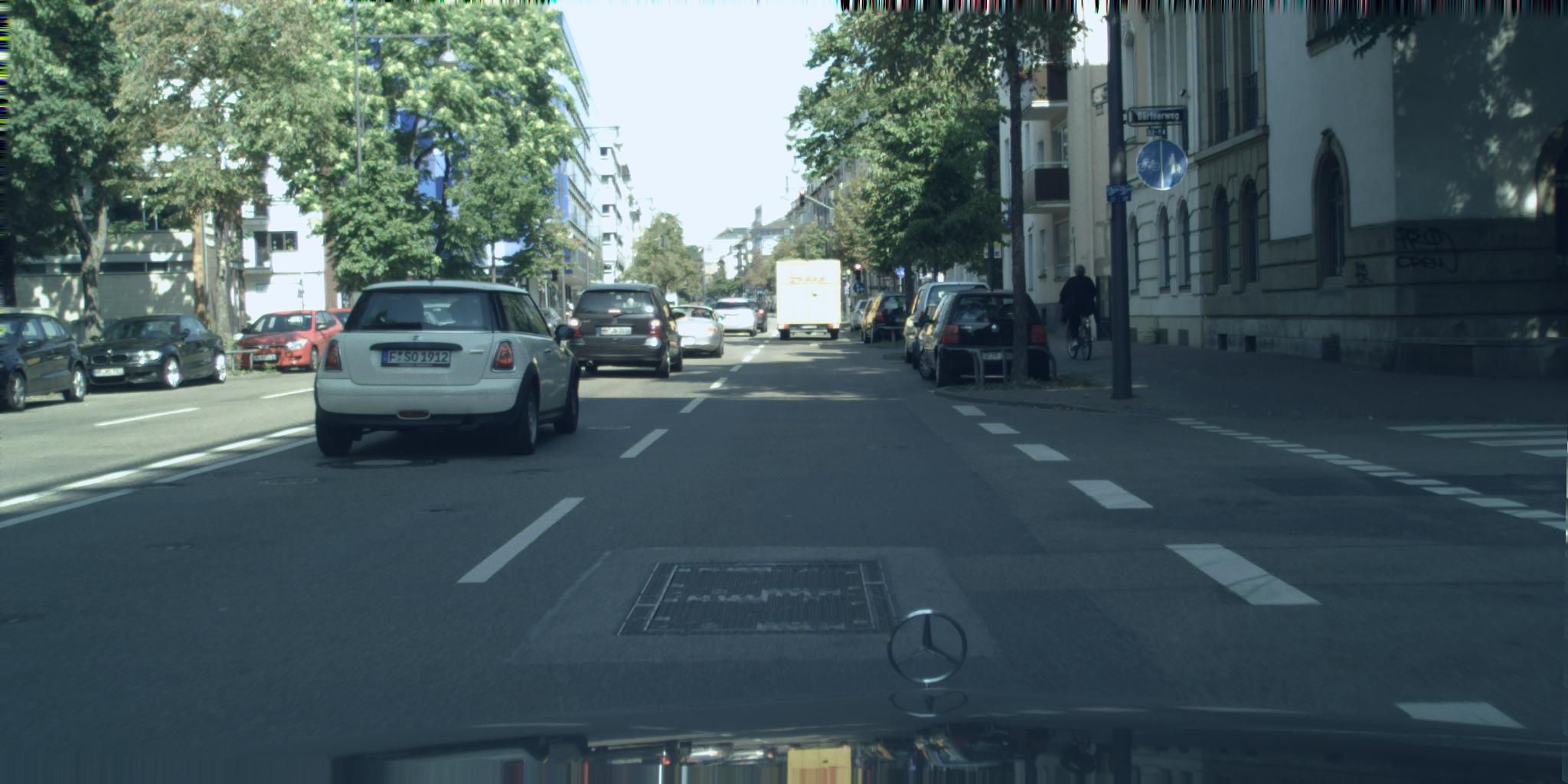} &
    \includegraphics[width=0.15\linewidth, valign=c]{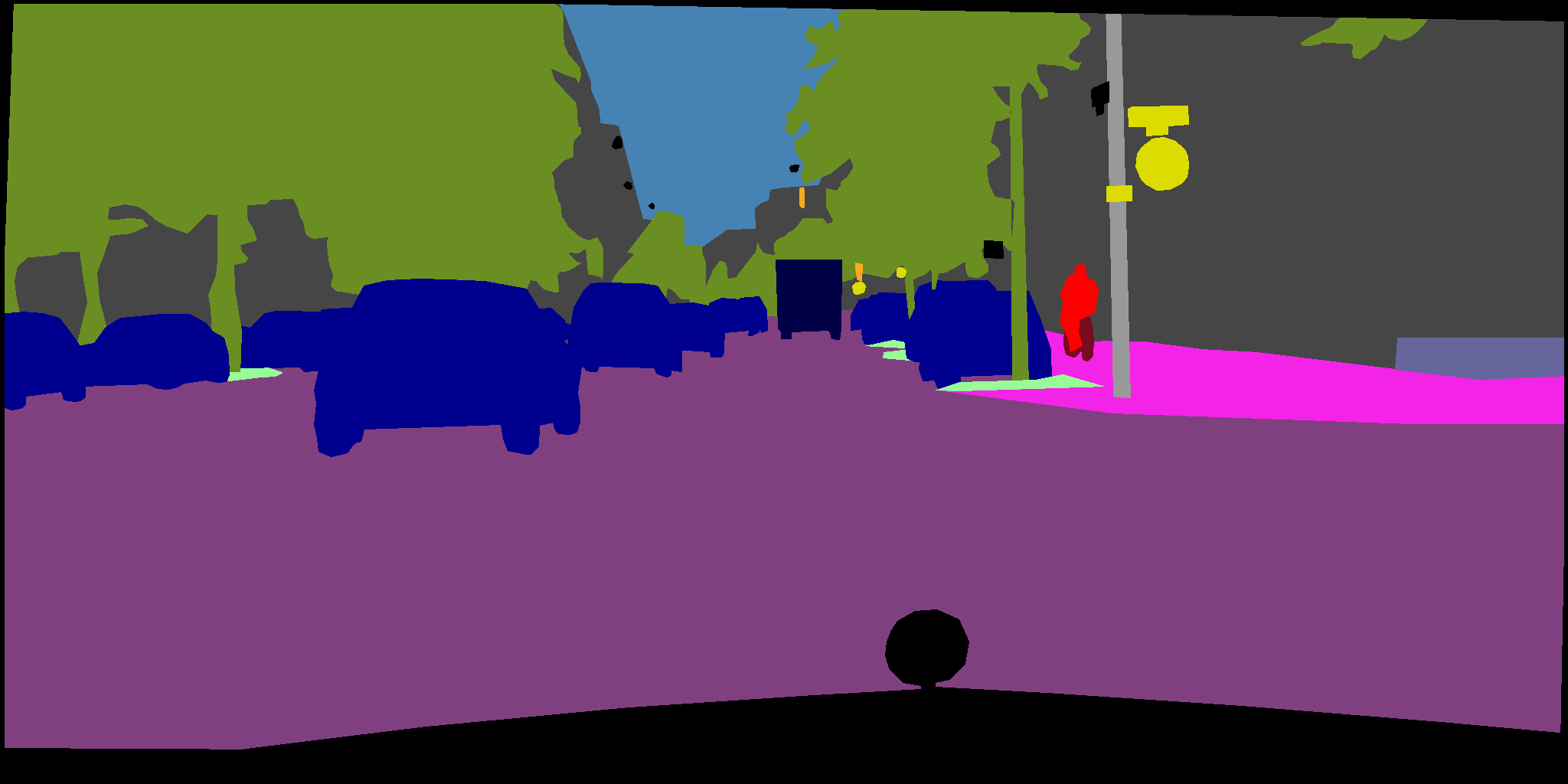} &
    \includegraphics[width=0.15\linewidth, valign=c]{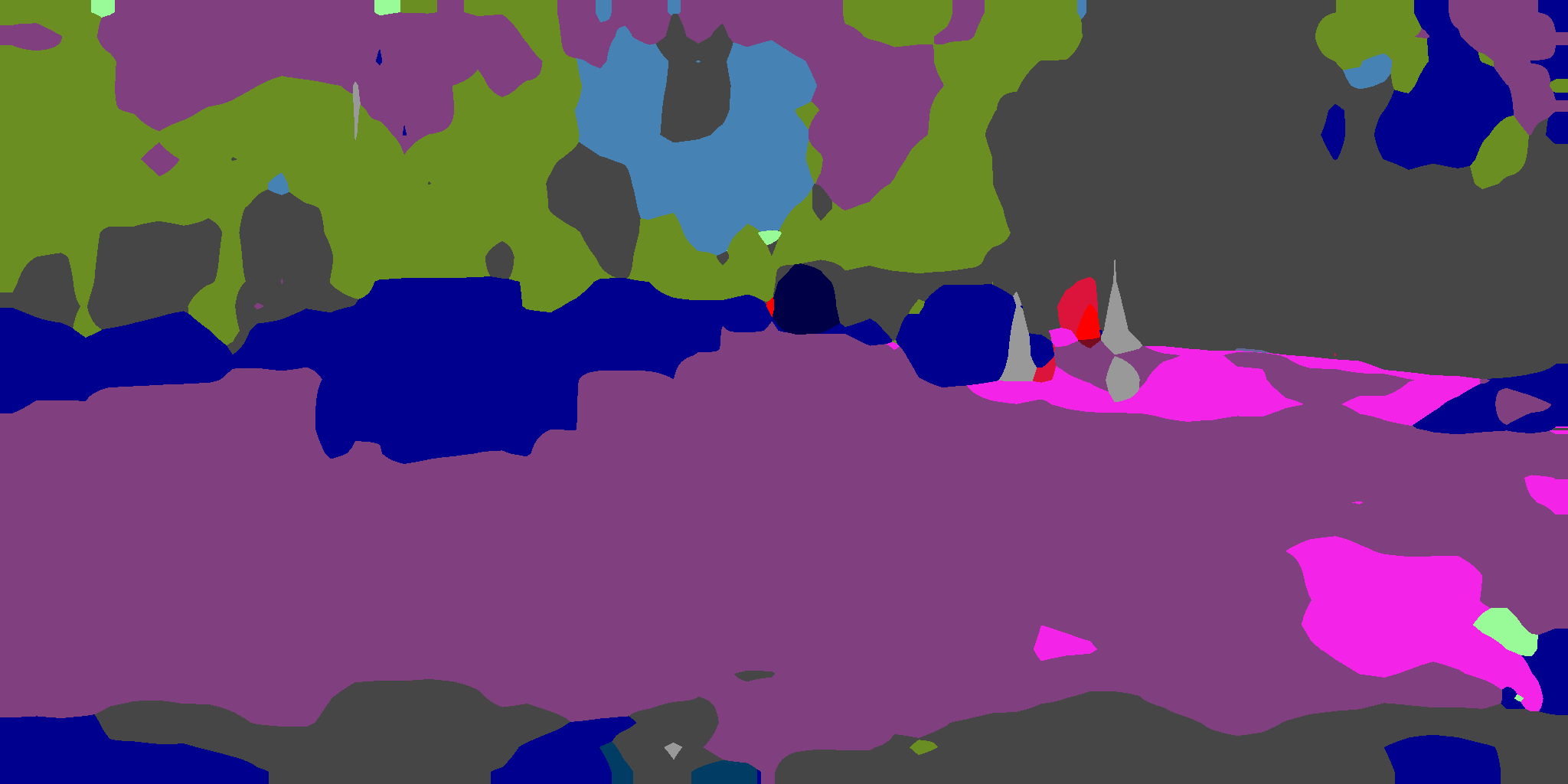} &
    \includegraphics[width=0.15\linewidth, valign=c]{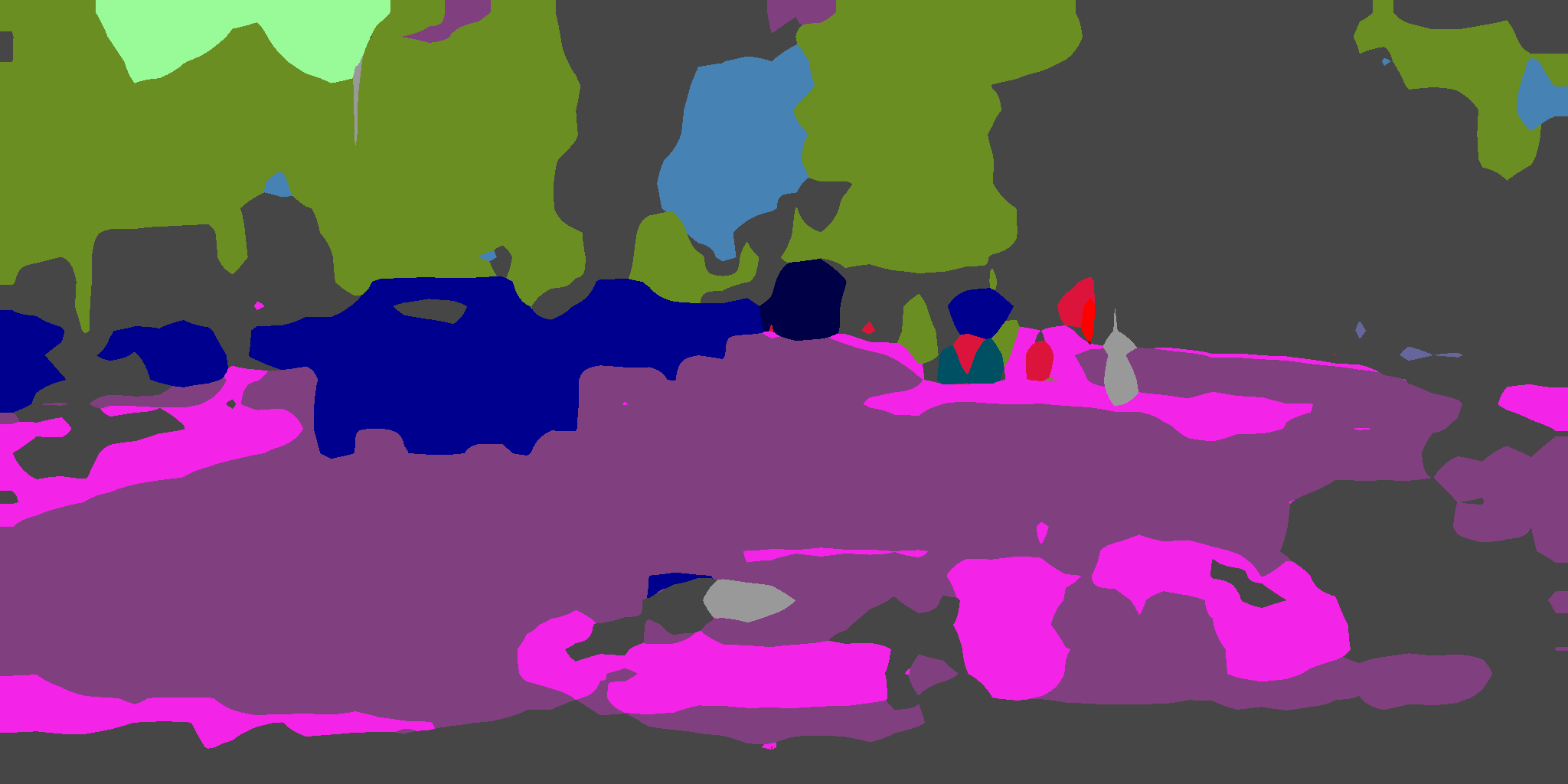} &
    \includegraphics[width=0.15\linewidth, valign=c]{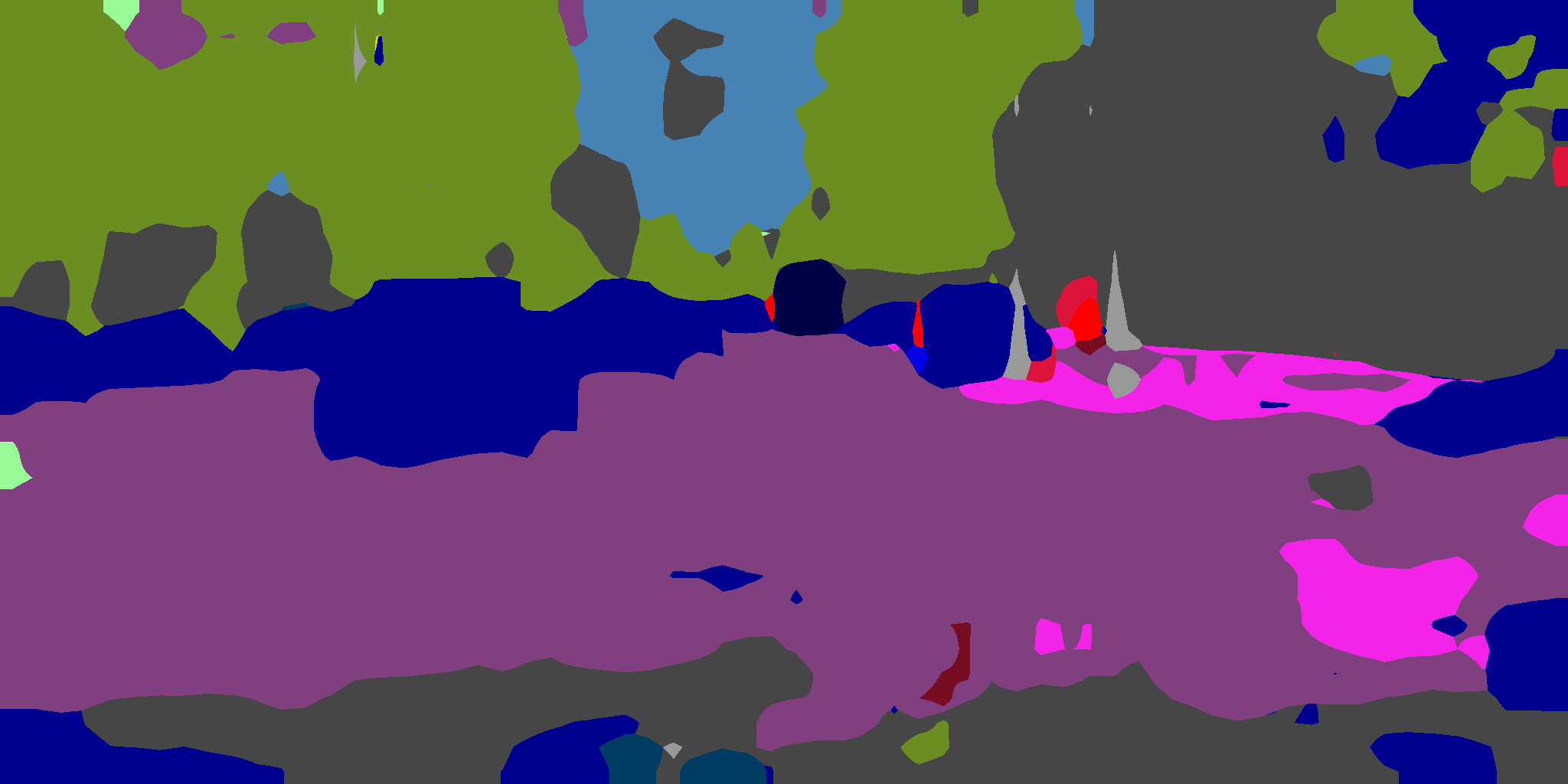} 
    \\
    \makecell{\textbf{DRONE} \\ \textbf{CLEAR}}
    & 
    \multirow[c]{2}{*}{ \includegraphics[width=0.15\linewidth, trim={0 0 0 -20cm}]{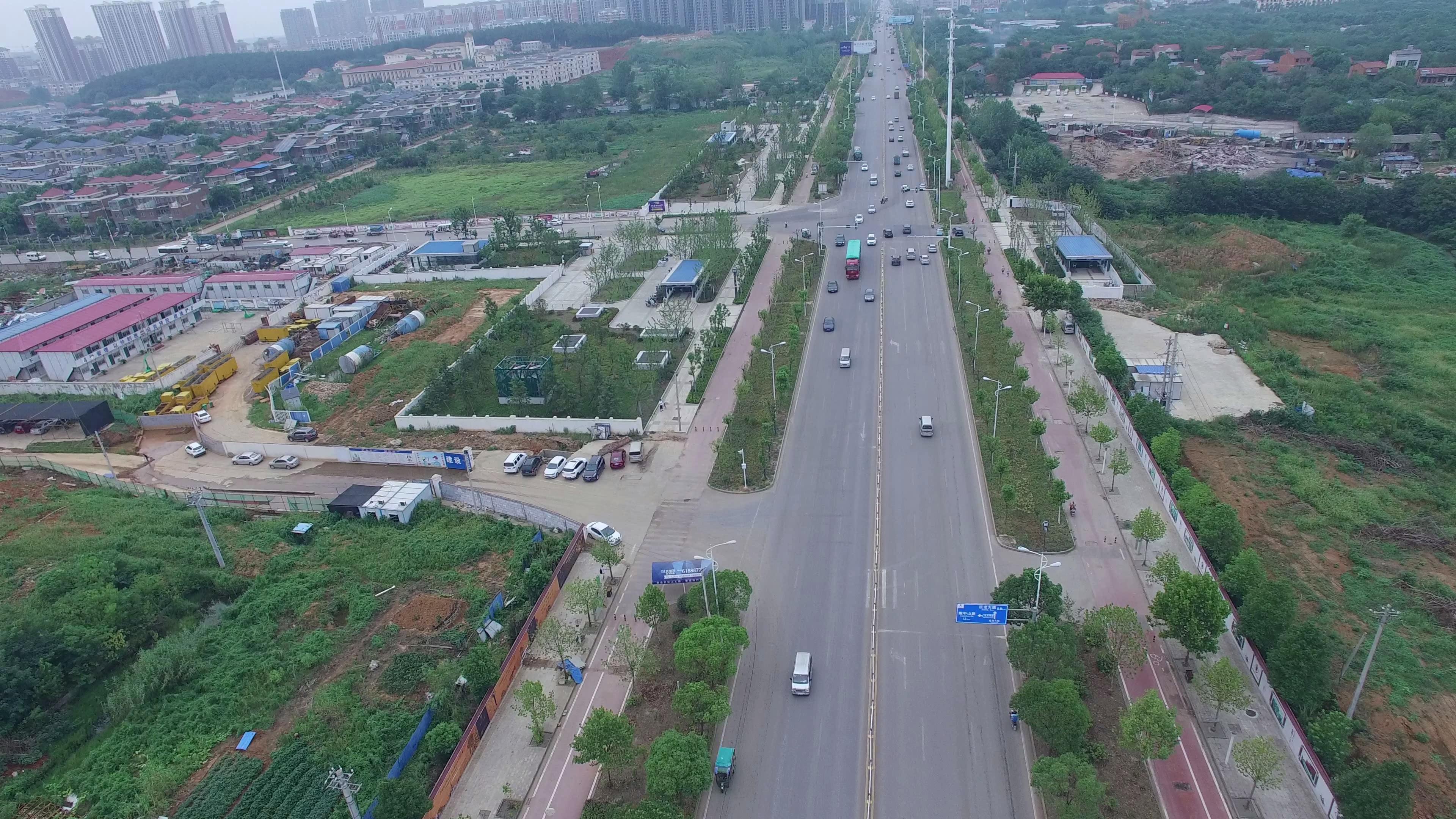}} &
    \multirow[c]{2}{*}{ \includegraphics[width=0.15\linewidth, trim={0 0 0 -20cm}]{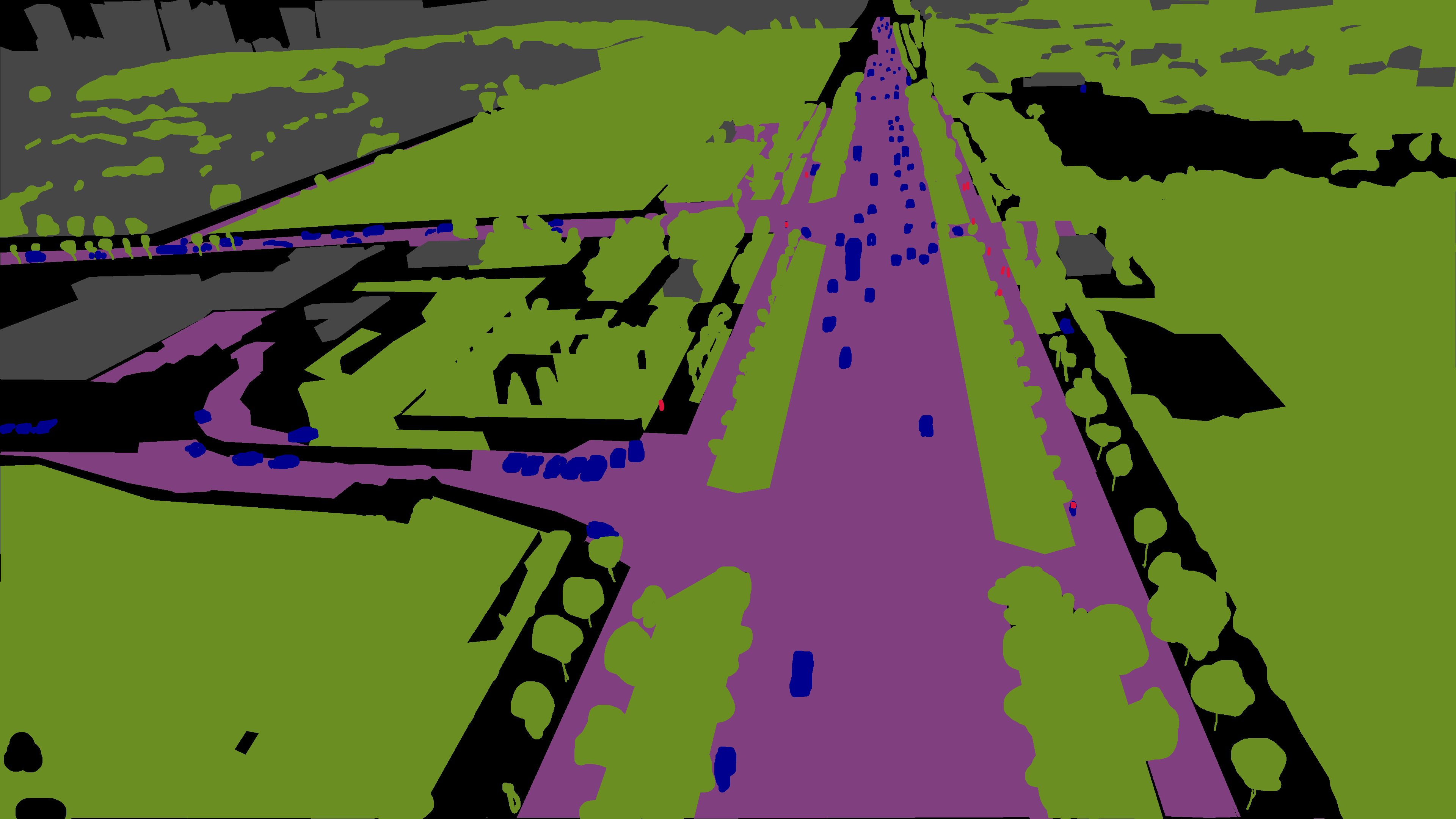}} &
    \includegraphics[width=0.15\linewidth, valign=c]{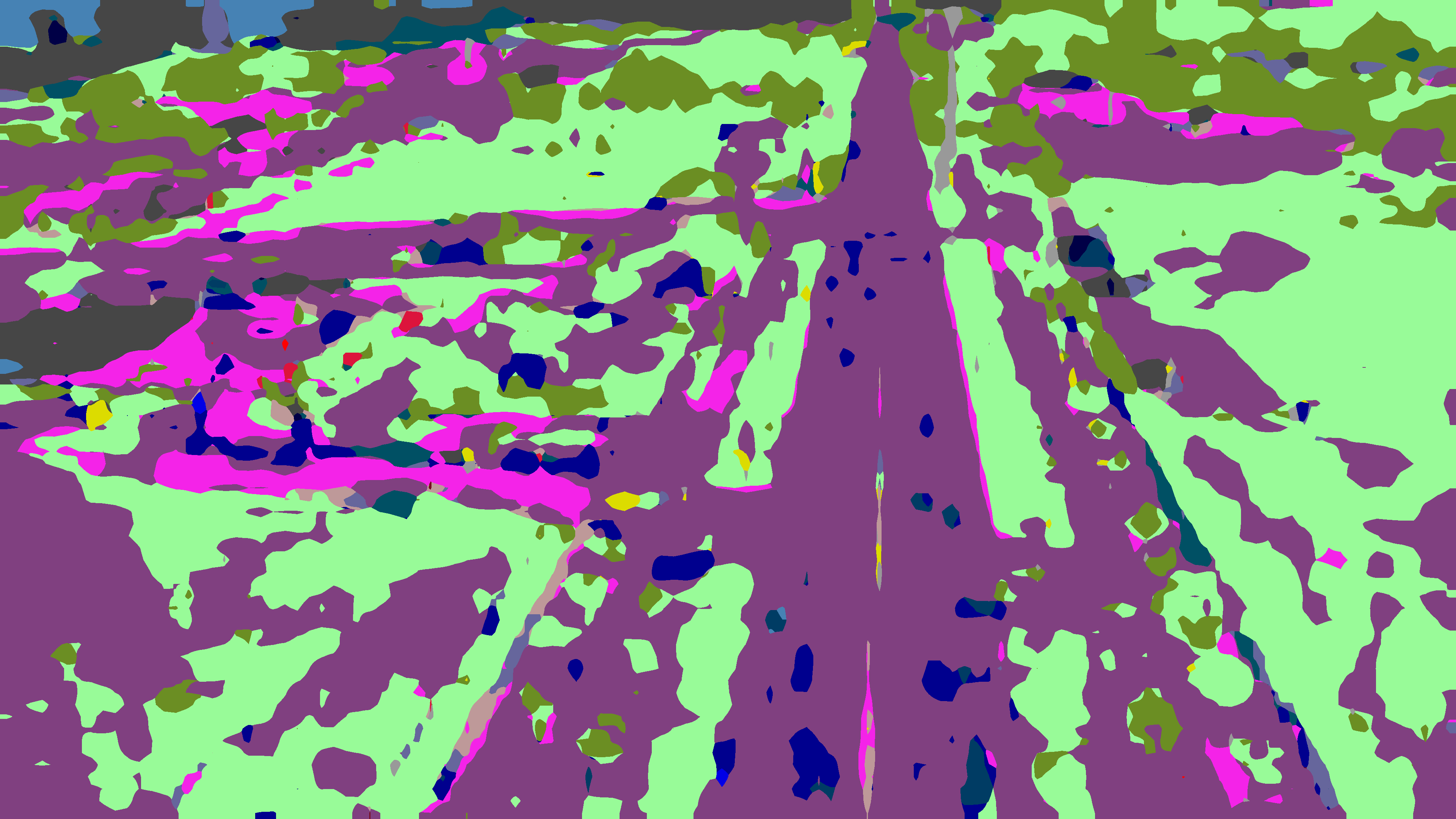} &
    \includegraphics[width=0.15\linewidth, valign=c]{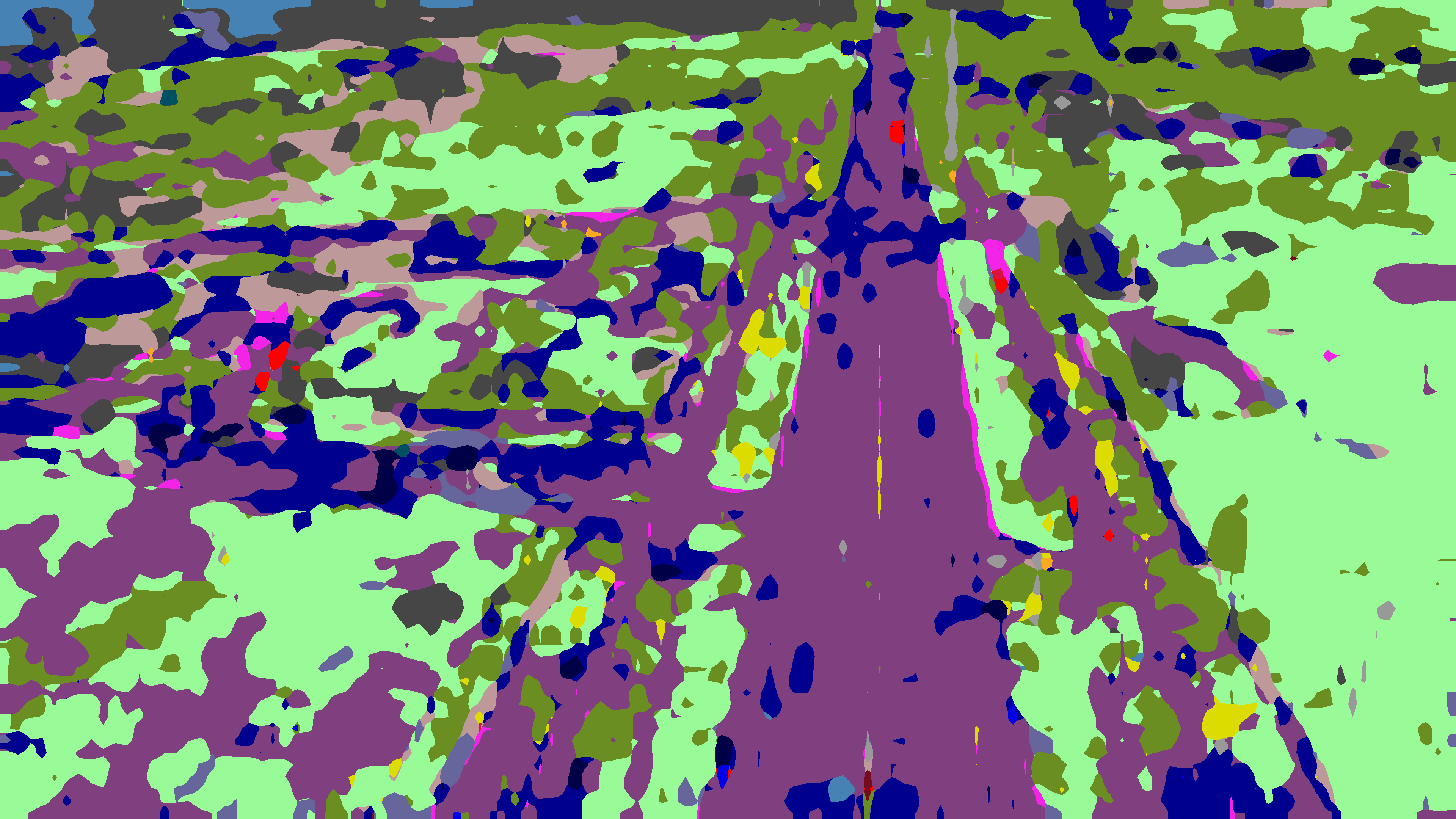} &
    \includegraphics[width=0.15\linewidth, valign=c]{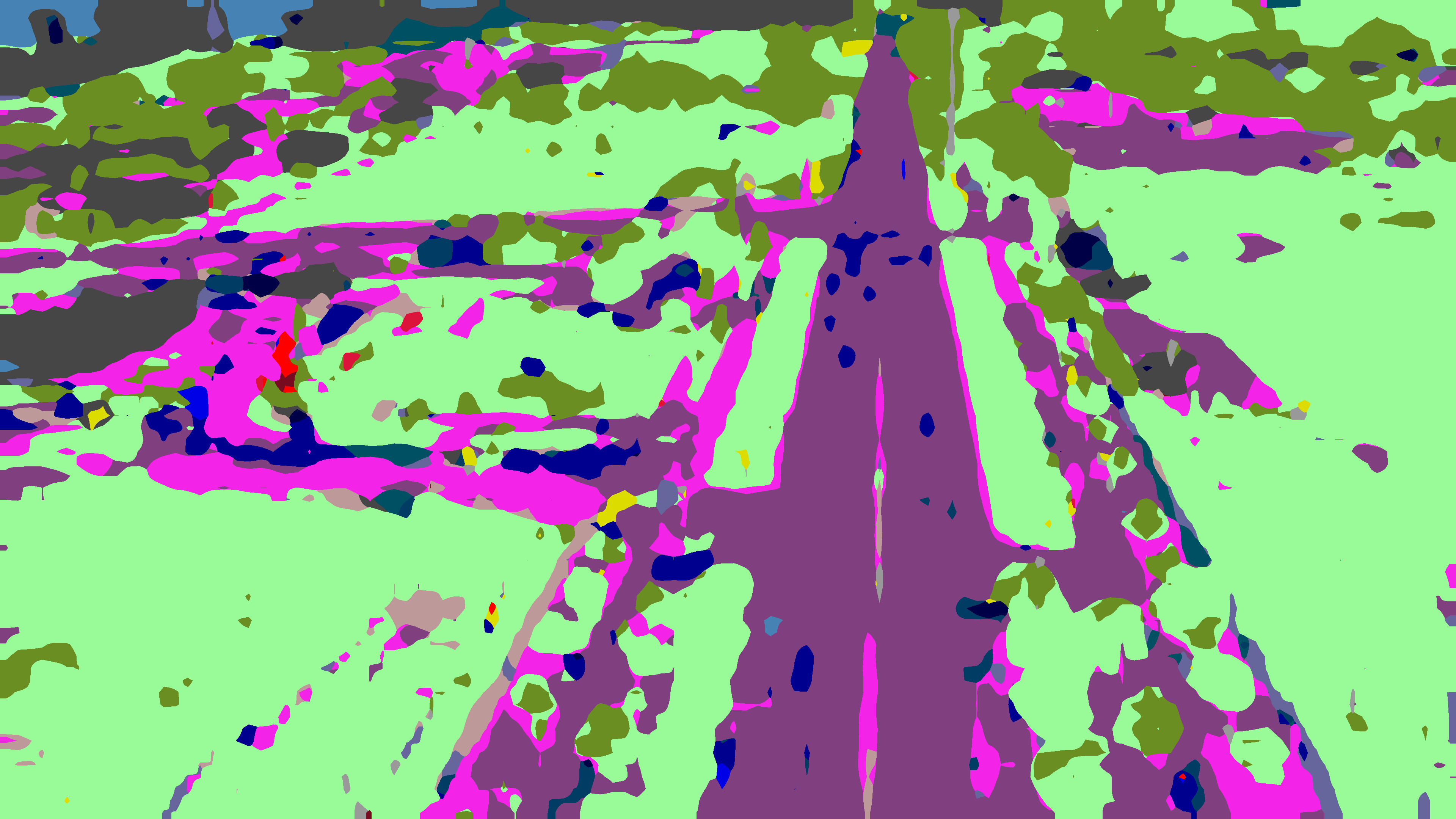} 
    \\
    \makecell{\textbf{DRONE} \\ \textbf{CLEAR}\\ (remapped)}
    & 
    &%
    &%
    \includegraphics[width=0.15\linewidth, valign=c]{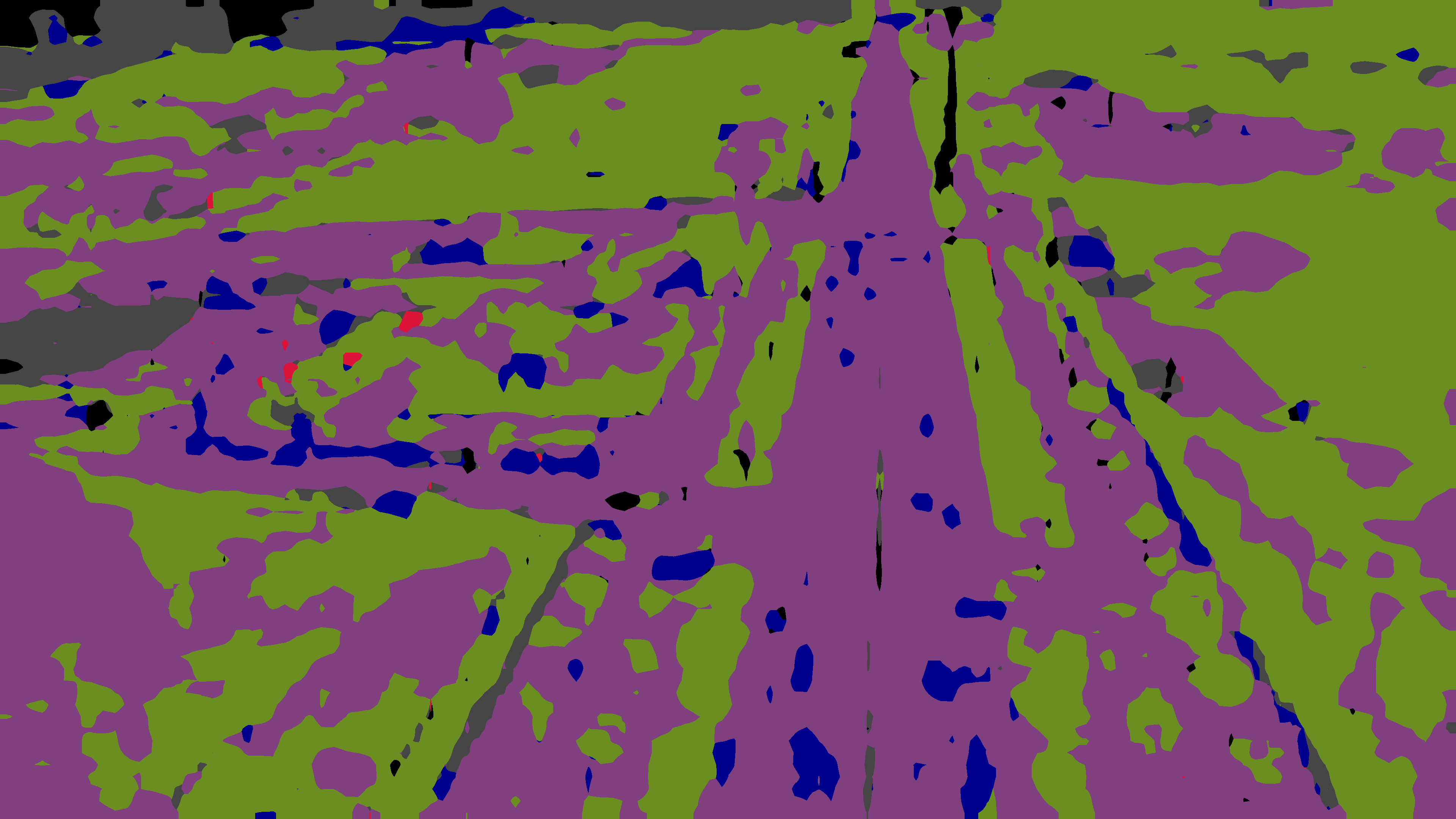} &
    \includegraphics[width=0.15\linewidth, valign=c]{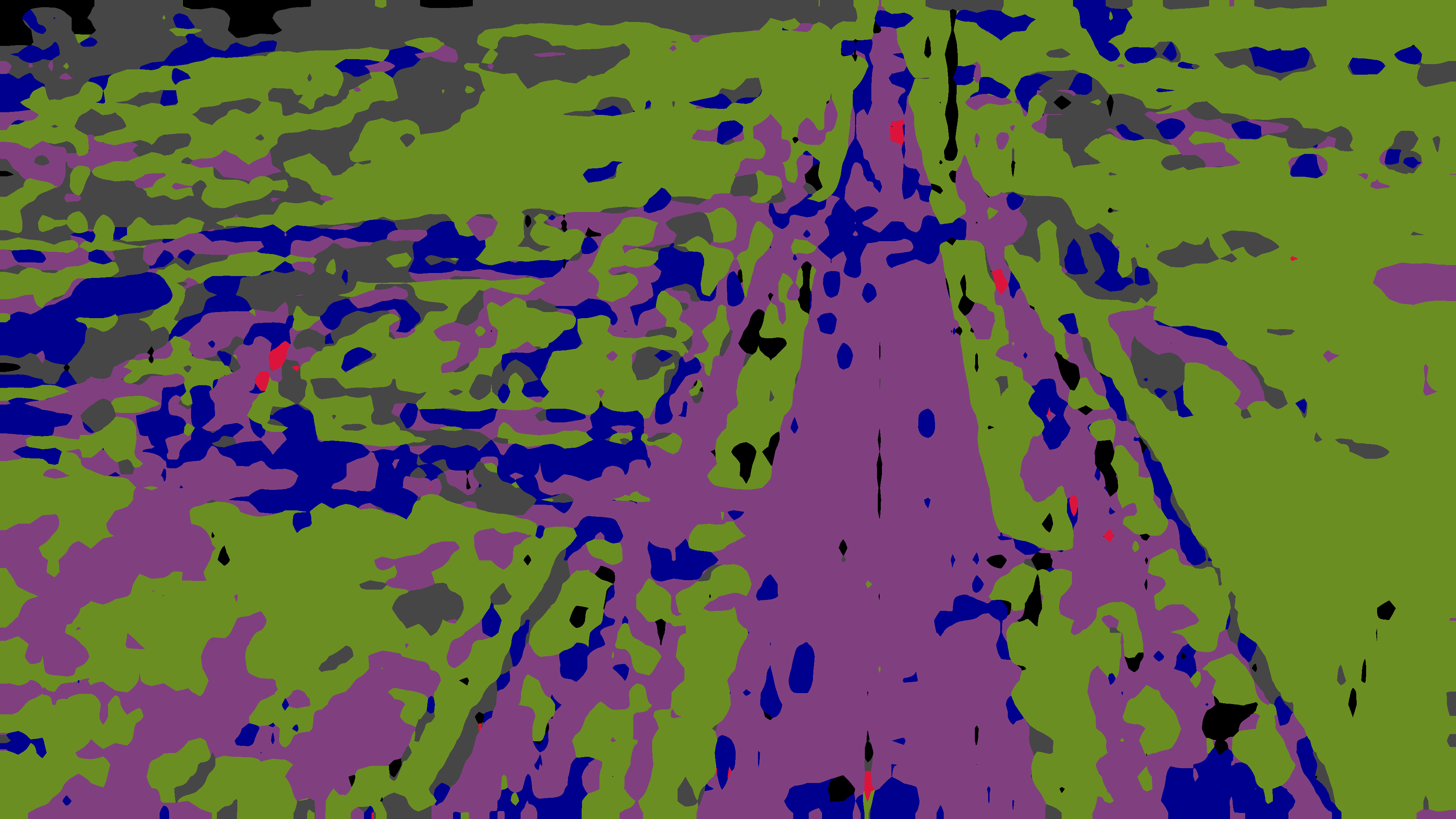} &
    \includegraphics[width=0.15\linewidth, valign=c]{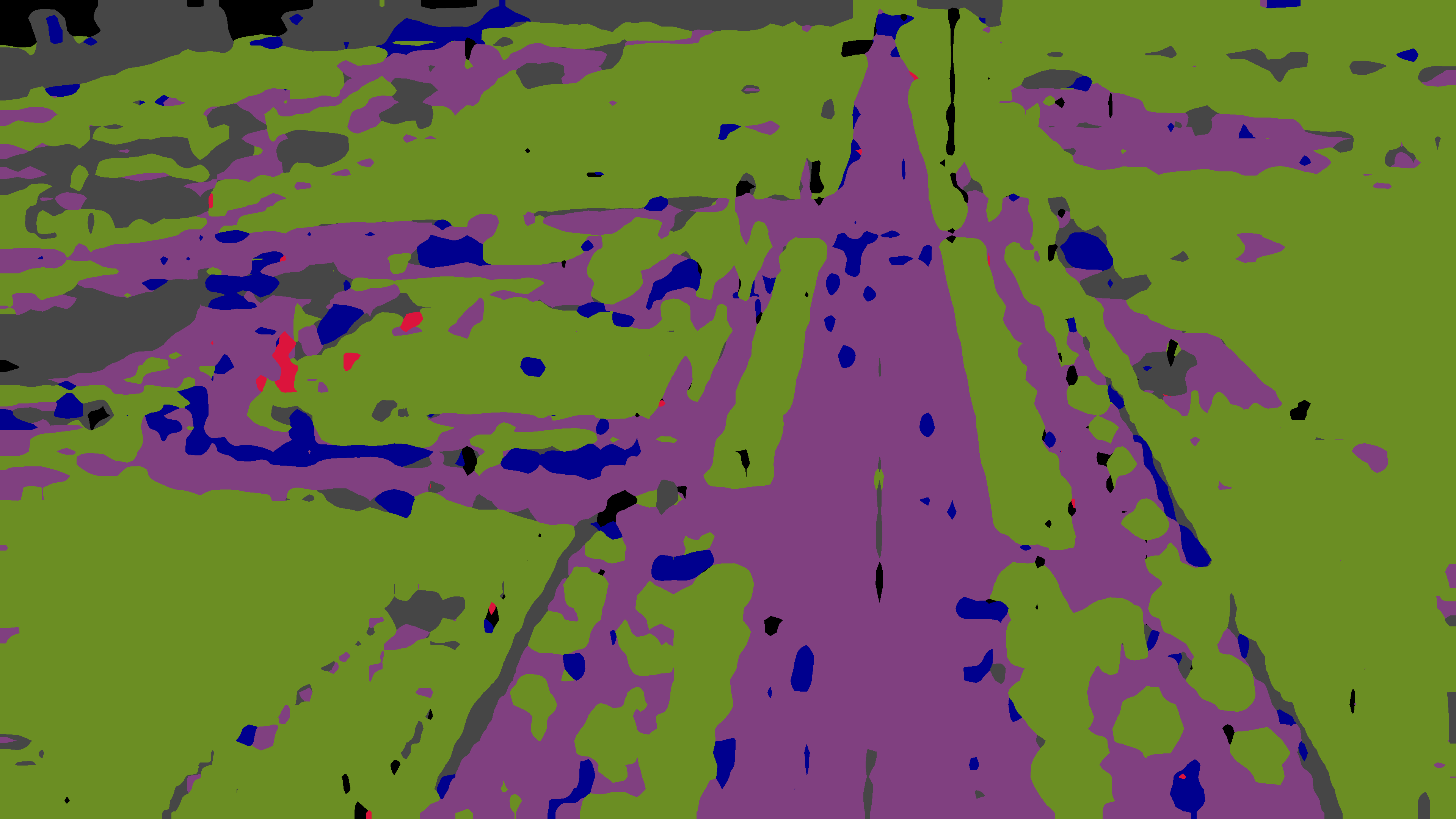} 
    \\
    \makecell{\textbf{CAR} \\ \textbf{ADVERSE}}
    & 
    \includegraphics[width=0.15\linewidth, valign=c]{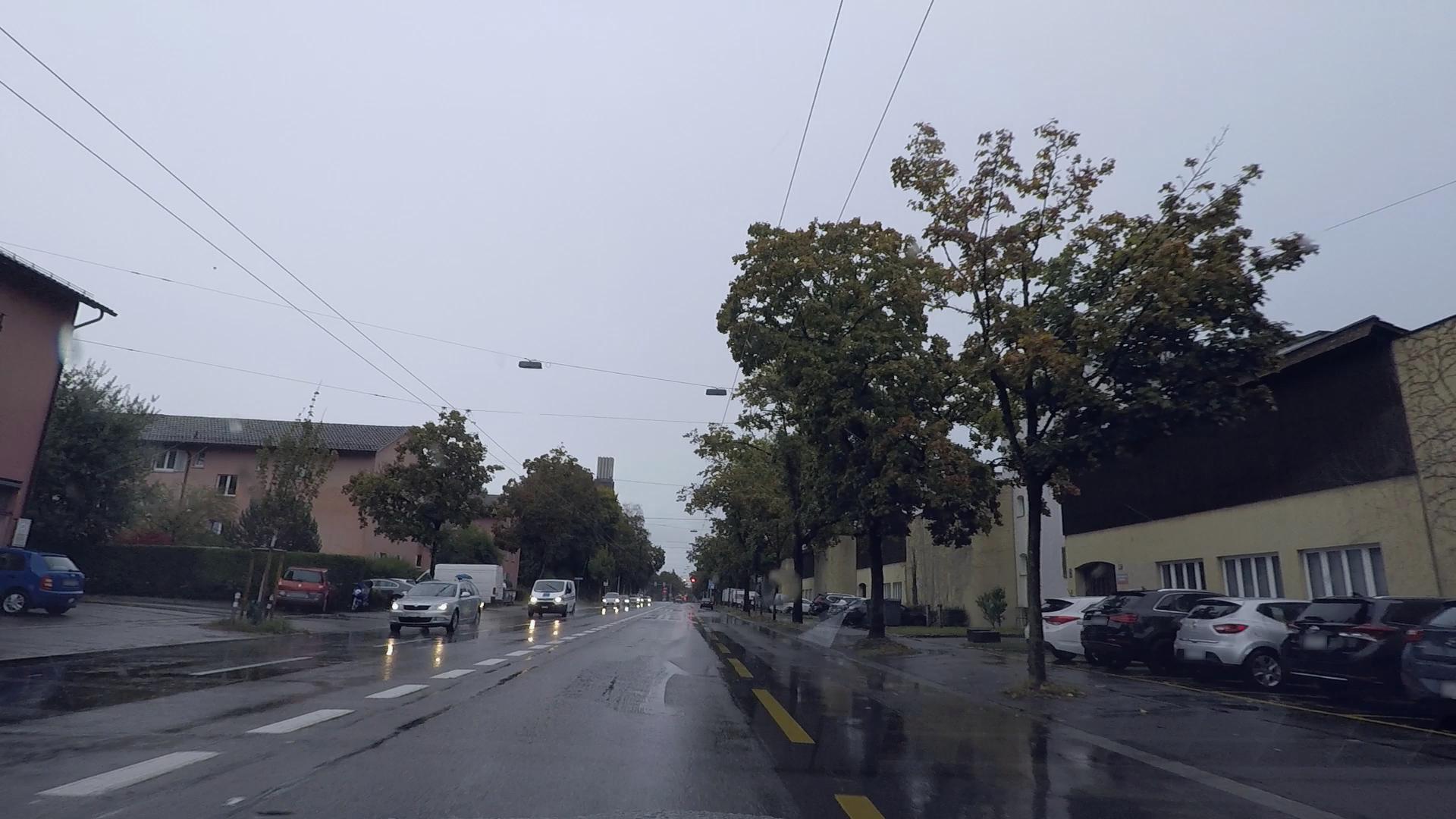} &
    \includegraphics[width=0.15\linewidth, valign=c]{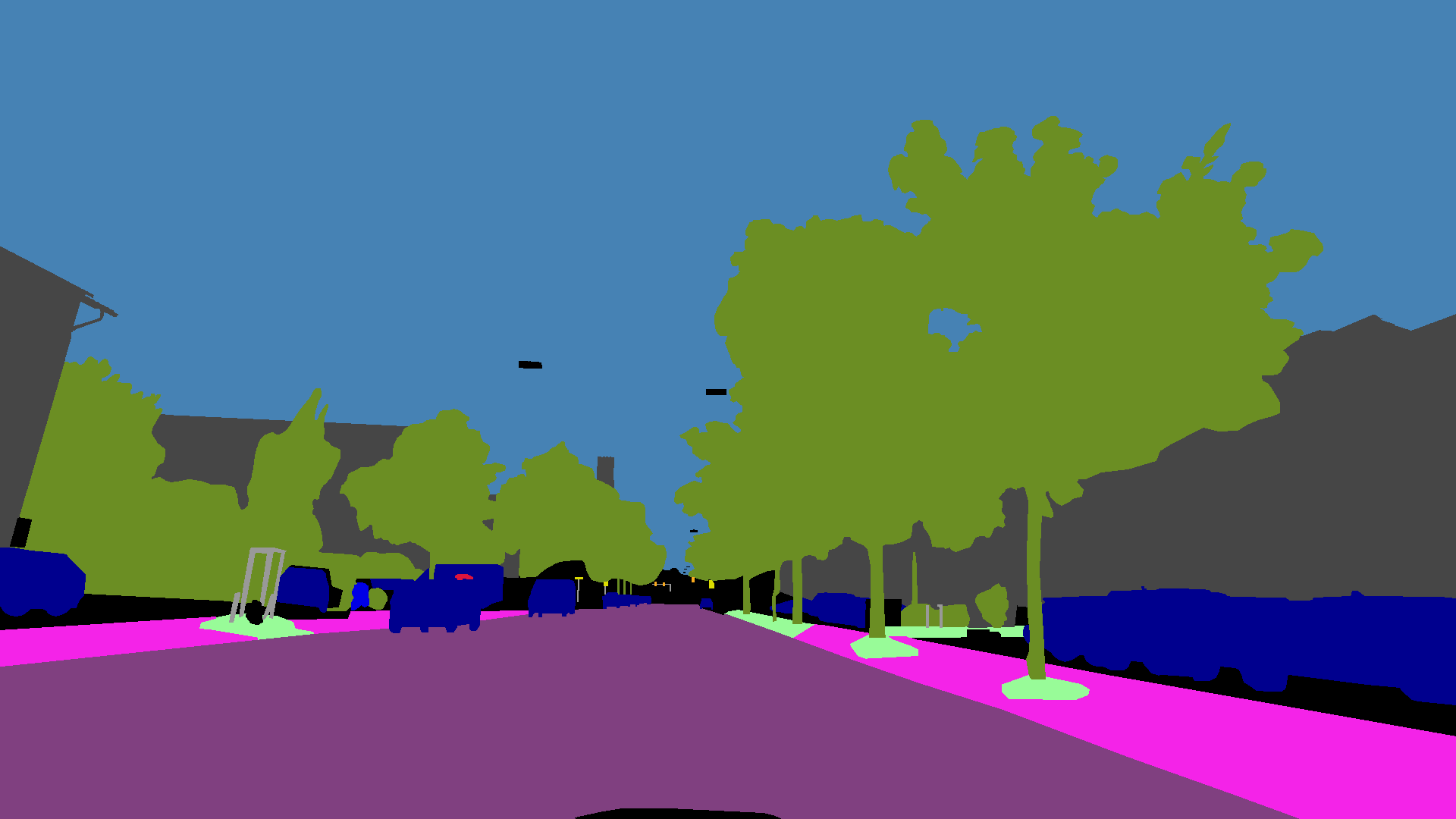} &
    \includegraphics[width=0.15\linewidth, valign=c]{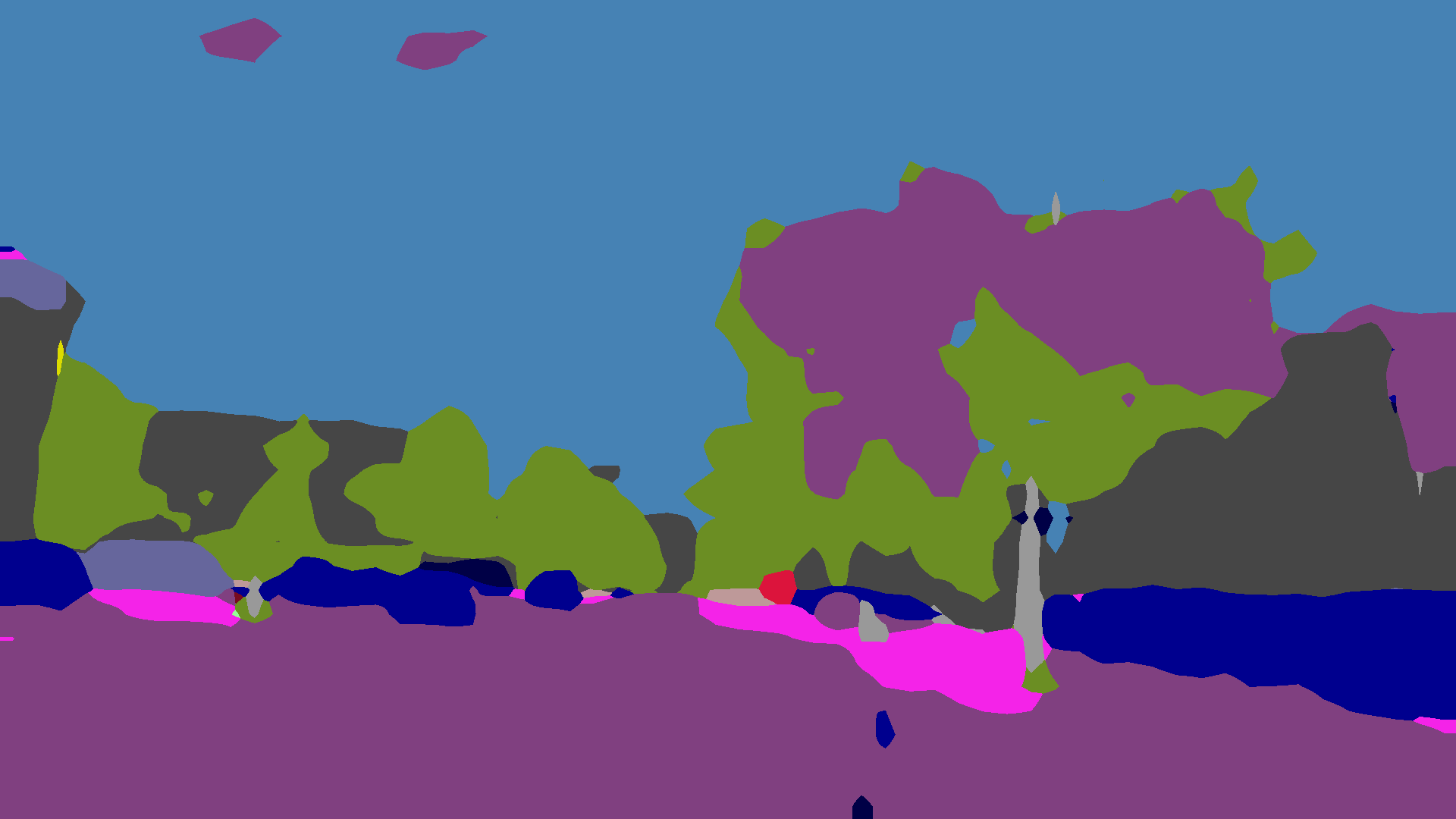} &
    \includegraphics[width=0.15\linewidth, valign=c]{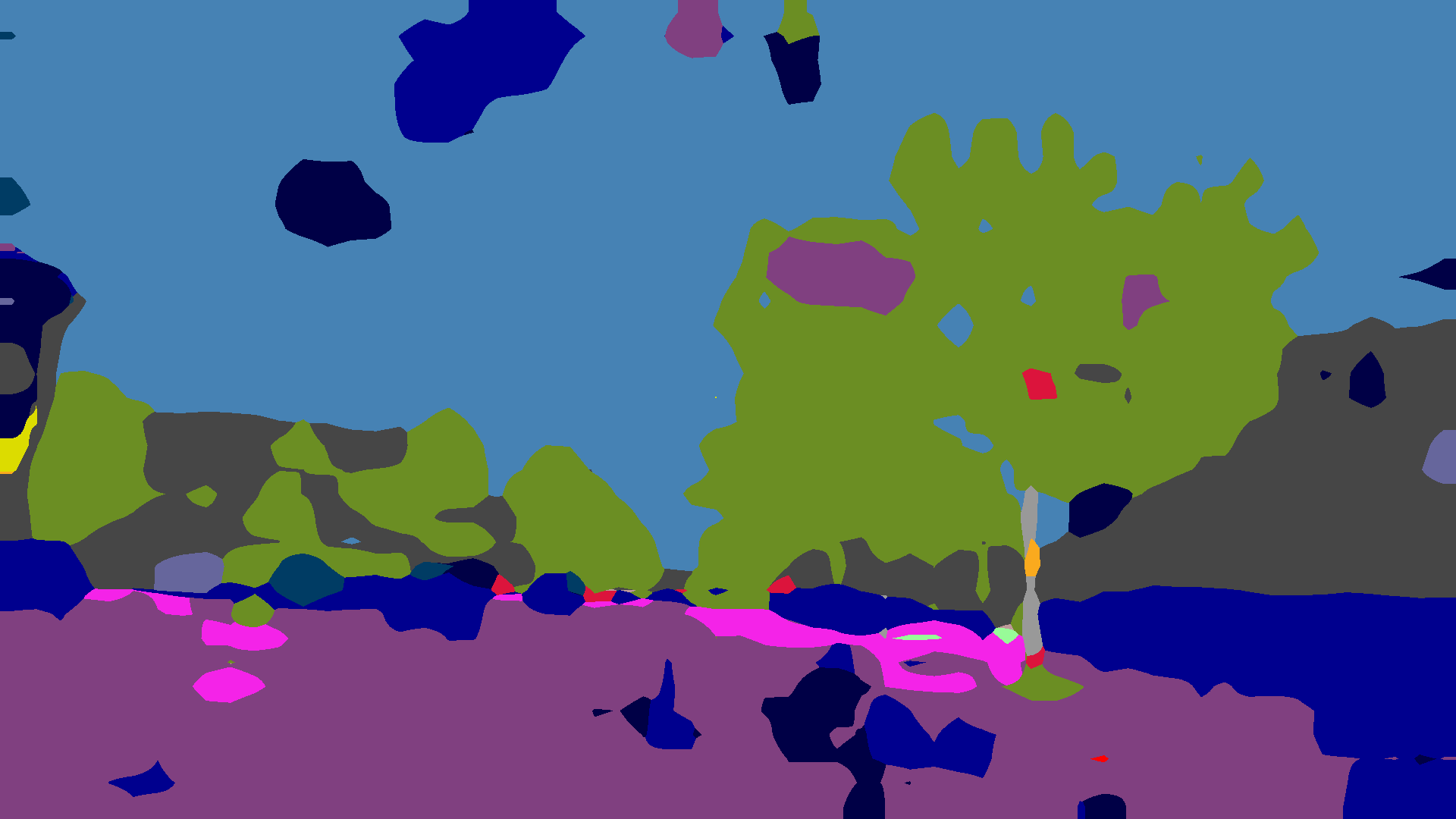} &
    \includegraphics[width=0.15\linewidth, valign=c]{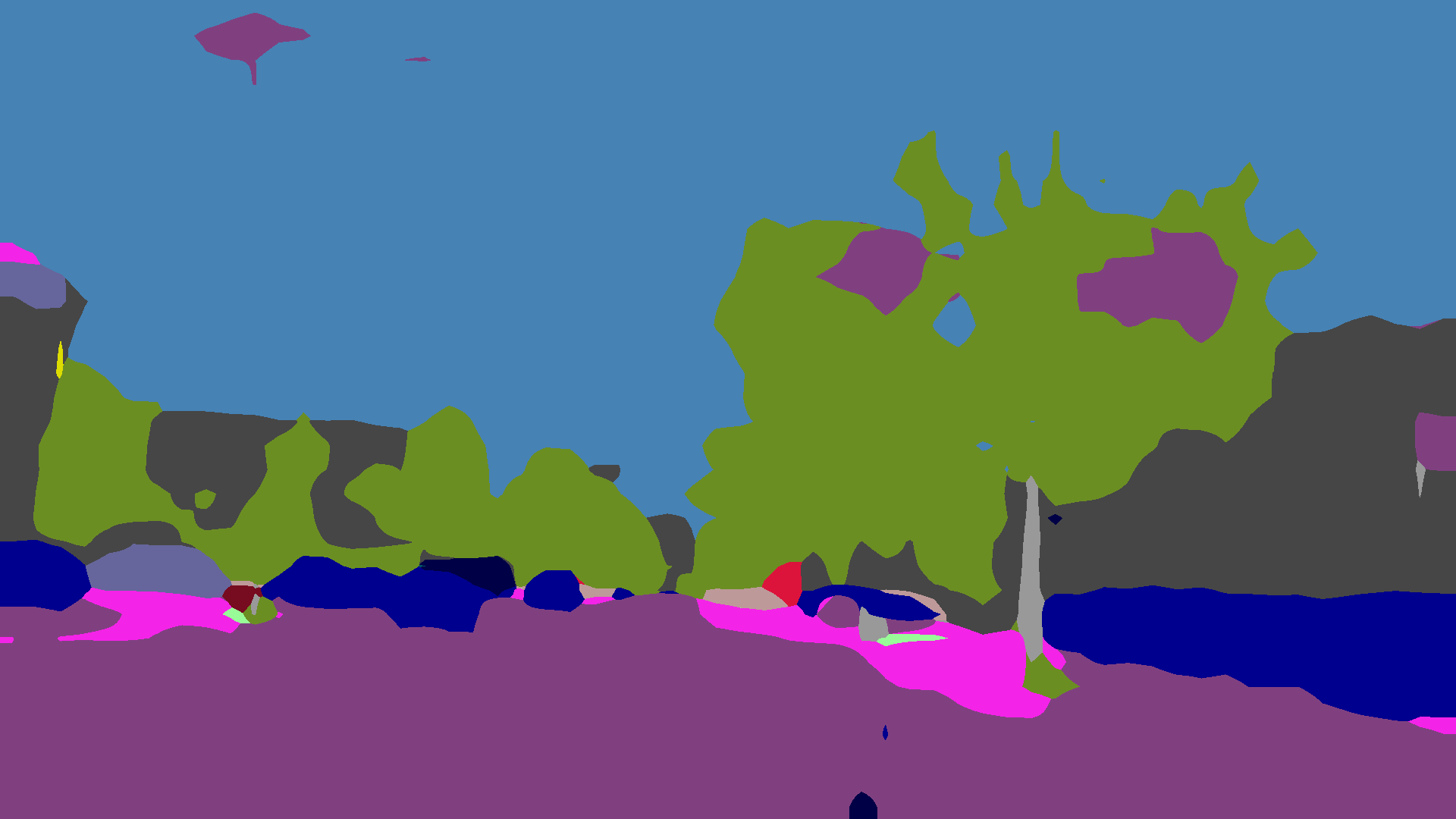} 
    \\
    \makecell{\textbf{DRONE} \\ \textbf{ADVERSE}}
    & 
    \multirow[c]{2}{*}{ \includegraphics[width=0.15\linewidth, trim={0 0 0 -20cm}]{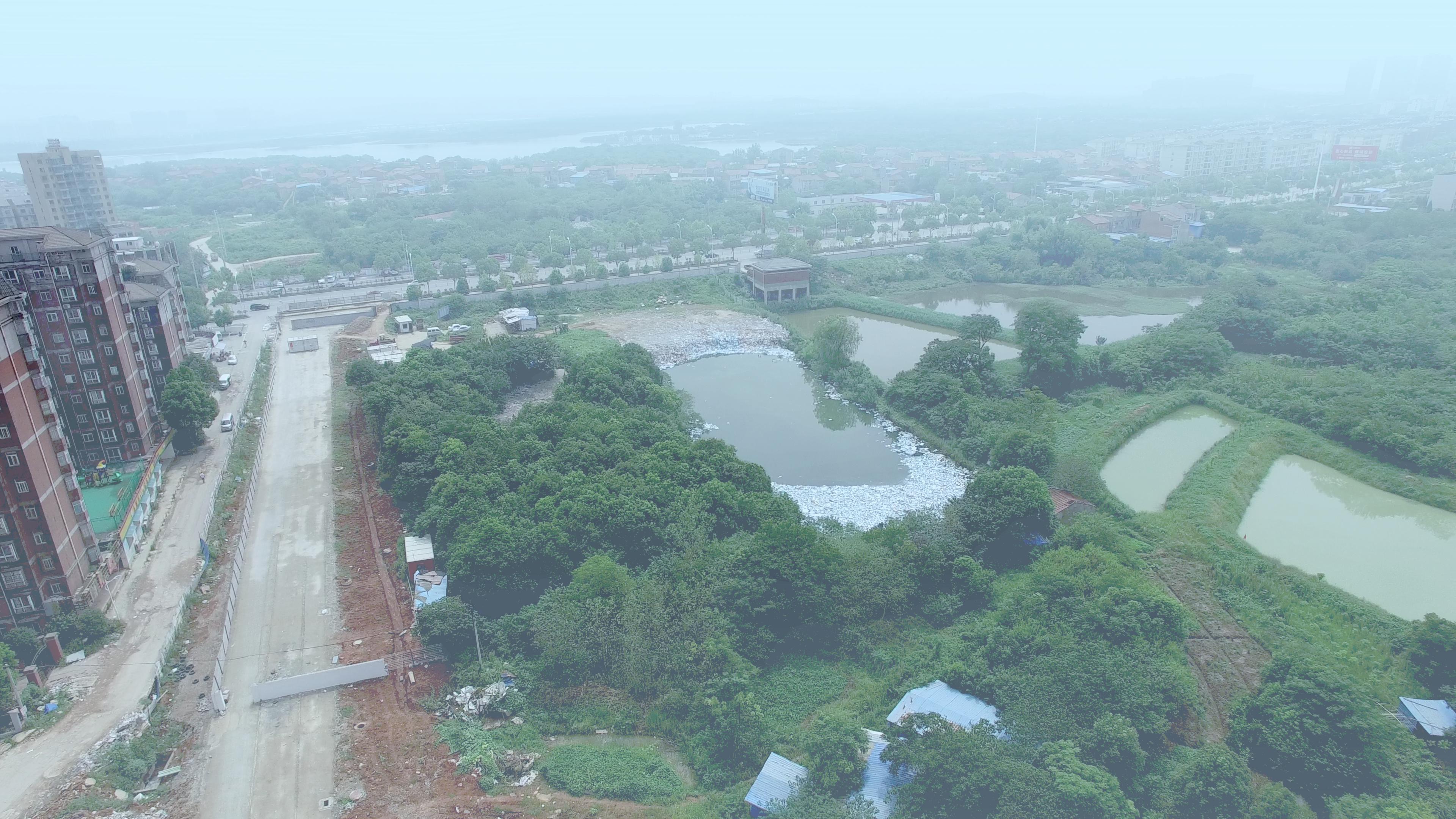}} &
    \multirow[c]{2}{*}{\includegraphics[width=0.15\linewidth, trim={0 0 0 -20cm}]{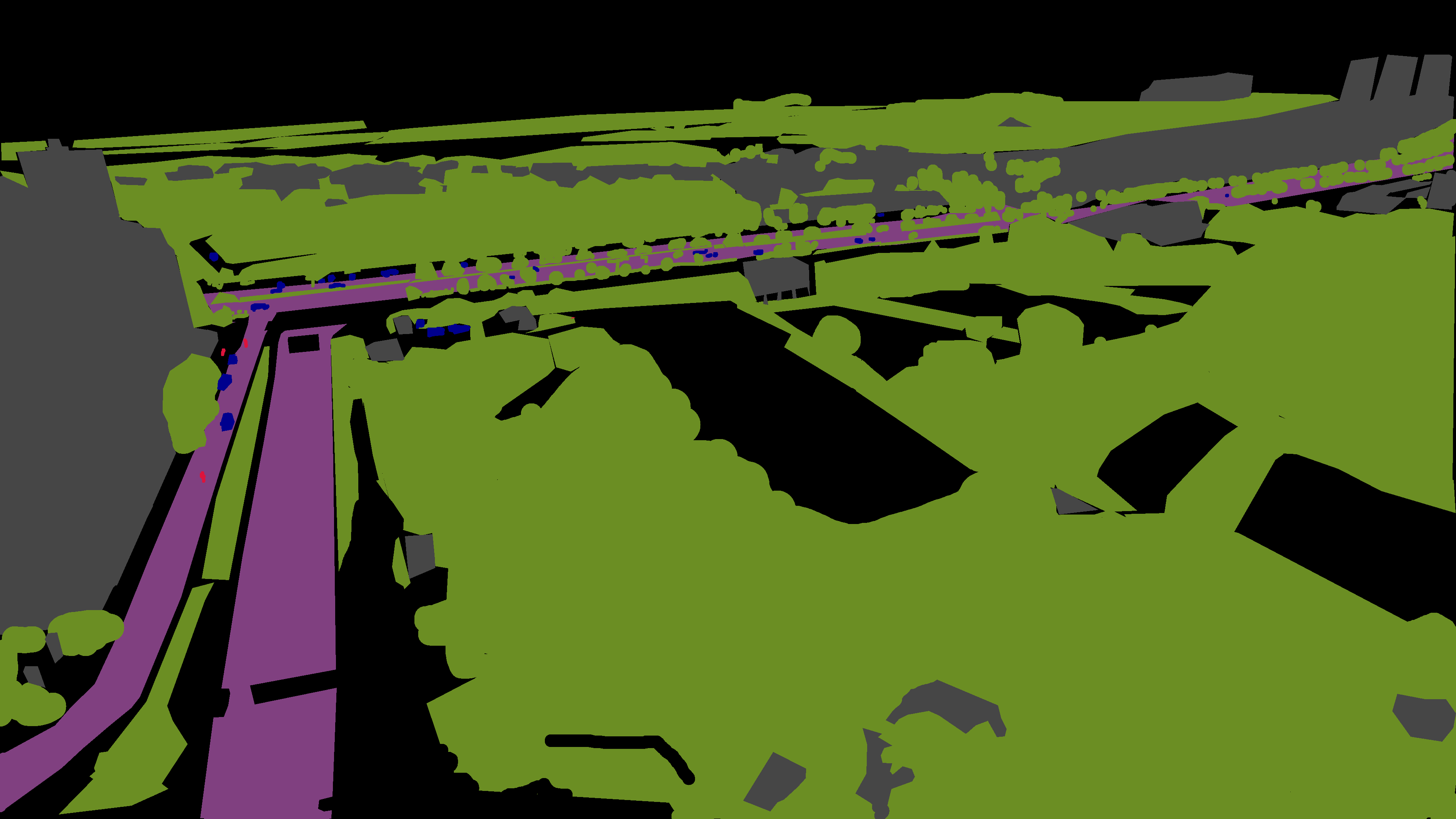}} &
    \includegraphics[width=0.15\linewidth, valign=c]{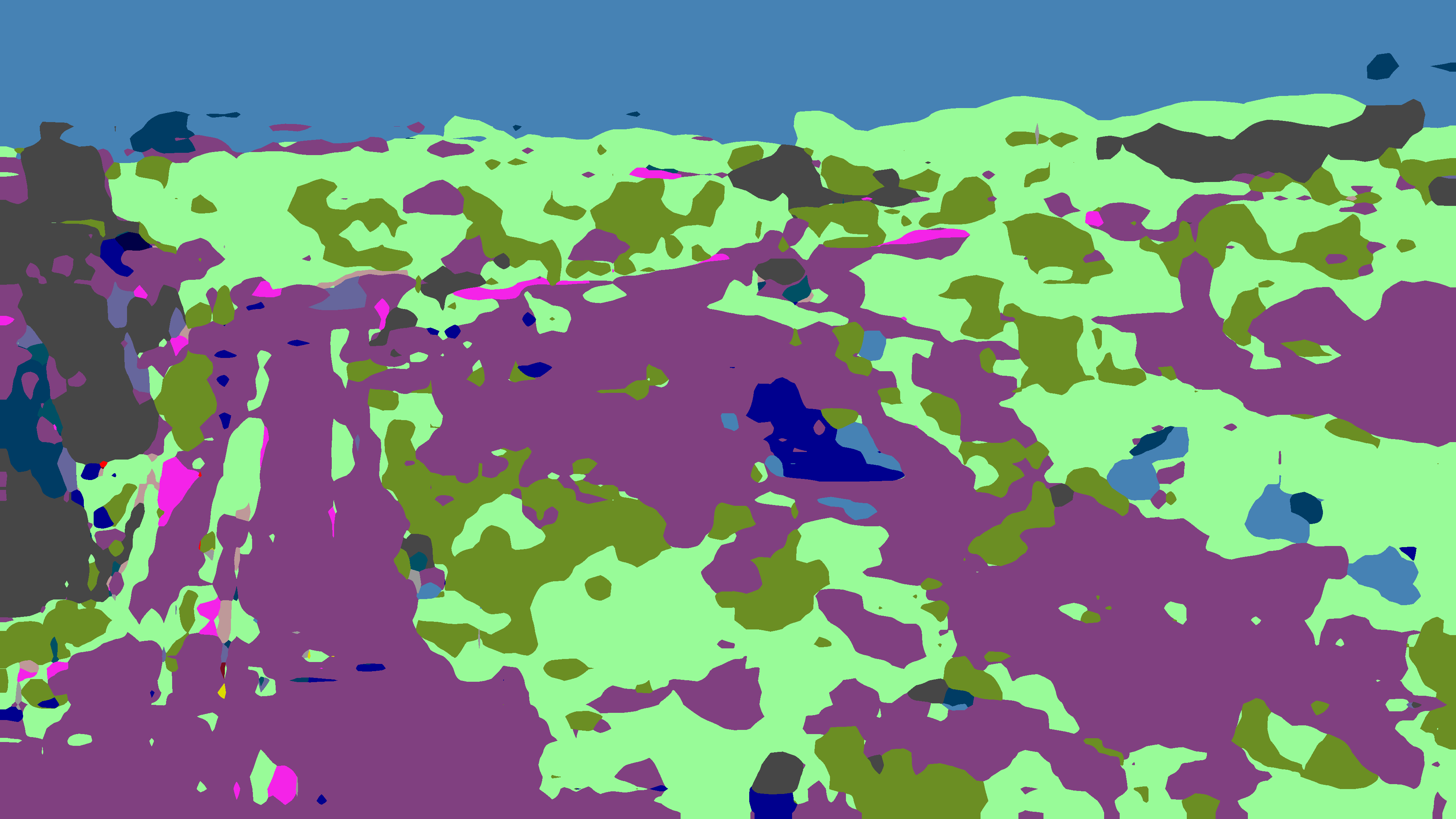} &
    \includegraphics[width=0.15\linewidth, valign=c]{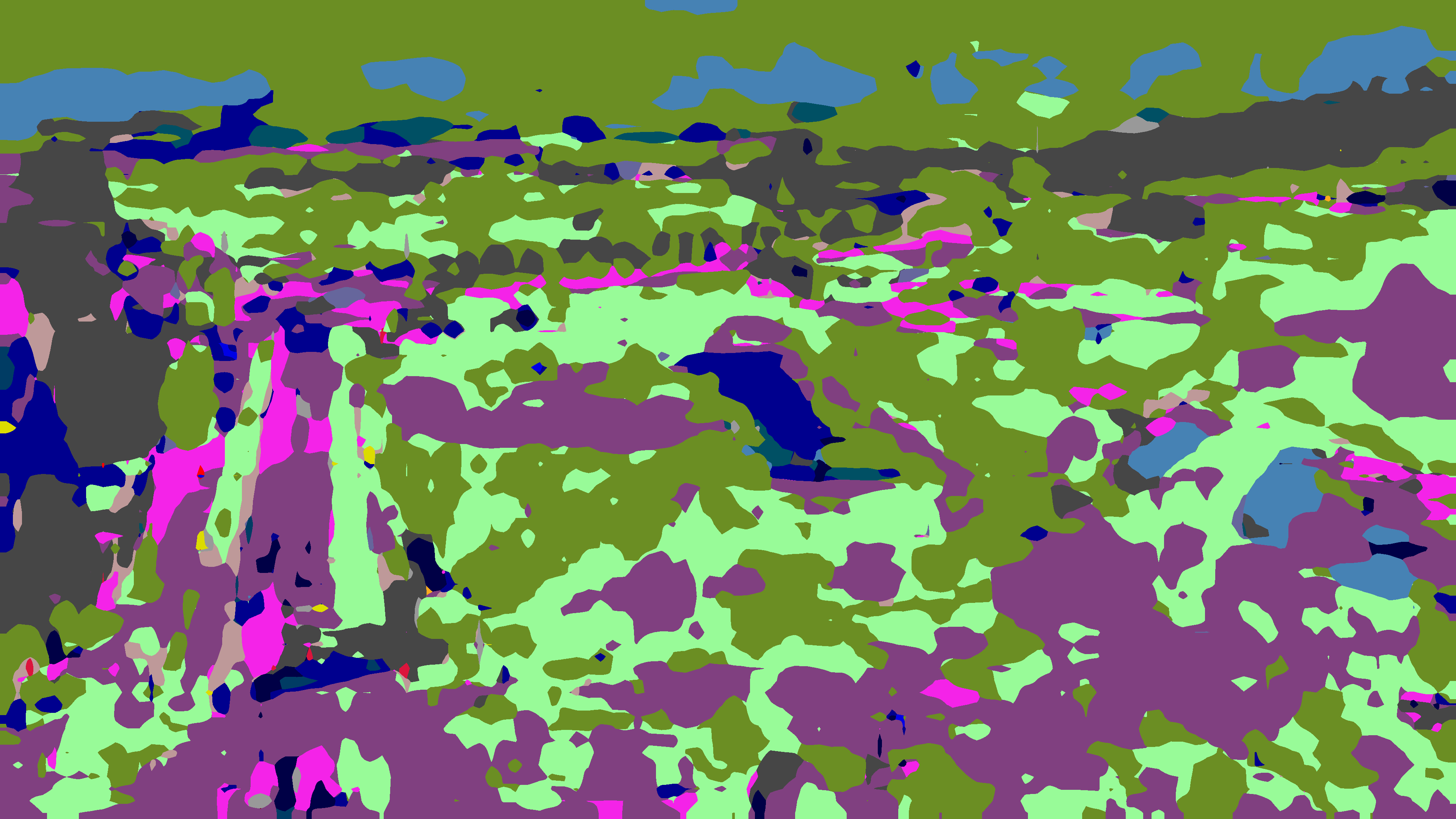} &
    \includegraphics[width=0.15\linewidth, valign=c]{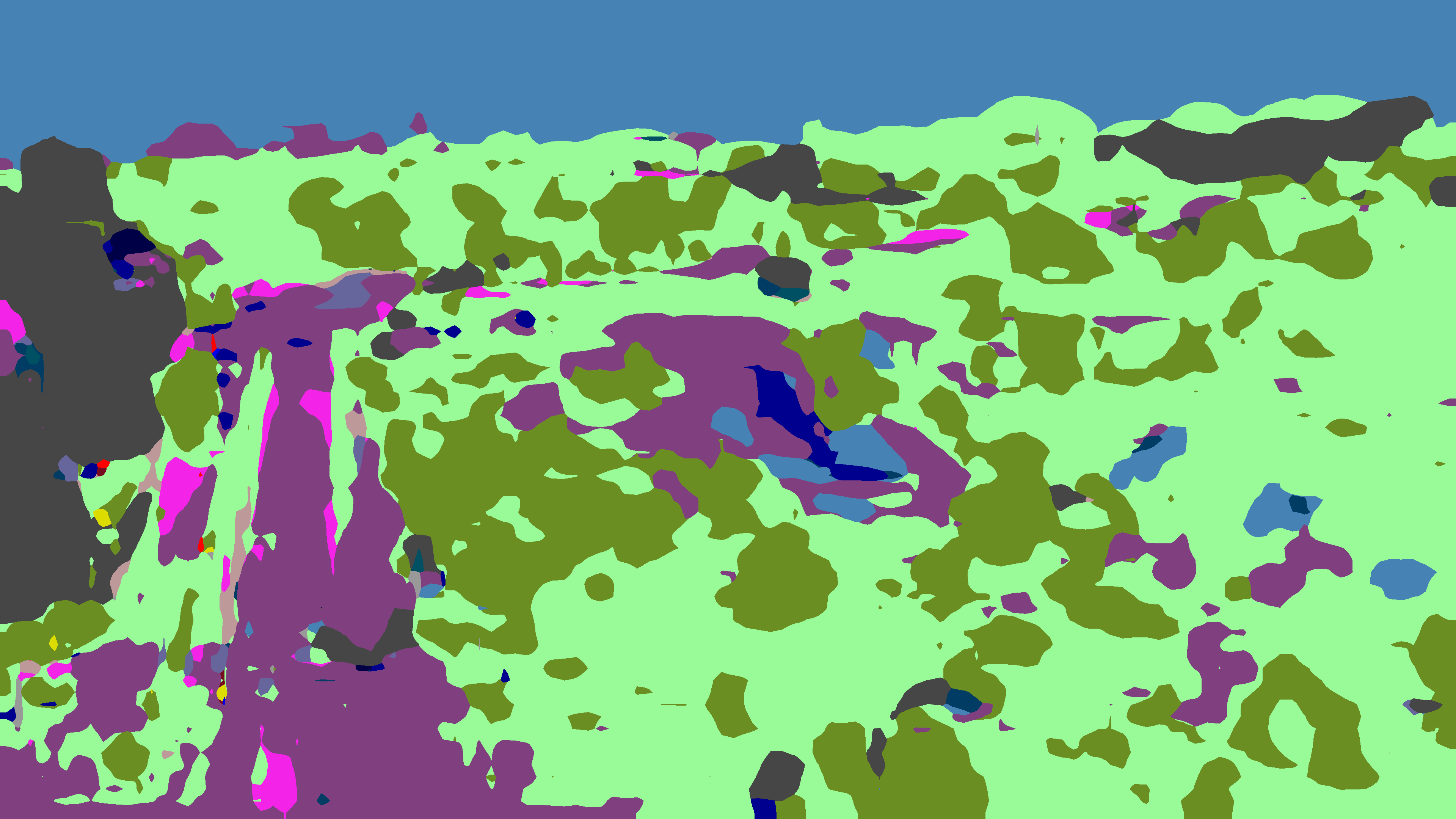} 
    \\
      \makecell{\textbf{DRONE} \\ \textbf{ADVERSE} \\ (remapped)}
    & 
    &%
    &
    \includegraphics[width=0.15\linewidth, valign=c]{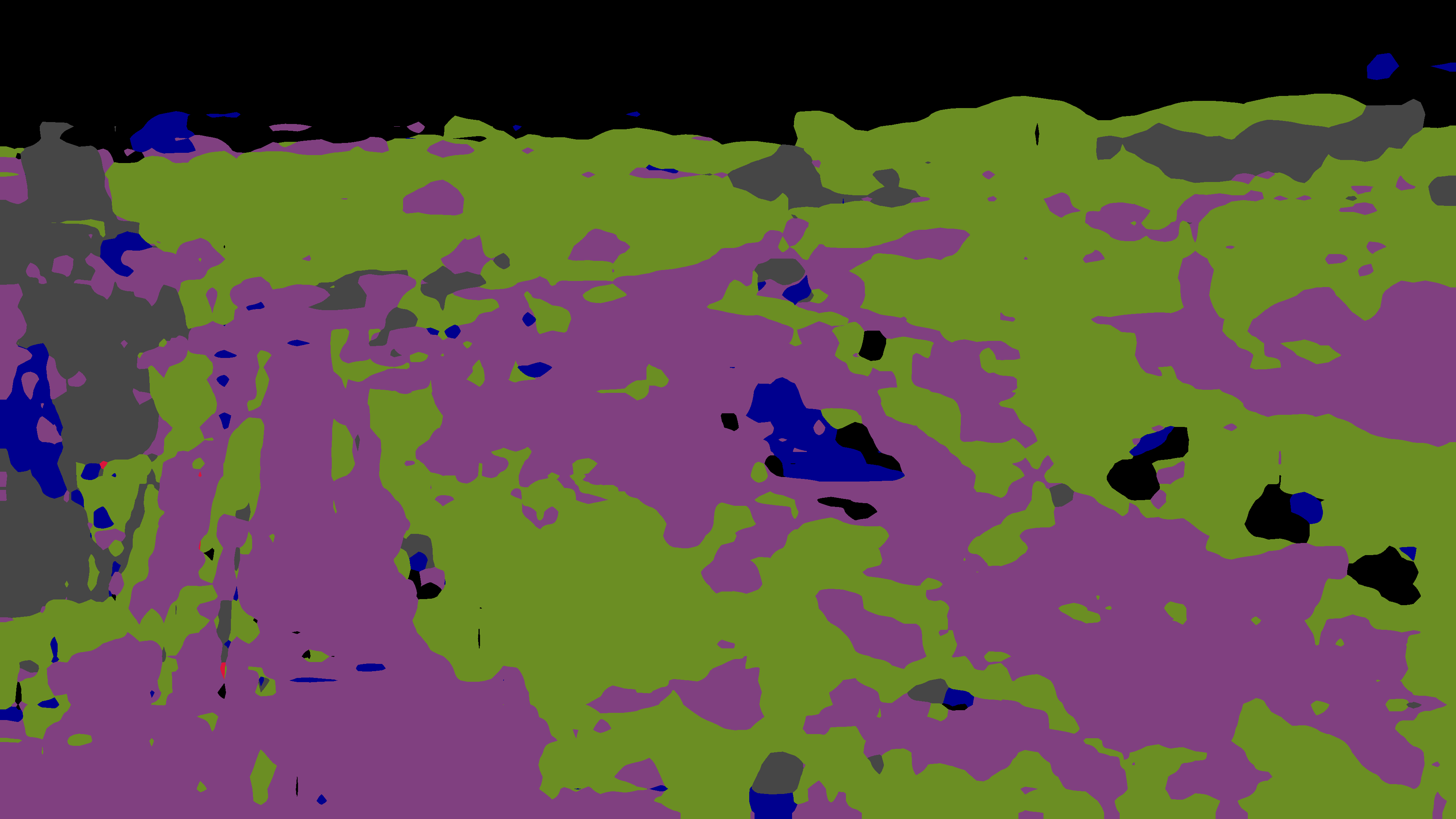} &
    \includegraphics[width=0.15\linewidth, valign=c]{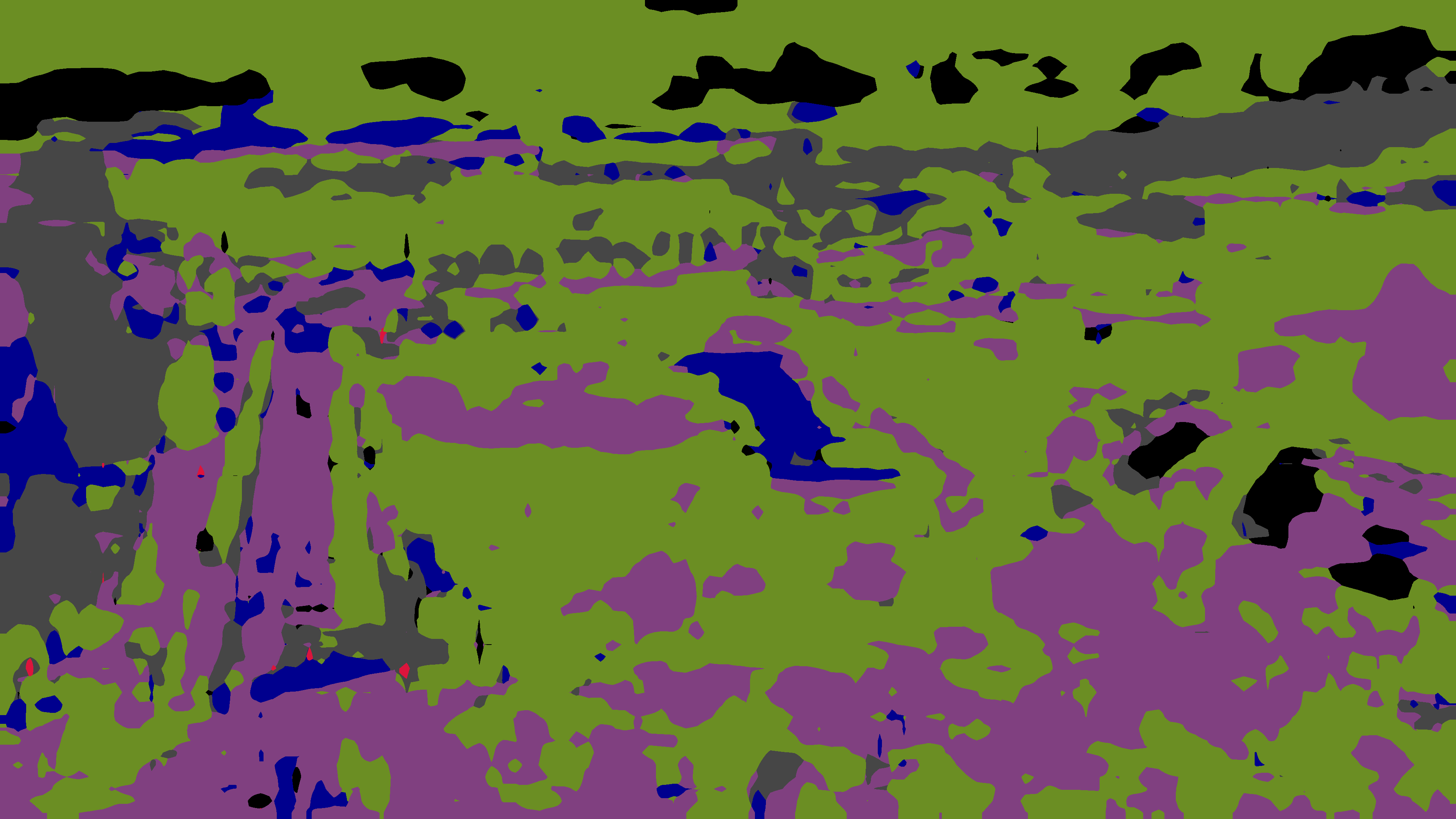} &
    \includegraphics[width=0.15\linewidth, valign=c]{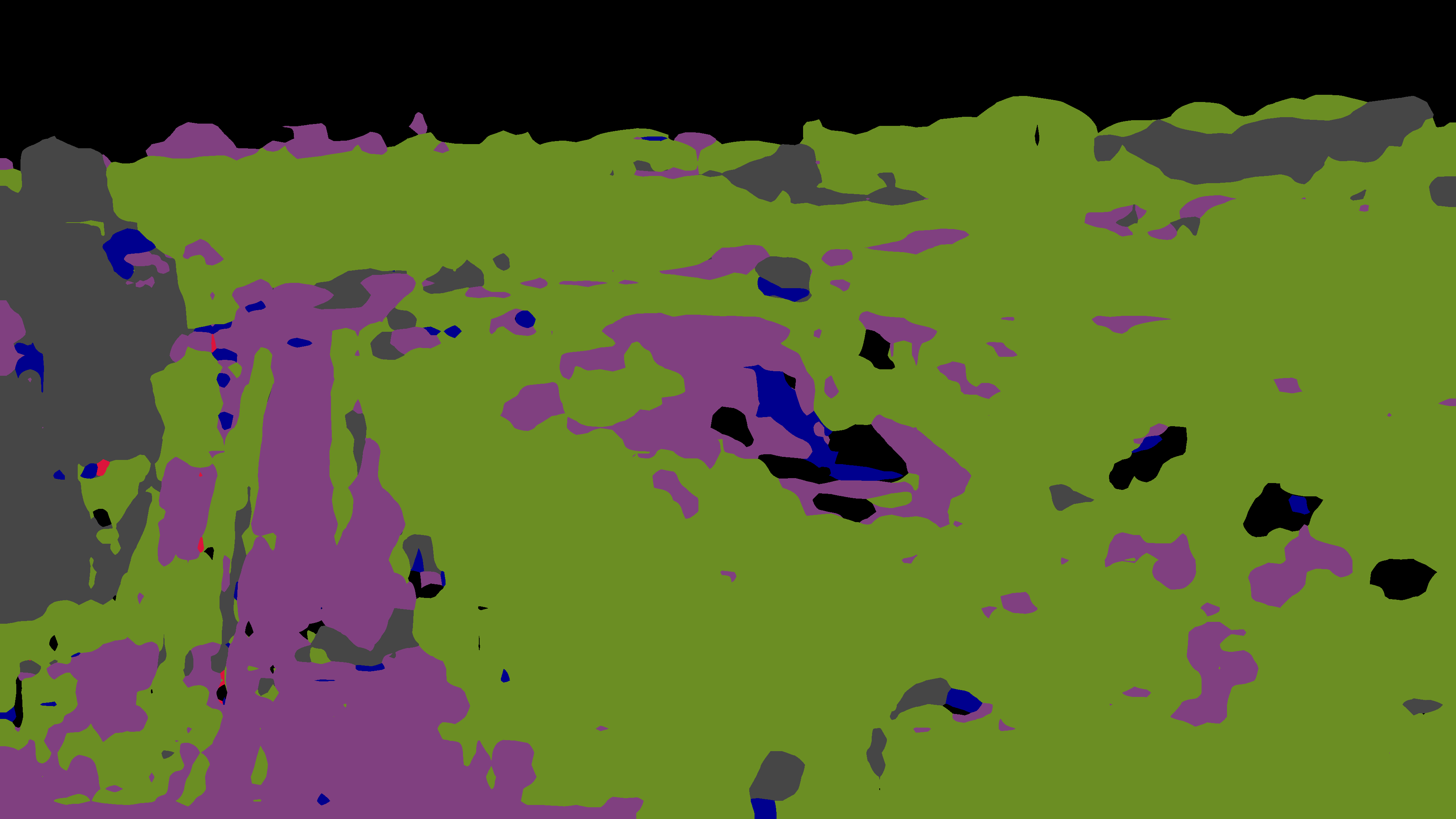} \\
  \end{tabular}
}
  \resizebox{\textwidth}{!}{%
    \begin{tabular}{c c c c c c c c c c}
                \cellcolor{road} \textcolor{white}{Road} & \cellcolor{sidewalk} Sidewalk & \cellcolor{building} \textcolor{white}{Building} & \cellcolor{wall} \textcolor{white}{Wall} & \cellcolor{fence} Fence & \cellcolor{pole} Pole & \cellcolor{tlight} T. Light & \cellcolor{tsign} T. Sign & \cellcolor{vegetation} \textcolor{white}{Vegetation} & \cellcolor{terrain} Terrain \\
                \cellcolor{sky} Sky & \cellcolor{person} \textcolor{white}{Person} & \cellcolor{rider} \textcolor{white}{Rider} & \cellcolor{car} \textcolor{white}{Car} & \cellcolor{truck} \textcolor{white}{Truck} & \cellcolor{bus} \textcolor{white}{Bus} &  \cellcolor{train} \textcolor{white}{Train} & \cellcolor{motorbike} \textcolor{white}{Motorbike} & \cellcolor{bicycle} \textcolor{white}{Bicycle} & \cellcolor{unlabelled} \textcolor{white}{Unlabeled}
    \end{tabular}
  }
  \caption{Qualitative results.}
  \label{fig:grid}
\end{figure*}

\section{Comparison over GTA $\rightarrow$ Cityscapes}
We demonstrate the advantage of our approach in transferring a more generalizable representation from synthetic to real-world urban environments, even in scenarios not involving drones. To this end, we evaluated our method on the GTA $\rightarrow$ Cityscapes domain adaptation task, with results presented in \cref{tab:gta2cityscapes}.
We utilized the same federated split of Cityscapes %
as employed in previous works \citesupp{fantauzzo2022feddrive,shenaj2023learning} to ensure a fair comparison. 
Our approach outperforms previous unsupervised methods and narrows the gap with supervised techniques. Specifically, we achieve results that are only 5.6 mIoU lower than the supervised method in \citesupp{fantauzzo2022feddrive}, while using a less complex model.

    {\small
    \bibliographystylesupp{ieee_fullname}
    \bibliographysupp{main}
    }

\end{document}